\documentclass{tlp}
\usepackage{mathptmx}

\DeclareMathAlphabet{\mathcal}{OMS}{cmsy}{m}{n}

 \usepackage{multicol}

\usepackage{booktabs}
\usepackage{enumitem}

 \usepackage{amssymb}
 \usepackage{latexsym}

\DeclareFontFamily{U}{mathb}{\hyphenchar\font45}
\DeclareFontShape{U}{mathb}{m}{n}{
      <5> <6> <7> <8> <9> <10> gen * mathb
      <10.95> mathb10 <12> <14.4> <17.28> <20.74> <24.88> mathb12
      }{}
\DeclareSymbolFont{mathb}{U}{mathb}{m}{n}
\DeclareMathSymbol{\blackdiamond}{2}{mathb}{"0C}

\usepackage{braket}

\usepackage{extpfeil}

\usepackage{pifont}

\usepackage{tikz}

\usepackage{threeparttable}

\usepackage[mathscr]{eucal}

\usepackage{verbatim}
 
\usepackage{ifthen}
\newcommand{\brifnotempty}[1]{\ifthenelse{\equal{#1}{}}{}{ \br{#1}}}
\newenvironment{lemma*}[2][]
	{\pagebreak[2] \par \noindent \textbf{Lemma~\ref{#2}}\brifnotempty{#1}.\it}{\par}
\newenvironment{theorem*}[2][]
	{\pagebreak[2] \par \noindent \textbf{Theorem~\ref{#2}}\brifnotempty{#1}.\it}{\par}
\newenvironment{proposition*}[2][]
	{\pagebreak[2] \par \noindent \textbf{Proposition~\ref{#2}}\brifnotempty{#1}.\it}{\par}
\newenvironment{corollary*}[2][]
	{\pagebreak[2] \par \noindent \textbf{Corollary~\ref{#2}}\brifnotempty{#1}.\it}{\par}

\newcommand{\an}[1]{\langle #1 \rangle}

\newcommand{\llbracket}{[\kern-.3ex[}
\newcommand{\llbracketscr}{[\kern-.25ex[}
\newcommand{\Llbracket}{\left[\kern-.6ex\left[}
\newcommand{\rrbracket}{]\kern-.3ex]}
\newcommand{\rrbracketscr}{]\kern-.25ex]}
\newcommand{\Rrbracket}{\right]\kern-.6ex\right]}
\newcommand{\sqbbr}[1]{\llbracket\hspace{.2ex}#1\hspace{.2ex}\rrbracket}
\newcommand{\sqbbrscr}[1]{\llbracketscr\hspace{.2ex}#1\hspace{.2ex}\rrbracketscr}

\newcommand{\llangle}{\langle\kern-.5ex\langle}
\newcommand{\llanglescr}{\langle\kern-.4ex\langle}
\newcommand{\Llangle}{\left\langle\kern-.8ex\left\langle}
\newcommand{\rrangle}{\rangle\kern-.5ex\rangle}
\newcommand{\rranglescr}{\rangle\kern-.4ex\rangle}
\newcommand{\Rrangle}{\right\rangle\kern-.8ex\right\rangle}
\newcommand{\anbbr}[1]{\llangle #1 \rrangle}

\newcommand{\Anbbr}[1]{\Llangle #1 \Rrangle}

\newcommand{\br}[1]{(#1)}
\newcommand{\Br}[1]{\left(#1\right)}

\newcommand{\tpl}[1]{\br{#1}}
\newcommand{\Tpl}[1]{\Br{#1}}

\newcommand{\seq}[1]{\langle #1 \rangle}

\newcommand{\lng}{n}
\newcommand{\lia}{i}
\newcommand{\lib}{j}

\newcommand{\sta}{\mathcal{S}}
\newcommand{\stb}{\mathcal{T}}

\newcommand{\ela}{s}
\newcommand{\elb}{t}

\newcommand{\pws}[1]{2^{#1}}
\newextarrow{\myxlongequal}{3300}{\Relbar\Relbar\Relbar}
\newcommand{\mlthen}{\Longrightarrow}
\newcommand{\mymodels}{\mathrel\mid\joinrel=}
\newcommand{\ent}{\mymodels}
\newcommand{\nent}{\not\ent}
\newcommand{\eq}{\equiv}

\newcommand{\prefix}[1]{\textsf{#1}}
\newcommand{\prefm}[1]{\ensuremath{\scriptscriptstyle \mathsf{#1}}}
\newcommand{\prefixm}[1]{\raisebox{-1pt}{\prefm{#1}}}

\newcommand{\pstl}[2][]{\textsf{#1#2}}
\newcommand{\const}[1]{\ensuremath{\mathit{#1}}}

\newcommand{\vara}{\mathbf{x}}

\newcommand{\prest}[1]{\ensuremath{\mathsf{#1}}}
\newcommand{\frm}{\phi}
\newcommand{\frma}{\phi}
\newcommand{\frmb}{\psi}
\newcommand{\frmu}{\mu}
\newcommand{\frmv}{\nu}
\newcommand{\twi}{I}
\newcommand{\twia}{I}
\newcommand{\twib}{J}
\newcommand{\twic}{K}
\newcommand{\twid}{L}

\newcommand{\twis}{\mathscr{I}}

\renewcommand{\mod}[1]{\sqbbr{#1}}
\newcommand{\modscr}[1]{\sqbbrscr{#1}}

\newcommand{\modc}[1]{\llbracket #1 \rrbracket_{\mC}}

\newcommand{\lang}{\mathcal{L}}

\newcommand{\atm}{\prest{p}}
\newcommand{\atma}{\prest{p}}
\newcommand{\atmb}{\prest{q}}
\newcommand{\atmc}{\prest{r}}

\newcommand{\olit}{l}
\newcommand{\olits}{\mathscr{L}}
\newcommand{\cmpl}[1]{\overline{#1}}

\newcommand{\lit}{L}
\newcommand{\slit}{S}
\newcommand{\slitp}[1][\slit]{#1^+}
\newcommand{\slitn}[1][\slit]{#1^-}
\newcommand{\lpnot}{\mathop{\sim\!}}
\newcommand{\lpif}{\leftarrow}

\newcommand{\rl}{\pi}
\newcommand{\rla}{\pi}
\newcommand{\rlb}{\sigma}

\newcommand{\hrl}[1][\rl]{H(#1)}
\newcommand{\hrla}{\hrl[\rla]}
\newcommand{\hrlb}{\hrl[\rlb]}
\newcommand{\hrlp}{\slitp[\hrl]}
\newcommand{\hrln}{\slitn[\hrl]}

\newcommand{\brl}[1][\rl]{B(#1)}

\newcommand{\brlb}{\brl[\rlb]}
\newcommand{\brlp}{\slitp[\brl]}
\newcommand{\brln}{\slitn[\brl]}

\newcommand{\prg}{\mathcal{P}\hspace{-.1ex}}
\newcommand{\prga}{\mathcal{P}\hspace{-.1ex}}
\newcommand{\prgb}{\mathcal{Q}}
\newcommand{\prgc}{\mathcal{R}}
\newcommand{\prgu}{\mathcal{U}}
\newcommand{\prgv}{\mathcal{V}}
\newcommand{\modsm}[1]{\sqbbr{#1}_{\mSM}}

\newcommand{\labbr}[1]{\pstl[\small]{(BR#1)}}
\newcommand{\smallbr}[1]{\hyperref[pstl:bu:#1]{\pstl[\scriptsize]{(BR#1)}}}

\newcommand{\labbu}[1]{\pstl[\small]{(BU#1)}}
\newcommand{\smallbu}[1]{\hyperref[pstl:bu:#1]{\pstl[\scriptsize]{(BU#1)}}}

\newcommand{\oas}{\omega}

\newcommand{\sympo}{\leq}
\newcommand{\po}[2]{\sympo^{#1}_{#2}}

\renewcommand{\pod}[1]{\po{#1}{\oas}}

\newcommand{\poeld}{\pod{\ela}}

\newcommand{\potwd}{\pod{\twia}}

\newcommand{\symspo}{<}
\newcommand{\spo}[2]{\symspo^{#1}_{#2}}

\newcommand{\spod}[1]{\spo{#1}{\oas}}
\newcommand{\spow}[1]{\spo{#1}{\prefixm{W}}}

\newcommand{\spotwd}{\spod{\twia}}
\newcommand{\spotww}{\spow{\twia}}

\newcommand{\ropb}{\mathbin{\circ}}
\newcommand{\copb}{\mathbin{\ast}}
\newcommand{\bigcopb}{\ensuremath \raisebox{-1pt}{{\Large$\ast$}}}

\newcommand{\uopb}{\mathbin{\diamond}}
\newcommand{\uopbw}{\mathbin{\uopb_{\prefixm{W}}}}
\newcommand{\C}[1]{\prefix{C}\protect\nobreakdash#1\hspace{0pt}}
\newcommand{\mC}{\prefixm{C}}

\newcommand{\dprg}{\boldsymbol{P}}

\newcommand{\expv}[1]{#1^{\mathsf{e}}}

\newcommand{\confl}[2]{\Join^{#1}_{#2}}

\newcommand{\all}[1]{\mathsf{all}(#1)}

\newcommand{\rej}[2][]{\mathsf{rej}_{#1}(#2)}

\newcommand{\defs}[1]{\mathsf{def}(#1)}

\newcommand{\least}[1]{\mathsf{least}(#1)}

\renewcommand{\S}[2][S]{\prefix{#1}\protect\nobreakdash#2\hspace{0pt}}

\newcommand{\mS}{\prefixm{S}}
\newcommand{\mods}[1]{\sqbbr{#1}_{\mS}}

\newcommand{\moduopr}[1]{\sqbbr{#1}_{\textsf{\S{}}_{\uopr}}}

\newcommand{\JU}[1]{\prefix{JU}\protect\nobreakdash#1\hspace{0pt}}
\newcommand{\mJU}{\prefixm{JU}}
\newcommand{\rejju}[1]{\rej[\mJU]{#1}}
\newcommand{\modju}[1]{\sqbbr{#1}_{\mJU}}

\newcommand{\AS}[1]{\prefix{AS}\protect\nobreakdash#1\hspace{0pt}}
\newcommand{\mAS}{\prefixm{AS}}
\newcommand{\rejas}[1]{\rej[\mAS]{#1}}
\newcommand{\modas}[1]{\sqbbr{#1}_{\mAS}}

\newcommand{\DS}[1]{\prefix{DS}\protect\nobreakdash#1\hspace{0pt}}
\newcommand{\mDS}{\prefixm{DS}}
\newcommand{\rejds}[1]{\rej[\mDS]{#1}}
\newcommand{\modds}[1]{\sqbbr{#1}_{\mDS}}

\newcommand{\RD}[1]{\prefix{RD}\protect\nobreakdash#1\hspace{0pt}}

\newcommand{\mRD}{\prefixm{RD}}
\newcommand{\rejrd}[1]{\rej[\mRD]{#1}}
\newcommand{\modrd}[1]{\sqbbr{#1}_{\mRD}}

\newcommand{\rejlm}[1]{\rej[\lm]{#1}}

\newcommand{\PRZ}[1]{\prefix{PRZ}\protect\nobreakdash#1\hspace{0pt}}
\newcommand{\mPRZ}{\prefixm{PRZ}}
\newcommand{\modprz}[1]{\sqbbr{#1}_{\mPRZ}}

\newcommand{\PRDi}[1]{\prefix{PRD\ensuremath{_i}}\protect\nobreakdash#1\hspace{0pt}}
\newcommand{\PRWi}[1]{\prefix{PRW\ensuremath{_i}}\protect\nobreakdash#1\hspace{0pt}}
\newcommand{\PRBi}[1]{\prefix{PRB\ensuremath{_i}}\protect\nobreakdash#1\hspace{0pt}}

\newcommand{\PRXi}[1]{\prefix{PRX\ensuremath{_i}}\protect\nobreakdash#1\hspace{0pt}}
\newcommand{\PRXtwo}[1]{\prefix{PRX\ensuremath{_2}}\protect\nobreakdash#1\hspace{0pt}}
\newcommand{\mPRXi}{\prefixm{PRX_i}}
\newcommand{\mPRXiX}[2]{\prefixm{PR{#1}_{#2}}}
\newcommand{\modprxi}[1]{\sqbbr{#1}_{\mPRXi}}
\newcommand{\modprxiX}[3]{\sqbbr{#3}_{\mPRXiX{#1}{#2}}}

\newcommand{\RVS}[1]{\prefix{RVS}\protect\nobreakdash#1\hspace{0pt}}
\newcommand{\mRVS}{\prefixm{RVS}}
\newcommand{\modrvs}[1]{\sqbbr{#1}_{\mRVS}}

\newcommand{\RVD}[1]{\prefix{RVD}\protect\nobreakdash#1\hspace{0pt}}
\newcommand{\mRVD}{\prefixm{RVD}}
\newcommand{\modrvd}[1]{\sqbbr{#1}_{\mRVD}}
\newcommand{\preds}[1]{\mathsf{pr}(#1)}

\newcommand{\ctau}{\tau}

\newcommand{\prgand}{\mathbin{\dot{\wedge}}}
\newcommand{\prgor}{\mathbin{\dot{\lor}}}

\newcommand{\uopr}{\mathbin{\oplus}}
\newcommand{\biguopr}{\mathop{\bigoplus}}
\newcommand{\ropr}{\mathbin{\otimes}}
\newcommand{\tri}{X}
\newcommand{\tria}{X}
\newcommand{\trib}{Y}

\newcommand{\tris}{\mathscr{X}}
\newcommand{\twiab}{\tpl{\twia, \twib}}
\newcommand{\twibb}{\tpl{\twib, \twib}}

\newcommand{\stri}{\mathcal{M}}
\newcommand{\stria}{\mathcal{M}}
\newcommand{\strib}{\mathcal{N}}

\newcommand{\sstria}{\mathcal{S}}
\newcommand{\sstrib}{\mathcal{T}}

\newcommand{\tr}{\mathsf{T}}
\newcommand{\fa}{\mathsf{F}}
\newcommand{\un}{\mathsf{U}}
\newcommand{\val}{\mathsf{V}}

\newcommand{\potrd}{\pod{\tria}}
\newcommand{\potrsd}{\pod{\tria^*}}

\newcommand{\spotrd}{\spod{\tria}}

\newcommand{\HT}[1]{\prefix{HT}\protect\nobreakdash#1\hspace{0pt}}
\newcommand{\SE}[1]{\prefix{SE}\protect\nobreakdash#1\hspace{0pt}}
\newcommand{\mHT}{\prefixm{HT}}
\newcommand{\eqHT}{\equiv_{\mHT}}
\newcommand{\entHT}{\ent_{\mHT}}
\newcommand{\modHT}[1]{\sqbbr{#1}_{\mHT}}
\newcommand{\modsescr}[1]{\sqbbrscr{#1}_{\mHT}}

\newcommand{\labprse}[1]{\pstl[\small]{(PR#1)}$_{\prefix{\tiny HT}}$}
\newcommand{\smallprse}[1]{\hyperref[pstl:prse:#1]{\pstl[\scriptsize]{(PR#1)$_{\prefix{\tiny HT}}$}}}

\newcommand{\labpuse}[1]{\pstl[\small]{(PU#1)}$_{\prefix{\tiny HT}}$}
\newcommand{\smallpuse}[1]{\hyperref[pstl:puse:#1]{\pstl[\scriptsize]{(PU#1)$_{\prefix{\tiny HT}}$}}}

\newcommand{\labpurr}[1]{\pstl[\small]{(PU#1)}$_{\prefix{\tiny RR}}$}

\newcommand{\X}[2][X]{\prefix{#1}\protect\nobreakdash#2\hspace{0pt}}
\newcommand{\mSM}{\prefixm{SM}}
\newcommand{\SMR}[1]{\prefix{SMR}\protect\nobreakdash#1\hspace{0pt}}
\newcommand{\RE}[1]{\prefix{RE}\protect\nobreakdash#1\hspace{0pt}}
\newcommand{\mRE}{\prefixm{RE}}
\newcommand{\eqRE}{\equiv_{\mRE}}
\newcommand{\modre}[1]{\sqbbr{#1}_{\mRE}}
\newcommand{\modrer}[1]{\anbbr{#1}_{\mRE}}

\newcommand{\Modrer}[1]{\Anbbr{#1}_{\mRE}}

\newcommand{\RR}[1]{\prefix{RR}\protect\nobreakdash#1\hspace{0pt}}
\newcommand{\mRR}{\prefixm{RR}}
\newcommand{\eqRR}{\equiv_{\mRR}}
\newcommand{\uope}{\oplus}
\newcommand{\e}{\varepsilon}
\newcommand{\te}[1]{\ensuremath{\e}\protect\nobreakdash#1\hspace{0pt}}

\newcommand{\er}{\delta}
\newcommand{\erone}{\er_\mathsf{a}}
\newcommand{\terone}[1]{$\delta_\textsf{a}$\protect\nobreakdash#1\hspace{0pt}}
\newcommand{\ertwo}{\er_\mathsf{b}}
\newcommand{\tertwo}[1]{$\delta_\textsf{b}$\protect\nobreakdash#1\hspace{0pt}}
\newcommand{\terthree}[1]{$\delta_\textsf{c}$\protect\nobreakdash#1\hspace{0pt}}

\newcommand{\labpu}[1]{\pstl[\small]{(P#1)}}
\newcommand{\labputwotop}{\pstl[\small]{(P2.$\top$)}}
\newcommand{\labpup}[1]{\pstl[\small]{(#1)}}
\newcommand{\pu}[1]{\hyperref[tbl:semantic properties]{\labpu{#1}}}

\newcommand{\pup}[1]{\hyperref[tbl:semantic properties]{\labpup{#1}}}

\newcommand{\lm}{\ell}
\newtheorem{theorem}{Theorem}

\newtheorem{definition}[theorem]{Definition}
\newtheorem{example}[theorem]{Example}

\usepackage[latin1]{inputenc}
\usepackage{textcomp}

\begin{document}

\lefttitle{Jo\~ao Leite and Martin Slota}

\jnlPage{\pageref{firstpage}}{\pageref{lastpage}}
\jnlDoiYr{2022}
\doival{10.1017/xxxxx}

\title{A Brief History of Updates of Answer-Set Programs}

\begin{authgrp}
\author{JO\~AO LEITE}
\affiliation{NOVA LINCS, Universidade Nova de Lisboa, Portugal}
\author{MARTIN SLOTA}
\affiliation{NOVA LINCS, Universidade Nova de Lisboa, Portugal}
\end{authgrp}

\maketitle

\begin{abstract}
Over the last couple of decades, there has been a considerable effort devoted to the problem of updating logic programs under the stable model semantics (a.k.a. answer-set programs) or, in other words, the problem of characterising the result of bringing up-to-date a logic program when the world it describes changes. Whereas the state-of-the-art approaches are guided by the same basic intuitions and aspirations as belief updates in the context of classical logic, they build upon fundamentally different principles and methods, which have prevented a unifying framework that could embrace both belief and rule updates. In this paper, we will overview some of the main approaches and results related to answer-set programming updates, while pointing out some of the main challenges that research in this topic has faced.
\end{abstract}

  \begin{keywords}
belief update; belief change; logic programming; answer-set programming
  \end{keywords}

\tableofcontents

\section{Introduction}
\label{sec:intro}

%!TEX root =  paper.tex

In this paper, we will take a historical journey through some of the main approaches proposed to deal with the problem of updating logic programs under the stable model semantics.

Knowledge-based systems must keep a representation of the world --- often encoded in some logic-based language equipped with a formal semantics and reasoning mechanisms --- which is then used for reasoning, e.g., to automate decision making. Whereas languages that are based on classical logic --- hence monotonic --- like description logics, are often used, it has been known for several decades that non-monotonic features are important for common-sense reasoning, e.g., to properly deal with default information, preferences, the frame problem, etc. Of the many existing languages for knowledge representation that exhibit non-monotonic features, logic programming under the stable model semantics (a.k.a. answer-set programming, or ASP, for short), introduced by \citeN{Gelfond1988,Gelfond1991}, is perhaps the biggest success story so far. Established some 30 years ago, ASP is similar in syntax to traditional logic programming, and has a simple and well-understood non-monotonic declarative semantics with known relationships with other logic-based formalisms such as default logic, autoepistemic logic, propositional and predicate logic, etc. (cf. the work of \citeN{DBLP:conf/iclp/Lifschitz08} and references therein). Its rich expressive power allows to compactly represent all NP and coNP problems if non-disjunctive logic programs are used, while disjunctive logic programs capture the complexity class $\Sigma_{2}^{p}$ and $\Pi_{2}^{p}$ \citep{44EiterGM1997}. Additionally, the existence of efficient implementations such as \emph{clasp} \citep{DBLP:journals/aicom/GebserKKOSS11} and \emph{dlv} \citep{DBLP:journals/tocl/LeonePFEGPS06} has made it possible to use ASP in significant applications in diverse areas such as 
configuration, diagnosis and repair, planning, classification, scheduling, robotics, information integration, legal reasoning, computational biology and bioinformatics, e-medicine, and decision support systems (cf. the surveys by \citeN{DBLP:journals/aim/ErdemGL16}, \citeN{DBLP:journals/ki/ErdemP18},\citeN{DBLP:journals/ki/FalknerFSTT18}, and references therein).

One of the more recent challenges for knowledge engineering and information
management is to efficiently and plausibly deal with the incorporation of new,
possibly conflicting knowledge and beliefs. 
There are several domains where this may be required.  
For example, knowledge-based systems that check for legal and regulatory compliance --- such as a public procurement monitoring system --- need to keep track of changing laws and regulations, and reason with them. Simply adding the new laws and regulations to the knowledge base containing the older ones would not work, since the newer may be in conflict with the older. It may also not be as simple as deleting the old conflicting ones, since the conflict may be contingent on particular cases. For example, some initial regulation may state that institutions are allowed to enter a contract without a public offer if the contract's value is below some fixed amount, and a later regulation may state that publicly funded foundations (a special kind of an institution) that have not filed their previous year's tax return, are not allowed to enter a contract without a public offer. The knowledge-based system would have to automatically deal with these cases, in a way similar to the legal principle of \emph{lex posterior} used by judges and other legal practitioners, according to which a newer law repeals an earlier conflicting one. 
Other domains where dealing with dynamic, possibly conflicting, knowledge and beliefs is important include multi-agent systems, where agents need to change their knowledge and beliefs to properly reflect their observations, including incoming messages from other agents, and even new norms, possibly resulting in a change in behaviour; stream reasoning systems that learn/extract knowledge from streams of data, which may be in conflict with previously learnt knowledge; or even transfer learning, where what is learnt by one system in one domain serves as initial knowledge when that system is placed in a different domain, which is then changed as new knowledge is learnt in the new domain. Indeed, any knowledge-based system that maintains a knowledge base about a dynamic world, i.e., a world that changes, needs to efficiently and plausibly deal with the incorporation of new, possibly conflicting knowledge.

The problems associated with the evolution of knowledge have been extensively
studied over the years, in the context of classical logic. Most of this work was inspired by the seminal contribution of
Alchourr{\'{o}}n, G{\"{a}}rdenfors and Makinson (AGM) who proposed a set of
desirable properties of belief change operators, now called \emph{AGM
postulates} \citep{Alchourron1985}. Subsequently, \emph{update} and
\emph{revision} have been distinguished as two very much related but ultimately
different belief change operations \citep{Keller1985,Winslett1990,Katsuno1991}.
While revision deals with incorporating new \emph{better} information about a \emph{static}
world, update takes place when changes occurring in a \emph{dynamic} world are
recorded. Katsuno and Mendelzon formulated a separate set of postulates for
update, now known as \emph{KM postulates}. For a comprehensive treatment of the subject and further references on belief change in classical logic,
the reader is referred to the survey paper by \citeN{DBLP:journals/jphil/FermeH11a}.

Despite the large body of research on belief change in general, and on updates in particular, in the context of classical logic, 
the use of ASP for knowledge representation in dynamic domains, mainely due to its rule-based syntax and non-monotonic semantics, 
has called for a specific line of research on how to update an answer-set program, which constitutes the central topic of this paper.

To illustrate the problem at hand, consider an agent with knowledge represented by the following program $\prga$ (where $\lpnot$ denotes default negation):
	\begin{align}
		\prest{goHome} &\lpif \lpnot \prest{money}.
			\label{eq:ex:intro:rule update on models:1} \\
		\prest{goRestaurant} &\lpif \prest{money}.
			\label{eq:ex:intro:rule update on models:2} \\
		\prest{money} &.
			\label{eq:ex:intro:rule update on models:3}
	\end{align}
The only stable model of $\prga$ is $\twia = \Set{\prest{money},
	\prest{goRestaurant}}$, capturing that the agent has money by rule
	\eqref{eq:ex:intro:rule update on models:3}, so according to rule
	\eqref{eq:ex:intro:rule update on models:2} it plans to go to a restaurant.
Suppose that the beliefs of the agent are to be updated by the program $\prgu$ with the following two rules:
	\begin{align*}
		\lpnot \prest{money} &\lpif \prest{robbed}.
		& \prest{robbed} &.
	\end{align*}
What  should the agent's beliefs be after the update of $\prga$ by $\prgu$?

The central research question is then how to semantically characterise 
pairs or, more generally, sequences of ASP programs (a.k.a. \emph{dynamic logic programs}) where 
each component represents an update of the preceding ones, and, if possible,
produce an ASP program that encodes the result of the updates.

Going back to the previous example, when we update $\prga$ by $\prgu$, the intuitively correct result is that \prest{robbed} is true and $\prest{money}$ is false, because of the rules in the update $\prgu$, and
	that $\prest{goRestaurant}$ should now be false because its only
	justification, $\prest{money}$, is no longer true. Furthermore, we would expect
	that $\prest{goHome}$ be now true, i.e.,\ the rule \eqref{eq:ex:intro:rule
	update on models:1} should be triggered because $\prest{money}$ became
	false.

Over the years, many have tackled this issue, which has proved far more difficult and elusive than perhaps originally thought. 
Earlier approaches \citep{DBLP:conf/jelia/MarekT94,Marek1998,DBLP:conf/lpnmr/PrzymusinskiT95,DBLP:journals/jlp/PrzymusinskiT97,Alferes1996} were based on literal inertia, following some of the basic principles inherited from the \emph{possible models approach} by \citeN{Winslett1988} --- an approach to updates in propositional logic based on minimising the set of atoms whose truth value changes when an
interpretation is updated that satisfies the KM postulates. However, they were soon found to be inadequate, or at least not sufficiently expressive to capture the result of updating an answer-set program by means of another answer-set program \citep{Leite1997}.

Since then, many different approaches were put forward. Though the state-of-the-art approaches are guided by
the same basic intuitions and aspirations as belief change in classical logic, they build upon
fundamentally different principles and methods. 
While many are based on the so-called
\emph{causal rejection principle}
\citep{Leite1997,Alferes2000,Eiter2002,Alferes2005,Osorio2007}, others employ
syntactic transformations and other methods, such as abduction
\citep{Sakama2003}, forgetting \citep{Zhang2005}, prioritisation
\citep{Zhang2006}, preferences \citep{Delgrande2007}, or dependencies on
defeasible assumptions \citep{Sefranek2011,Krumpelmann2012}.

One interesting feature about these developments is that the resulting
operators often bear characteristics of \emph{revision} rather than \emph{update}, as viewed from the
perspective of belief change in classical logic, often blurring the frontier between these two types of operations. 
We will review not only those approaches that aim to deal with updates, but
also  some operators that are closer to revision, but whose authors position their contribution
as having similar goals, namely by comparing them with the operators specifically defined for updates. 
However, while evaluating the reviewed approaches, we will guide ourselves by the basic idea of 
what an update is, i.e., an operation that deals with incorporating new information about a \emph{dynamic} world, as opposed to 
a revision which deals with incorporating better knowledge about a \emph{static} world.
In our opinion, underlying this definition of update is the assumption that the information acquired at each point in time
is correct at that moment, as opposed to the assumption adopted in revision, whereby the information acquired at each time point is not necessarily correct, just better than what we had before. 
One relevant consequence is that an empty or tautological update (which we take to represent that nothing changed) should have no effect, because our beliefs about the world were correct and nothing changed, while it is acceptable that a revision by an empty or tautological update may lead to some change, for example restoring consistency, because we take it that in revision, our beliefs about the world were not necessarily correct.

More recently, both the AGM and KM postulates were revisited, taking into account a monotonic characterisation of ASP --- \HT-models \citep{Pearce1997,Lifschitz2001} --- as the basis for new classes of revision and update operators \citep{DBLP:journals/tocl/DelgrandeSTW13,Slota2013a}. Despite their own merits and shortcomings, these approaches based on \HT-models have opened up new avenues to investigate other belief change operations in ASP, such as forgetting \citep{WangZZZ12,WangWZ13,WangZZZ14,DBLP:conf/aaai/DelgrandeW15} (see the paper by \citeN{DBLP:conf/epia/GoncalvesKL16} for a survey on forgetting in ASP, or another survey in this volume), and new insights into unifying classical logic and ASP-based updates \citep{Slota2012,Slota2012a,DBLP:journals/ai/SlotaLS15}.

In this paper, we will briefly revisit the main landmarks in the history of updating answer-set programs. We will follow a chronological approach that divides it into three main eras, although not without intersection --- model updates, syntax-based updates, and semantics-based updates --- that correspond to the three main sections of this paper. These sections are preceded by a brief section (pre-history) on belief change in classical logic, and followed by an outlook where some current and future issues are discussed. Before we begin the journey, we provide a brief background section with the usual preliminaries on propositional logic, answer-set programming, and other basic concepts used throughout the paper.

Given the nature of this paper, we will often exercise some restraint on technical content in favour of better conveying the main intuitions and concepts underlying each approach and focusing on providing simple examples that bring forward their main differentiating features. Despite this compromise, this paper still conveys a novel and, we hope, significant scientific contribution, inasmuch as it is, to the best of our knowledge, the first encompassing and critical survey of the field, not only comparing different approaches but sometimes even providing intuitions and illustrative examples that were absent from the original papers.

\section{Background}
%!TEX root =  paper.tex

\noindent \textbf{Propositional Logic.}
We consider a propositional language over a finite set of propositional
variables $\lang$ and the usual set of propositional connectives to form
propositional formulae. A (two-valued) interpretation is any $\twi \subseteq
\lang$. The set of all (two-valued) interpretations is denoted by $\twis$.
Each atom $\atm$ is assigned one of two truth values in $\twi$:
$\twi(\atm) = \tr$ if $\atm \in \twi$ and $\twi(\atm) = \fa$ otherwise. This
assignment is generalised in the standard way to all propositional formulae.
The set of all models of a formula $\frm$ is denoted by $\mod{\frm}$. We say
$\frm$ is \emph{complete} if $\mod{\frm}$ is a singleton set. For two formulae
$\frma, \frmb$ we say that $\frma$ \emph{entails} $\frmb$, denoted by $\frma
\ent \frmb$, if $\mod{\frma} \subseteq \mod{\frmb}$, and that $\frma$ \emph{is
equivalent to} $\frmb$, denoted by $\frma \equiv \frmb$, if $\mod{\frma} =
\mod{\frmb}$.

\noindent \textbf{Answer-Set Programs.}
Answer-set programming (a.k.a. logic programming under the stable model semantics) has its roots in classical logic. However, answer-set programs diverge from classical semantics
by adopting the closed world assumption and allowing for non-monotonic
inferences. Here, we introduce the class of answer-set programs
that allow for both disjunction and default negation in heads of rules.

The basic syntactic building blocks of rules are also propositional atoms from
$\lang$. A \emph{negative literal} is an atom preceded by $\lpnot{}$ denoting
default negation. A \emph{literal} is either an atom or a negative literal. Throughout this paper, we adopt a convention that double default negation is absorbed, so that $\lpnot \lpnot
\atm$ denotes the atom $\atm$. Given a set $\slit$ of literals, we introduce
the following notation: $\slit^+ = \Set{\atm \in \lang | \atm \in \slit}$,
$\slit^- = \Set{\atm \in \lang | \lpnot \atm \in \slit}$, and $\lpnot \slit =
\Set{\lpnot \lit | \lit \in \slit}$.

A \emph{rule} is a pair of sets of literals $\rl = \an{\hrl, \brl}$. We say
that $\hrl$ is the \emph{head of $\rl$} and $\brl$ is the \emph{body of
$\rl$}. Usually, for convenience, we write $\rl$ as
\[
	\hrl^+; \lpnot \hrl^- \lpif \brl^+, \lpnot \brl^-.
\]
A rule is called \emph{non-disjunctive} if its head contains at most one
literal; a \emph{fact} if its head contains exactly one literal and its body
is empty; an \emph{integrity constraint} if its head is empty.
A \emph{program} is any set of rules. A program is
\emph{non-disjunctive} if all its rules are.

We define the class of \emph{acyclic programs} using level
mappings \citep{Apt1991}.
A \emph{level mapping} is a function $\ell$ that assigns a natural number to
	every atom, and is extended to default literals
	and sets of literals by putting $\ell(\lpnot \lit) = \ell(\lit)$ and
	$\ell(\slit) = \max \set{\ell(\lit) | \lit \in \slit}$.
We say that a program $\prg$ is \emph{acyclic} if there exists a level
	mapping $\ell$ such that for every rule $\rl \in \prg$ it holds that $\hrl\neq\emptyset$ and
	$\ell(\olit_H) > \ell(\olit_B)$ for every $\olit_H \in \hrl$ and every $\olit_B \in \brl$.

In the following, we define the answer-sets (a.k.a. stable models) of a program 
\citep{Gelfond1988,Gelfond1991} as well as two monotonic model-theoretic
characterisations of rules and programs. One is that of classical models, where a rule is
simply treated as a classical implication. The other, \emph{\HT-models}, is based on the logic of here-and-there
\citep{Heyting1930,Pearce1997} and is expressive enough to capture both
classical models and answer-sets.

Satisfaction of programs is obtained by treating rules as
classical implications. Table~\ref{table:lp:satisfaction} defines
satisfaction of literals $\olit$ and $\lpnot \olit$, a set of
literals $\slit$, a rule $\rl$ and a program $\prg$ in an
interpretation $\twib \subseteq \lang$. We say that $\twib$ is a
\emph{\C-model} (classical model) of a rule $\rl$ if $\twib \ent \rl$, and a \emph{\C-model} of a program $\prg$ if $\twib \ent \prg$.
The set of
all \C-models of a rule $\rl$ is denoted by $\modc{\rl}$ and for any program
$\prg$, $\modc{\prg} = \bigcap_{\rl \in \prg} \modc{\rl}$. A program $\prg$ is
\emph{consistent} if $\modc{\prg} \neq \emptyset$, and \emph{inconsistent} otherwise.

\begin{table}
	\caption{Satisfaction of literals, rules and programs}
	\begin{tabular}{l}
		\toprule
		$\twib \ent \olit$
			 \text{iff}
			 $\olit \in \twib$
			\\
		$\twib \ent \lpnot \olit$
			 \text{iff}
			 $\olit \notin \twib$
			\\
		$\twib \ent \slit$
			 \text{iff}
			 $\twib \ent \lit$ for all $\lit \in \slit$
			\\
		$\twib \ent \rl$
			 \text{iff}
			 $\exists \lit \in \brl : \twib \nent \lit$
				or $\exists \lit \in \hrl : \twib \ent \lit$
			\\
		$\twib \ent \prg$
			 \text{iff}
			 $\twib \ent \rl$ for all $\rl \in \prg$
			\\
		\bottomrule
	\end{tabular}
	\label{table:lp:satisfaction}
\end{table}

The stable and \HT-models are defined in terms of reducts. Given a program $\prg$ and an  
interpretation $\twib$, the \emph{reduct of $\prg$ w.r.t.\ $\twib$} is defined as 
\[
\prg^\twib = \Set{\an{\hrlp, \brlp}
			| 
			\rl \in \prg
			\land
			\twib \nent \an{\lpnot \hrln, \lpnot \brln}}.
\]

An interpretation $\twib$ is a \emph{stable model} of a program $\prg$ if
$\twib$ is a subset-minimal \C-model of $\prg^\twib$. The set of all stable
models of $\prg$ is denoted by $\modsm{\prg}$. A program $\prg$ is
\emph{coherent} if $\modsm{\prg} \neq \emptyset$, and \emph{incoherent} otherwise. 

\HT-models are semantic structures that can be seen as three-valued
interpretations. In particular, we call a pair of interpretations $\tri =
\an{\twia, \twib}$ such that $\twia \subseteq \twib$ a \emph{three-valued
interpretation}. Each atom $\atm$ is assigned one of three truth values in
$\tri$: $\tri(\atm) = \tr$ if $\atm \in \twia$; $\tri(\atm) = \un$ if $\atm
\in \twib \setminus \twia$; $\tri(\atm) = \fa$ if $\atm \in \lang \setminus
\twib$. The set of all three-valued interpretations is denoted by $\tris$. A
three-valued interpretation $\an{\twia, \twib}$ is an \emph{\HT-model} of a
rule $\rl$ if $\twib\ent\rl$ and $\twia\ent\rl^\twib$. The set of all \HT-models of a rule $\rl$ is denoted by
$\modHT{\rl}$ and for any program $\prg$, $\modHT{\prg} = \bigcap_{\rl \in
\prg} \modHT{\rl}$. Note that $\twib$ is a stable model of $\prg$ if and only
if $\an{\twib, \twib} \in \modHT{\prg}$ and for all $\twia \subsetneq \twib$,
$\an{\twia, \twib} \notin \modHT{\prg}$. Also, $\twib \in \modc{\prg}$ if and
only if $\an{\twib, \twib} \in \modHT{\prg}$. 
We write $\twiab \ent \prg$ if $\twiab \in \modHT{\prg}$. We say that \emph{$\prga$ is strongly
	equivalent to $\prgb$}, denoted by $\prga \eqHT \prgb$, if $\modHT{\prga} =
	\modHT{\prgb}$, and that \emph{$\prga$ strongly entails $\prgb$}, denoted by
	$\prga \entHT \prgb$, if $\modHT{\prga} \subseteq \modHT{\prgb}$.
	A rule $\rl$ is tautological if $\modHT{\rl}= \tris$. It follows from the results of \cite{Inoue2004} and \cite{Cabalar2007a} that a rule $\rl$ is tautological if $\brl\cap\hrl\neq\emptyset$ or $\brl^+\cap\brl^-\neq\emptyset$.

The class of programs defined above can be extended to allow for a second form
of negation, dubbed \emph{strong negation}. In a nutshell, $\lang$ is extended to 
also include, for each of its original atoms $\atm$, its (strong) negation $\lnot\atm$. Elements of this extended $\lang$ are dubbed \emph{objective literals} 
and we  use the following notation to refer to complementary objective literals: $\cmpl{\atm}= \lnot \atm$ and $\cmpl{\lnot \atm} = \atm$ for any atom $\atm$.
Each atom $\atm$ is interpreted separately of (though still consistently with) its strong negation $\lnot \atm$. Each interpretation naturally corresponds to a consistent subset of the extended set $\lang$. 
More formally, an \emph{(extended) interpretation} is a subset of $\lang$ that does
not contain both $\olit$ and $\cmpl{\olit}$ for any objective literal $\olit$. Note that this is in contrast with the definition of answer-set by \citeN{Gelfond1991}, which allows for certain programs to have their semantics be characterised by the so-called \emph{contradictory answer-set} $\lang$.

To simplify notation, first-order atoms with variables are often used in program rules. Such rules should be seen as a shortcut corresponding to the
set of rules obtained by replacing the variables with constants to form atoms in $\lang$, in all possible ways.

\noindent \textbf{Order Theory.}
Given a set $\sta$, a \emph{preorder over $\sta$} is a reflexive and
transitive binary relation over $\sta$; a \emph{strict preorder over $\sta$}
is an irreflexive and transitive binary relation over $\sta$; a \emph{partial
order over $\sta$} is a preorder over $\sta$ that is antisymmetric. Given a
preorder $\po{}{}$ over $\sta$, we denote by $\spo{}{}$ the strict preorder
induced by $\po{}{}$, i.e.\ $\ela \spo{}{} \elb$ if and only if $\ela \po{}{}
\elb$ and not $\elb \po{}{} \ela$. For any subset $\stb$ of $\sta$, the set of
\emph{minimal elements of $\stb$ w.r.t.\ $\po{}{}$} is 
$	\min(\stb, \po{}{})
	=
	\Set{
		\ela \in \stb
		|
		\lnot \exists \elb \in \stb : \elb \spo{}{} \ela
	}$.
A \emph{preorder assignment over $\sta$} is any
	function $\oas$ that assigns a preorder $\poeld$ over $\sta$ to each $\ela
	\in \sta$. A \emph{partial order assignment over $\sta$} is any preorder
	assignment $\oas$ over $\sta$ such that $\poeld$ is a partial order over
	$\sta$ for every $\ela \in \sta$. A \emph{total order assignment over $\sta$} is any preorder
	assignment $\oas$ over $\sta$ such that $\poeld$ is a total order over
	$\sta$ for every $\ela \in \sta$.

\section{Pre-History --- Belief Change in Classical Logic}
\label{sec:prehistory}
%!TEX root =  paper.tex

An \emph{update} is typically described as an operation that brings a
knowledge base \emph{up to date} when the \emph{world described by it changes}, whereas 
a \emph{revision} is typically described as an operation that deals with incorporating new \emph{better} knowledge about a \emph{world that did not change}
\citep{Keller1985,Winslett1990,Katsuno1991}. From a generic perspective, both forms of belief change operators --- update and revision --- were studied within the context of propositional logic.

Propositional belief change operators, either for update ($\uopb$) or for revision ($\ropb$), take two
propositional formulas, representing the original knowledge base and its
update, as arguments, and return a formula representing the updated knowledge
base. Any such operator $\copb\in\{\uopb,\ropb\}$
is inductively generalised to finite sequences $\seq{\frm_\lia}_{\lia < \lng}$
of propositional formulas as follows:
$\bigcopb \seq{\frm_0} = \frm_0$ and $\bigcopb \seq{\frm_\lia}_{\lia < \lng +
1} = (\bigcopb \seq{\frm_\lia}_{\lia < \lng}) \copb \frm_\lng$, $\lng>0$.
To further specify the desired properties of belief change operators, \citeN{Katsuno1989,Katsuno1991}
proposed two sets of postulates --- one for revision and one for update. Following the original order of presentation, we will first 
briefly review the postulates for revision, and then those for update. Even though
these postulate have been mostly absent from the literature in updates of answer-set programs,
more so during the era of syntax-based updates, they took a more prominent role during the era of semantic-based
updates, and will be revisited in Section \ref{sec:semanticUpdates}.

We start with the following
six postulates for a belief revision operator $\ropb$ and formulas $\frma$,
$\frmb$, $\frmu$, $\frmv$, proposed by \citeN{Katsuno1989}, which 
correspond to the AGM postulates for the case of propositional logic.

\begin{enumerate}[labelindent=18pt,labelwidth=10pt,leftmargin=!]
	\renewcommand{\theenumi}{\labbr{\arabic{enumi}}}
	\renewcommand{\labelenumi}{\theenumi\hfill}
	\setlength{\itemsep}{.1ex}
	\item[\labbr{1}]  $\frma \ropb \frmu \ent \frmu$.
		\label{pstl:br:1}

	\item[\labbr{2}] If $\mod{\frma\land\frmu} \neq \emptyset$, then 
		$\frma \ropb \frmu \eq \frma \land \frmu$.
		\label{pstl:br:2}

	\item[\labbr{3}] If $\mod{\frmu} \neq \emptyset$, then
		$\mod{\frma \ropb \frmu} \neq \emptyset$.
		\label{pstl:br:3}

	\item[\labbr{4}] If $\frma \eq \frmb$ and $\frmu \eq \frmv$, then $\frma \ropb
		\frmu \eq \frmb \ropb \frmv$.
		\label{pstl:br:4}

	\item[\labbr{5}] $(\frma \ropb \frmu) \land \frmv \ent \frma \ropb (\frmu \land
		\frmv)$.
		\label{pstl:br:5}

	\item[\labbr{6}] If $\mod{(\frma \ropb \frmu) \land \frmv} \neq \emptyset$,
		then $\frma \ropb (\frmu \land
		\frmv) \ent (\frma \ropb \frmu) \land \frmv$.
		\label{pstl:br:6}
\end{enumerate}

Most of these postulates can be given a simple intuitive reading. For
instance, \labbr{1} requires that information from the revision be retained in the
revised belief base. This is also frequently referred to as the
\emph{principle of primacy of new information} \citep{Dalal1988}. 
Postulate \labbr{2} requires that whenever the original formula and the revision are
jointly consistent, the result corresponds to their conjunction.
Postulate \labbr{3} requires that whenever the formula used for revision is satisfiable, 
then so should be the result of the revision. Postulate \labbr{4} encodes independence of syntax. 
Postulates
\labbr{5} and \labbr{6} require that revision should be accomplished with
minimal change.

The main idea behind these postulates is formally captured by the
notion of a belief revision operator characterised by an order assignment.

\begin{definition}
	[Belief Revision Operator Characterised by an Order Assignment]
	Let $\ropb$ be a belief revision operator and $\oas$ a preorder assignment
	over the set of all formulas. We say that $\ropb$ is \emph{characterised by $\oas$} if for
	all formulae $\frma$, $\frmu$,
	\[
		\mod{\frma \ropb \frmu}
		=
		\min \Br{ \mod{\frmu}, \pod{\frma} }
		.
	\]
\end{definition}

A set of natural conditions on the assigned orders is captured by the following notion of a \emph{faithful} order
assignment.

\begin{definition}
	[Faithful Preorder Assignment Over Formulas]
	A preorder assignment $\oas$ over the set of all formulas is \emph{faithful} if the following three conditions hold:
	\begin{itemize}[labelindent=2pt,labelwidth=6pt,leftmargin=!]
	\item $\text{If }\twia,\twib\in\mod{\frma}\text{, then }\twia\not\spod{\frma}\twib$.
	\item $\text{If }\twia\in\mod{\frma}\text{ and }\twib\not\in\mod{\frma}\text{, then }\twia\spod{\frma}\twib$.
	\item $\text{If }\mod{\frma}=\mod{\frmb}\text{, then }\spod{\frma}=\spod{\frmb}$.
	\end{itemize}
\end{definition}

The representation theorem of \citeN{Katsuno1989} states that operators
characterised by faithful total preorder assignments over the set of all formulas are exactly those that satisfy the
KM revision postulates.

\begin{theorem}
	[\citeN{Katsuno1989}]
	\label{thm:br:representation}
	Let $\ropb$ be a belief revision operator. Then the following conditions are
	equivalent:
	\begin{itemize}[labelindent=2pt,labelwidth=6pt,leftmargin=!]
		\item The operator $\ropb$ satisfies conditions \labbr{1} -- \labbr{6}.

		\item The operator $\ropb$ is characterised by a faithful total preorder
			assignment over the set of all formulas.
	\end{itemize}
\end{theorem}

%%%%%%%%%%%%%%%%%%%%%%%%%%%%%%%%%%%

Whereas according to \citeN{Katsuno1991}, \emph{revision} is used when we are
obtaining new information about a static world, \emph{updates} consists of bringing
a knowledge base up to date when the world described by it changes.
To characterise updates, \citeN{Katsuno1991} proposed the following
eight postulates for a belief update operator $\uopb$ and formulas $\frma$,
$\frmb$, $\frmu$, $\frmv$:

\begin{enumerate}[labelindent=18pt,labelwidth=10pt,leftmargin=!]
	\renewcommand{\theenumi}{\labbu{\arabic{enumi}}}
	\renewcommand{\labelenumi}{\theenumi\hfill}
	\setlength{\itemsep}{.1ex}
	\item[\labbu{1}]  $\frma \uopb \frmu \ent \frmu$.
		\label{pstl:bu:1}

	\item[\labbu{2}] If $\frma \ent \frmu$, then $\frma \uopb \frmu \eq \frma$.
		\label{pstl:bu:2}

	\item[\labbu{3}] If $\mod{\frma} \neq \emptyset$ and $\mod{\frmu} \neq \emptyset$, then
		$\mod{\frma \uopb \frmu} \neq \emptyset$.
		\label{pstl:bu:3}

	\item[\labbu{4}] If $\frma \eq \frmb$ and $\frmu \eq \frmv$, then $\frma \uopb
		\frmu \eq \frmb \uopb \frmv$.
		\label{pstl:bu:4}

	\item[\labbu{5}] $(\frma \uopb \frmu) \land \frmv \ent \frma \uopb (\frmu \land
		\frmv)$.
		\label{pstl:bu:5}

	\item[\labbu{6}] If $\frma \uopb \frmu \ent \frmv$ and $\frma \uopb \frmv \ent \frmu$,
		then $\frma \uopb \frmu \eq \frma \uopb \frmv$.
		\label{pstl:bu:6}

	\item[\labbu{7}] If $\frma$ is complete, then $(\frma \uopb \frmu) \land (\frma \uopb
		\frmv) \ent \frma \uopb (\frmu \lor \frmv)$.
		\label{pstl:bu:7}

	\item[\labbu{8}] $(\frma \lor \frmb) \uopb \frmu \eq (\frma \uopb \frmu) \lor (\frmb
		\uopb \frmu)$.
		\label{pstl:bu:8}
\end{enumerate}

Postulates \labbu{1}--\labbu{5} correspond to postulates \labbr{1}--\labbr{5}. However,
when $\frma$ is consistent, then \labbu{2} is weaker than \labbr{2}. 
The property expressed by \labbu{8} is at the heart of belief updates:
Alternative models of the original belief base $\frma$ or $\frmb$ are treated as possible
real states of the modelled world. Each of these models is updated
independently of the others to make it consistent with the update $\frmu$,
obtaining a new set of interpretations --- the models of the updated belief
base. Based on this view of updates, \citeN{Katsuno1991} proved an important
representation theorem that makes it possible to constructively characterise and evaluate every operator
$\uopb$ that satisfies postulates \labbu{1} -- \labbu{8}.
    The main idea, based on postulate \labbu{8}, is formally captured by the
notion of a belief update operator characterised by an order assignment.
\begin{definition}
	[Belief Update Operator Characterised by an Order Assignment]
	Let $\uopb$ be a belief update operator and $\oas$ a preorder assignment
	over $\twis$. We say that $\uopb$ is \emph{characterised by $\oas$} if for
	all formulae $\frma$, $\frmu$,
	\[
		\mod{\frma \uopb \frmu}
		=
		\bigcup_{\twi \in \modscr{\frma}}
		\min \Br{ \mod{\frmu}, \potwd }
		.
	\]
\end{definition}

A natural condition on the assigned orders is that every interpretation be the
closest to itself, captured by the following notion of a \emph{faithful} order
assignment.

\begin{definition}
	[Faithful Order Assignment over Interpretations]
	A preorder assignment $\oas$ over $\twis$ is \emph{faithful} if for every
	interpretation $\twia$ the following condition is satisfied:
	\[
		\text{For every } \twib \in \twis \text{ with } \twib \neq \twia
		\text{ it holds that } \twia \spotwd \twib
		.
	\]
\end{definition}

The representation theorem of \citeN{Katsuno1991} states that operators
characterised by faithful order assignments over $\twis$ are exactly those that satisfy the
KM update postulates.

\begin{theorem}
	[\citeN{Katsuno1991}]
	\label{thm:bu:representation}
	Let $\uopb$ be a belief update operator. Then the following conditions are
	equivalent:
	\begin{itemize}[labelindent=2pt,labelwidth=6pt,leftmargin=!]
		\item The operator $\uopb$ satisfies conditions \labbu{1} -- \labbu{8}.

		\item The operator $\uopb$ is characterised by a faithful preorder
			assignment over $\twis$.

		\item The operator $\uopb$ is characterised by a faithful partial order
			assignment over $\twis$.
	\end{itemize}
\end{theorem}
  
Katsuno and Mendelzon's result provides a framework for belief update
operators, each specified on the semantic level by strict preorders assigned
to each propositional interpretation. The most influential instance of this
framework is the \emph{possible models approach} by \citeN{Winslett1988}, based
on minimising the set of atoms whose truth value changes when an
interpretation is updated. Formally, for all interpretations $\twia$, $\twib$
and $\twic$, the strict preorder $\spotww$ is defined as follows:
$\twib \spotww \twic$ if and only if $(\twib \div \twia) \subsetneq (\twic
\div \twia)$,
where $\div$ denotes set-theoretic symmetric difference. The operator
$\uopbw$ by Winslett, unique up to equivalence of its inputs and
output, thus satisfies the following equation:
\[
	\mod{\frma \uopbw \frmu}
	=\bigcup_{\twi \in \modscr{\frma}}
	\Set{
		\twib \in \mod{\frmu}
		|
		\lnot \exists \twic \in \mod{\frmu} :
			(\twic \div \twia) \subsetneq (\twib \div \twia)
	}.
\]
Note that it follows from Theorem~\ref{thm:bu:representation} that $\uopbw$
satisfies postulates \labbu{1} -- \labbu{8}.

\section{The First Era --- Model Updates}
\label{sec:modelUpdates}
%!TEX root =  paper.tex

The first authors to address the issue of updates and logic programs using the stable model semantics were
\citeN{DBLP:conf/jelia/MarekT94,Marek1998}, although not as a means to update logic program. Instead, they used a rule based
language, and a semantics similar to the stable models semantics, to specify the updates of a database. 

The underlying idea was that the rules would specify constraints that had to be satisfied by a database. Then, given an initial database and a set of rules, \citeN{DBLP:conf/jelia/MarekT94,Marek1998} defined a semantics that assigns a set of databases that are the justified result of the update. 

In a nutshell, a database $DB^{\prime}$ is considered to be a justified update of a database $DB$ by a program $U$ if $DB^{\prime}$, viewed as an interpretation, is a model of $U$, and no other database $DB^{\prime\prime}$, that is also a model of $U$, is closer to $DB$ than $DB^{\prime}$. 

Procedurally, this notion of justified update corresponds to following the common-sense law of inertia, whereby \emph{only} those elements that \emph{need} to be changed due to the 
update specification are \emph{actually changed}, the remaining staying the same, in line with the ideas proposed by \citeN{Winslett1988}.

Representing databases as interpretations, update programs as logic programs, and with $\div$ denoting set-theoretic symmetric difference, as before, the set of interpretations resulting from updating  
 $\twia$ by $\prgu$ is given by
 \[
 \Set{
		\twib \in \modc{\prgu}
		|
		\lnot\exists \twic \in \modc{\prgu} :
			(\twic \div \twia) \subsetneq (\twib \div \twia)
	}.
 \] 
\begin{example}
Consider an initial interpretation $\twia=\Set{\prest{cold},\prest{sun}}$
and suppose we want to update it according to the following program $\prgu$:
\begin{align*}
		\prest{rain}.&
		&\prest{clouds} \lpif \prest{rain}.&
		&\lpnot\prest{sun} \lpif \prest{clouds}.	
\end{align*}

The only justified update is the interpretation $\twib=\Set{\prest{cold}, \prest{clouds}, \prest{rain}}$. Declaratively, $\prga$ has two models $\Set{\prest{cold}, \prest{clouds}, \prest{rain}}$ and $\Set{\prest{clouds}, \prest{rain}}$, where the former is obviously closer to $\twia$ than the latter. Procedurally, $\prest{rain}$ is true in $\twib$ because of the fact in $\prgu$, $\prest{clouds}$ is true in $\twib$ because of the second rule in $\prgu$, together with the fact that  $\prest{rain}$ is now true, $\prest{sun}$ is false because of the third rule in the program, while $\prest{cold}$ is true in $\twib$, \emph{by inertia}, because it was true in $\twia$ and there is nothing in $\prgu$ forcing it to become false.
\end{example}

Subsequently, \citeN{DBLP:conf/lpnmr/PrzymusinskiT95,DBLP:journals/jlp/PrzymusinskiT97} showed how the framework proposed by \citeN{DBLP:conf/jelia/MarekT94,Marek1998} could be captured by logic programming under the stable models semantics, by encoding the initial database as a set of facts, the rules proposed by \citeN{DBLP:conf/jelia/MarekT94,Marek1998} as rules of logic programming, and by adding additional rules encoding the common-sense law of inertia, so that the stable models of the resulting logic program would correspond to the justified updates of the initial database.

It was \citeN{Alferes1996} who first proposed to use a logic program to update another logic program, under the stable model semantics. Following the possible models approach proposed by \citeN{Winslett1988}, a knowledge base $DB^{\prime}$ is considered to be the
update of a knowledge base $DB $ by $U$ if each model of $DB^{\prime}$ is an update of a model of $DB $ by $U$. 

According to this approach, dubbed the \emph{model update approach}, the problem of finding
an update of a logic program $\prga$ is reduced to the problem of individually
finding updates of each of its stable models $\twia$. Each stable model would be updated following the ideas proposed by \citeN{DBLP:conf/jelia/MarekT94,Marek1998}, i.e., following the common-sense law of inertia, or minimal change. 

Just like \citeN{DBLP:conf/lpnmr/PrzymusinskiT95,DBLP:journals/jlp/PrzymusinskiT97}, \citeN{Alferes1996} proposed an encoding of the problem of updating a program $\prga$ by a program $\prgu$ into logic programming, producing another logic program, written in an extended language, whose stable models correspond to the updates of each of the stable models of $\prga$ by $\prgu$.

According to \citeN{Alferes1996}, the update of a program $\prga$ by a program $\prgu$ is characterised by the following set of interpretations:
 \[
	\bigcup_{\twi \in \modsm{\prga}}
 \Set{
		\twib \in \modc{\prgu}
		|
		\lnot\exists \twic \in \modc{\prgu} :
			(\twic \div \twia) \subsetneq (\twib \div \twia)
	}.
 \] 

As it turns out, when we take a closer look at this semantics, we soon realise that things do not behave exactly as one might expect.
\begin{example}
Consider the same example from the introduction, where an agent had its beliefs represented by the program $\prga$:
	\begin{align*}
		\prest{goHome} \lpif \lpnot \prest{money}.&
		&\prest{goRestaurant} \lpif \prest{money}.&
		&\prest{money}.
	\end{align*}
which was then the subject of an update by $\prgu$ with the following two rules:
	\begin{align*}
		\lpnot \prest{money} &\lpif \prest{robbed}.
		& \prest{robbed} &.
	\end{align*}
	The only stable model of $\prga$ is $\twia = \Set{\prest{money},
	\prest{goRestaurant}}$.
	Program $\prgu$ has the following four models:
	$\twic_1=\Set{\prest{robbed}}$, $\twic_2=\Set{\prest{robbed}, \prest{goRestaurant}}$,
	$\twic_3=\Set{\prest{robbed}, \prest{goHome}}$, and $\twic_4=\Set{\prest{robbed}, \prest{goRestaurant},\prest{goHome}}$. 
Of these four models, $\twic_2$ is the only one that is closest to $\twia$, hence it is the only one that characterises the update of $\prga$ by $\prgu$ according to
\citeN{Alferes1996}.
	
Procedurally, if we update $\prga$ by $\prgu$ following the fundamental ideas behind
	the possible models approach and the common sense law of inertia, the result must be characterised by the stable models of
	$\prga$ after they are minimally changed to become consistent with $\prgu$.
	And in order to make $\twia$ consistent with the rules in $\prgu$, one needs
	to modify the truth value of two atoms, $\prest{robbed}$ and
	$\prest{money}$, arriving at the interpretation $\twic_2 =
	\Set{\prest{robbed}, \prest{goRestaurant}}$. So, after the update, the agent
	has no money but still plans to go to a restaurant, different from the intuitively correct result 
	where, after being robbed, the agent would not go to the restaurant, and instead go home, because he has no money.
\end{example}

The undesirable behaviour illustrated by the previous example was first observed by \citeN{Leite1997}, leading to the beginning of the second era.

\section{The Second Era --- Syntax-Based Updates}
\label{sec:syntacticUpdates}
%!TEX root =  paper.tex

%%% Here starts the part from the survey

The reason for the problem illustrated by the example at the end of the previous section 
is that modifications on the level of individual stable models, akin to model-based belief update operators, are unable
to capture the essential relationships between literals encoded in rules. This
was first argued by \citeN{Leite1997}, who took a closer look
at Newton's first law, also known as the \emph{law of inertia}, which states that
\emph{``every body remains at rest or moves with constant velocity in a
straight line, unless it is compelled to change that state by an unbalanced
force acting upon it''} \citep{Principia}.\footnote{The original text of Newton is as follows: \emph{``Corpus omne perseverare in statu suo
         quiescendi vel movendi uniformiter in directum, nisi quatenus illud a
         viribus impressis cogitur statum suum mutare.''}.} In their discussion, Leite and Pereira pointed out that the 
common-sense interpretation of this law as \emph{``things keep as they are unless some kind of
force is applied to them''} is true, but does not exhaust its meaning. It is the result of all applied forces that governs the outcome. Take
a body to which several forces are applied, and which is in a state of
equilibrium due to those forces cancelling out. Later, one of those forces is
removed and the body starts to move. The same kind of behaviour presents itself when updating programs. Before
obtaining the truth value, by inertia, of those elements not directly affected
by the update program, one should verify whether the truth of such elements is
not indirectly affected by the updating of other elements or, in other words, whether there
is still some rule that supports such truth. 

To rectify this problem, a number of approaches were proposed. Despite being 
based on fundamentally different
principles and methods when compared to their model update counterparts, they 
all take into account the syntactic rule-based form of the programs involved.
 
This section provides an
overview of existing rule update semantics, pointing at some of the technical
as well as semantic differences between them, often relying on examples
to show how these semantics are interrelated.

Rule update semantics typically deal only with ground non-disjunctive rules
and some do not allow for default negation in their heads. While some of them
follow the belief update tradition and construct an updated program given the
original program and its update, others only assign a set of stable models to
a pair or sequence of programs where each represents an update of
the preceding ones. In order to compare these semantics, we adopt the latter,
less restrictive point of view. The ``input'' of a rule update semantics is
thus defined as follows:

\begin{definition}
	[Dynamic Logic Program]
	A \emph{dynamic logic program} (DLP) is a finite sequence of ground
	non-disjunctive logic programs.
	Given a DLP $\dprg$, we denote by $\all{\dprg}$ the set of all rules
	belonging to the programs in $\dprg$. We say that $\dprg$ is \emph{acyclic}
	if $\all{\dprg}$ is acyclic.
\end{definition}

In order to avoid issues with rules that are repeated in multiple components
of a DLP, we assume throughout this section that every rule is uniquely
identified in all set-theoretic operations. This could be formalised by
assigning a unique name to each rule and performing operations on names
instead of on the rules themselves. 

The set of stable models assigned to dynamic logic programs under a particular
update semantics will be denoted as follows:

\begin{definition}
	[Rule Update Semantics]
	A \emph{rule update semantics \S{}} is characterised by a (partial) function
	$\mods{\cdot}$ that assigns a set $\mods{\dprg}$ of interpretations to a
	dynamic logic program $\dprg$. We call each member of $\mods{\dprg}$ an
	\emph{\S-model of $\dprg$}.
\end{definition}

Whenever an approach is defined through a rule update operator, such an operator is understood as a function that assigns a program to
each pair of programs. A rule update operator $\uopr$ is extended to DLPs as
follows: $\biguopr \an{\prg_0} = \prg_0$; $\biguopr \an{\prg_\lia}_{\lia <
\lng+1} = (\biguopr \an{\prg_\lia}_{\lia < \lng}) \uopr \prg_\lng$, $\lng>0$. Note that
such an operator naturally induces a rule update semantics $\textsf{\S{}}_{\uopr}$:
given a DLP $\dprg$, the $\textsf{\S{}}_{\uopr}$-models of $\dprg$ are the
stable models of $\biguopr \dprg$. In the rest of this paper, we exercise a
slight abuse of notation by referring to the operators and their associated
update semantics interchangeably.

We first discuss a major group of semantics based on the \emph{causal
rejection principle}
\citep{Leite1997,Buccafurri1999,Alferes2000,Eiter2002,Alferes2005,Osorio2007}, followed by semantics based on
\emph{preferences} \citep{Zhang2006,Delgrande2007}. We then proceed to discuss
semantics that bear some characteristics of revision rather than update
\citep{Sakama2003,Osorio2007a,Delgrande2010} and touch upon approaches that
manipulate dependencies on default assumptions induced by rules
\citep{Sefranek2011,Krumpelmann2010,Krumpelmann2012}. Towards the end of this section, we formulate some fundamental
properties of rule update semantics.

The use and effect of integrity constraints within the semantics presented in this section has not received significant attention in the literature. Whereas most semantics are only defined for programs without integrity constraints, often pointing to the fact that one can replace the integrity constraint $\lpif\brl.$ with the rule $a_\rl\lpif\lpnot a_\rl,\brl.$, where $a_\rl$ is a new atom, with the same effect, other semantics are defined for programs with integrity constraints, but without any specific provisions regarding their role. Since all these semantics that were defined for programs with integrity constraints are preserved under the above transformation that eliminates them, in this section we will restrict the definition of semantics to only consider DLPs without integrity constraints.

\subsection{Causal Rejection-Based Semantics}

\label{sec:ru:causal rejection}

The \emph{causal rejection principle} \citep{Leite1997} forms the basis of a
number of rule update semantics. Informally, it can be stated as follows:
\begin{center}
	A rule should be \emph{rejected} when it is directly contradicted by a more
	recent rule.
\end{center}
The common understanding of this principle has been to consider a direct contradiction between rules
to mean a conflict between the heads of rules, i.e., that the head of the rejecting rule is the
negation of the head of the rejected one. 

In the first proposals based on the causal rejection principle, only conflicts between \emph{objective} literals in rule
heads were considered, and default negation in rule heads was not allowed
\citep{Leite1997,Eiter2002}. Later, it was found that this approach has
certain limitations, namely that some belief states, represented by stable
models, become unreachable \citep{Alferes2000,Leite2003}. 
For example, no
update of the program $\prga = \set{\atma.}$ leads to a stable model where
neither $\atma$ nor $\lnot \atma$ is true. 
Default negation in rule heads was
thus used to regain reachability of such states. For instance, the update
$\prgu = \Set{\lpnot \atma., \lpnot \lnot \atma.}$ forces $\atma$ to be
unknown, regardless of its previous state. 
Hence, strong negation is used to express that an atom \emph{becomes explicitly
false}, while default negation allows for more fine-grained control: the atom only
\emph{ceases to be true}, but its truth value may not be unknown. The latter
also makes it possible to move between any pair of epistemic states by means
of updates, as illustrated by the following example from the book by \cite{Leite2003}:

\begin{example}
	[Railway crossing]
	\label{ex:railway crossing}
	Suppose that we use the following logic program to choose an action at a
	railway crossing:
	\begin{align*}
		\prest{cross} &\lpif \lnot \prest{train}.
		&
		\prest{wait} &\lpif \prest{train}.
		&
		\prest{listen} &\lpif \lpnot \prest{train}, \lpnot \lnot \prest{train}.
	\end{align*}
	The intuitive meaning of these rules is as follows: one should \prest{cross}
	if there is evidence that no train is approaching; \prest{wait} if there is
	evidence that a train is approaching; \prest{listen} if there is no such
	evidence.

	Consider a situation where a train is approaching, represented by the fact
	$(\prest{train}.)$. After this train has passed by, we want to update our
	knowledge to an epistemic state where we lack evidence with regard to the
	approach of a train. If this was accomplished by updating with the fact
	$(\lnot \prest{train}.)$, we would cross the tracks at the subsequent state,
	risking being killed by another train that was approaching. Therefore, we
	need to express an update stating that all past evidence for an atom is to
	be removed. The proposal was to accomplish this by allowing default negation in heads
	of rules. In this scenario, the intended update could be expressed by the fact
	$(\lpnot \prest{train}.)$.
\end{example}

In the following, we thus present
the semantics from \citeN{Leite1997} and \citeN{Eiter2002} in generalised forms that allow
default negation in rule heads, but coincide with their original definitions on programs without such feature \citep{Leite2003}.

The notion of conflicting rules plays an important role in all the semantics based on the causal rejection principle. When generalised logic programs are used, a conflict between rules occurs when the head literal $\olit$ of one rule is the
default or strong negation of the head literal of the other rule, $\lpnot\olit$ or $\cmpl{\olit}$ respectively. Following the proposal of \citeN{Leite2003}, we consider the conflicts between a rule with an objective literal $\olit$ in its head and a rule with the default negation of the same literal $\lpnot
\olit$ in its head as primary conflicts, while conflicts between rules with complementary objective literals in their heads, $\olit$ and $\cmpl{\olit}$, are handled by \emph{expanding} DLPs. Expansion of a DLP corresponds to the following operation: whenever
a DLP contains a rule with an objective literal $\olit$ in its head, its
expansion also contains a rule with the same body and the literal $\lpnot
\cmpl{\olit}$ in its head. Formally:

\begin{definition}
	[Expanded Version of a DLP]
	Let $\dprg = \seq{\prg_\lia}_{\lia < \lng}$ be a DLP. The \emph{expanded
	version of $\dprg$} is the DLP $\expv{\dprg} = \seq{\expv{\prg_\lia}}_{\lia
	< \lng}$ where for every $\lia < \lng$,
\[		\expv{\prg_\lia} = \prg_\lia \cup \{
			\lpnot \cmpl{\olit} \lpif \brl.
			|
			\rl \in \prg_\lia
			\land \hrl = \set{\olit}
			\land \olit \in \olits
		\}.
\]
\end{definition}

The additional rules in the expanded version capture the coherence principle:
when an objective literal $\olit$ is derived, its complement $\cmpl{\olit}$
cannot be concurrently true and thus $\lpnot \cmpl{\olit}$ must be true. In
this way, every conflict between complementary objective literals directly
translates into a conflict between an objective literal and its default
negation. By ensuring that we always use expanded versions of DLPs, we can adopt the following definition of a conflict between a pair of rules. 
\begin{definition}[Conflicting rules]
\label{def:conflicting_rules}
We say that rules $\rla$, $\rlb$ \emph{are in conflict}, denoted by
$\rla \confl{}{} \rlb$, if and only if
\[
	\hrla \neq \emptyset \enspace\text{ and }\enspace
		\hrla = \lpnot \hrlb
	.
\]
\end{definition}

\subsubsection{The \JU-semantics and the \AS-semantics}
The historically first rule update semantics in answer-set programming is the \emph{justified update
semantics}, or \emph{\JU-semantics} for short \citep{Leite1997}, with the idea
to define a set of \emph{rejected rules}, which depends on a stable model
candidate, and then verify that the candidate is indeed a stable model of the
remaining rules.

\begin{definition}
	[\JU-Semantics \citep{Leite1997}]
	\label{def:ru:ju}
	Let $\dprg = \seq{\prg_\lia}_{\lia < \lng}$ be a DLP and $\twib$ an 
	interpretation. We define the set $\rejju{\dprg, \twib}$ of rejected rules
	as
\[
		\rejju{\dprg, \twib} = \{
			\rla \in \prg_\lia
			|
			\exists \lib \, \exists \rlb
			:
			\lia < \lib < \lng 
			\land \rlb \in \prg_\lib 
			\land \rla \confl{}{} \rlb
			\land \twib \ent \brlb
		\}
		.
\]
	The set $\modju{\dprg}$ of \emph{\JU-models of a DLP $\dprg$} consists of
	all stable models $\twib$ of the program
	\[
		\all{\expv{\dprg}} \setminus \rejju{\expv{\dprg}, \twib}
		.
	\]
\end{definition}

Under the \JU-semantics, a rule $\rla$ is rejected if and only if a more
recent rule $\rlb$ is in conflict with $\rla$ and the body of $\rlb$ is
satisfied in the stable model candidate $\twib$. Note that the latter
condition is essential --- without it, rules might get rejected simply because
a more recent rule $\rlb$ has a conflicting head, without a guarantee that
$\rlb$ will actually be activated.

Before we illustrate the \JU-semantics with an example, we first present
a related semantics, which prevents rejected rules from rejecting other rules. It is dubbed the \emph{update answer-set semantics}, or \emph{\AS-semantics} for short
\citep{Eiter2002}: 

\begin{definition}
	[\AS-Semantics \citep{Eiter2002}]
	Let $\dprg = \seq{\prg_\lia}_{\lia < \lng}$ be a DLP and $\twib$ an
	interpretation. We define the set of rejected rules $\rejas{\dprg, \twib}$
	as\footnote{%
			Note that although the definition is recursive, the defined set is
			unique. This is because we assume that every rule is uniquely identified
			and to determine whether a rule from $\prg_\lia$ is rejected, the
			recursion only refers to rejected rules from programs $\prg_\lib$ with
			$\lib$ strictly greater than $\lia$. One can thus first find the
			rejected rules in $\prg_{\lng - 1}$ (always $\emptyset$ by the definition),
			then those in $\prg_{\lng - 2}$ and so on until $\prg_0$.
		}
\[
		\rejas{\dprg, \twib} = \{
			\rla \in \prg_\lia
			|
			\exists \lib \, \exists \rlb
			:
			\lia < \lib < \lng
			\land \rlb \in \prg_\lib \setminus \rejas{\dprg, \twib}
			\land \rla \confl{}{} \rlb
			\land \twib \ent \brlb
		\}.
\]
	The set $\modas{\dprg}$ of \emph{\AS-models of a DLP $\dprg$} consists of
	all stable models $\twib$ of the program
	\[
		\all{\expv{\dprg}} \setminus \rejas{\expv{\dprg}, \twib}
		.
	\]
\end{definition}

The definitions of the \JU- and \AS-semantics are fairly straightforward and
reflect intuitions about rule updates better than an approach based on a
belief update construction, such as the one by \citeN{Alferes1996}. As an illustration,
let us look at the result of these semantics when applied to the example in the introduction:

\begin{example}
	Consider again the program $\prga$ from the example in the introduction which contains the rules
	\begin{align*}
		\prest{goHome} \lpif \lpnot \prest{money}.&
		&\prest{goRestaurant} \lpif \prest{money}.&
		&\prest{money}.
	\end{align*}
		and its update $\prgu$ with the rules
	\begin{align*}
		\lpnot \prest{money} &\lpif \prest{robbed}.
		& \prest{robbed} &.
	\end{align*}
	Following the discussion in the introduction,
	the expected stable model of the DLP $\seq{\prga, \prgu}$ is $\twib =
	\set{\prest{robbed}, \prest{goHome}}$. Also, $\rejju{\expv{\seq{\prga,
	\prgu}}, \twib} = \rejas{\expv{\seq{\prga, \prgu}}, \twib} =
	\Set{\prest{money}.}$, and $\twib$ is indeed a stable model of the remaining
	rules in $\all{\expv{\seq{\prga, \prgu}}}$. Furthermore, $\twib$ is the only
	interpretation with these properties, so
	\[
		\modju{\seq{\prga, \prgu}}
		= \modas{\seq{\prga, \prgu}}
		= \set{\twib}
		.
	\]
\end{example}

\begin{example}
	To illustrate the expansion mechanism and its interplay with the rejection mechanism, consider again the program $\prga$ from the example in the introduction which contains the rules
	\begin{align*}
		\prest{goHome} \lpif \lpnot \prest{money}.&
		&\prest{goRestaurant} \lpif \prest{money}.&
		&\prest{money}.
	\end{align*}
		but now its update $\prgu$ is modified by replacing $\lpnot \prest{money}$ with $\lnot\prest{money}$:
	\begin{align*}
		\lnot \prest{money} &\lpif \prest{robbed}.
		& \prest{robbed} &.
	\end{align*}
	The expanded version of both programs are:
	\begin{align*}
\expv{\prga}&		&\prest{goHome} &\lpif \lpnot \prest{money}.
		&\prest{goRestaurant} &\lpif \prest{money}.&
		&\prest{money}.\\
		&		&\lpnot\lnot\prest{goHome} &\lpif \lpnot \prest{money}.
		&\lpnot\lnot\prest{goRestaurant} &\lpif \prest{money}.&
		&\lpnot\lnot\prest{money}.\\
\expv{\prgu}&		&\lnot \prest{money} &\lpif \prest{robbed}.
		 &\prest{robbed}.& &\\
 		 &		&\lpnot \prest{money} &\lpif \prest{robbed}.
		 &\lpnot\lnot\prest{robbed}.& &
	\end{align*}

	The expected stable model of the DLP $\seq{\prga, \prgu}$ is $\twib =
	\set{\prest{robbed},\lnot\prest{money}, \prest{goHome}}$. Also, $\rejju{\expv{\seq{\prga,
	\prgu}}, \twib} = \rejas{\expv{\seq{\prga, \prgu}}, \twib} =
	\Set{\prest{money}.,\lpnot\lnot\prest{money}.}$, and $\twib$ is indeed a stable model of the remaining
	rules in $\all{\expv{\seq{\prga, \prgu}}}$. Furthermore, $\twib$ is the only
	interpretation with these properties, so
	\[
		\modju{\seq{\prga, \prgu}}
		= \modas{\seq{\prga, \prgu}}
		= \set{\twib}
		.
	\]
\end{example}

Nevertheless, problematic examples which are not handled correctly by these
semantics have also been identified \citep{Leite2003}. Many of them involve
tautological updates, the intuition being that a tautological rule (i.e.,\ a
rule whose head literal also belongs to its body) cannot indicate a change in
the modelled world because it is always true. It thus follows that a
tautological update should not affect the stable models of the original
program.  Interestingly, immunity to tautological updates is a desirable
property of belief updates in classical logic, being a direct consequence of postulate \labbu{2}.
The following example illustrates a misbehaviour of the \AS-semantics
\begin{example}
Consider the DLP
$\dprg_1 = \seq{\set{\atma.}, \set{\lnot \atma.}, \set{\atma \lpif \atma.}}$. Under the \AS-semantics, we obtain
$ \modas{\dprg_1}
		= \set{\set{\lnot \atma}, \set{\atma}}$, 
where the expected result is $\set{\set{\lnot \atma}}$. 
\end{example}

In
the previous example, the \JU-semantics provides an adequate solution: since it allows rejected
rules to reject, the initial rule is always rejected and $\modju{\dprg_1} =
\set{\set{\lnot \atm}}$ as expected. Unfortunately, there are also numerous
DLPs to which the \JU-semantics assigns unwanted models, as illustrated b the following example.
\begin{example}
\label{eq:ru:problem with ju}
Consider the DLP
$\dprg_2 = \seq{\set{\atma.}, \set{\lpnot \atma \lpif \lpnot \atma.}}$. Under the \JU- and \AS-semantics, we obtain
$\modju{\dprg_2}
		= \modas{\dprg_2}
		= \set{\emptyset, \set{\atma}}$, where the expected result is $\set{\set{\atma}}$.
\end{example}
The unwanted stable model $\emptyset$ arises because rejecting $\atma.$ causes 
the default assumption $\lpnot \atma$ to be ``reinstated'', i.e., $\atma$ to be assumed false \emph{by default} --- despite $\atma$ being initially asserted as a
fact. 

\subsubsection{The \DS-semantics}

The problem illustrated with Example \ref{eq:ru:problem with ju} is addressed in the \emph{dynamic stable model semantics}, or \emph{\DS-semantics} for
short \citep{Alferes2000}, by constraining the set of atoms that can be assumed false by default.

\begin{definition}
	[\DS-Semantics \citep{Alferes2000}]
	Let $\dprg = \seq{\prg_\lia}_{\lia < \lng}$ be a DLP and $\twib$ an
	interpretation. The set of rejected rules $\rejds{\dprg, \twib}$ is
	identical to the set $\rejju{\dprg, \twib}$ and we define the set of default
	assumptions $\defs{\dprg, \twib}$ as
\[
		\defs{\dprg, \twib} = \{
			\lpnot \olit
			|
			\olit \in \olits
			\land \lnot \exists \rl \in \all{\dprg} :
				\hrl = \set{\olit} \land \twib \ent \brl
		\}.
\]
	The set $\modds{\dprg}$ of \emph{\DS-models of a DLP $\dprg$} consists of
	all interpretations $\twib$ such that
	\[
		\twib' = \least{
			[ \all{\expv{\dprg}} \setminus \rejds{\expv{\dprg}, \twib} ]
			\cup \defs{\expv{\dprg}, \twib}
		}
		,
	\]
	where $\twib' = \twib \cup \lpnot\,(\olits \setminus \twib)$ and
	$\least{\cdot}$ denotes the least model of the argument program with all
	literals treated as atoms.
\end{definition}

Note that it follows from the definition of a (regular) stable model that
$\twib$ is a \JU-stable model of a DLP $\dprg$ if and only if
\[
	\twib' = \least{
		[ \all{\expv{\dprg}} \setminus \rejds{\expv{\dprg}, \twib} ]
		\cup \lpnot\,(\olits \setminus \twib)
	}
	.
\]
Hence, the difference between the \JU- and \DS-semantics is only in the set of
default assumptions that can be adopted to construct the model. In particular,
if a rule that derives an objective literal $\olit$ is present in
$\all{\dprg}$, then $\lpnot \olit$ is not among the default assumptions in the
\DS-semantics although it could be used as a default assumption in the \JU-semantics. In other words, according to the \DS-semantics, once some objective literal $\olit$ can be derived
by some (\emph{older}) rule, we can no longer assume $\lpnot \olit$ by default.
The \DS-semantics thus resolves problems with examples such as the previous one encoded by $\dprg_2$, i.e.,\ it holds that $\modds{\dprg_2} =
\set{\set{\atma}}$. But even the \DS-semantics exhibits problematic behaviour
when tautological updates are involved, as illustrated by the following example.
\begin{example}
Consider the DLP
$\dprg_3 = \seq{\set{\atma., \lnot \atma.}, \set{\atma \lpif \atma.}}$. Under the \DS-semantics, just as the case with the \JU- and the \AS-semantics, we obtain 
$ \modju{\dprg_3}
		= \modas{\dprg_3}
		= \modds{\dprg_3}
		= \set{\set{\atma}}$, where the expected result is $\set{}$.
\end{example}
The expected result here is that no stable model should be assigned to
$\dprg_3$ because initially it has none and the tautological update should not
change anything about that situation. 
Whereas one might wonder why not simply achieve immunity to tautologies by preprocessing programs and removing them, it is important to note that 
the problem runs deeper. Immunity to tautologies is a simple, easy to understand manifestation of a deeper problem concerning updates that should be considered tautological, even though 
there are no tautological rules. These tautological updates are characterised by the existence of cycles involving more than one rule, which could not be dealt with by simply removing tautologies, not even sets of cyclic rules within a single program. 

\subsubsection{The \RD-semantics}
The trouble with tautological and some other
types of irrelevant updates has been discussed and finally resolved by \citeN{Alferes2005}
who defined the so-called \emph{refined extension principle} --- a principle encoding the
desirable immunity to tautological, cyclic and other irrelevant updates --- 
as well as a rule update
semantics satisfying the principle. The definition of this semantics is very similar to the \DS-semantics, the only difference being
that in the set of rejected rules, $\lia \leq \lib$ is required instead of
$\lia < \lib$, which seems to be a technical trick with little correspondence to any intuition.
The semantics is called the \emph{refined dynamic stable
model semantics}, or \emph{\RD-semantics} for short:

\begin{definition}
	[\RD-Semantics \citep{Alferes2005}]
	Let $\dprg = \seq{\prg_\lia}_{\lia < \lng}$ be a DLP and $\twib$ an
	interpretation. We define the set of rejected rules $\rejrd{\dprg, \twib}$
	as
\[
		\rejrd{\dprg, \twib} = \{
			\rla \in \prg_\lia
			|
			\exists \lib \, \exists \rlb
			:
			\lia \leq \lib < \lng
			\land \rlb \in \prg_\lib
			\land \rla \confl{}{} \rlb
			\land \twib \ent \brlb
		\}.
\]
	The set $\modrd{\dprg}$ of \emph{\RD-models of a DLP $\dprg$} consists of
	all interpretations $\twib$ such that
	\[
		\twib' = \least{
			[ \all{\expv{\dprg}} \setminus \rejrd{\expv{\dprg}, \twib} ]
			\cup \defs{\expv{\dprg}, \twib}
		}
		,
	\]
	where $\twib'$ and $\least{\cdot}$ are as before.
\end{definition}

Due to satisfying the refined extension principle, the \RD-semantics is
completely immune to tautological updates. For instance, in case of the previous example
encoded by $\dprg_3$ we obtain $\modrd{\dprg_3} = \emptyset$. As we
shall see, a vast majority of rule update semantics, even those that have been
developed much later and are not based on causal rejection, are not immune to
tautological updates.

\cite{Banti2005} present an alternative, equivalent, characterisation for the 
\RD-semantics, based on level mappings, which is perhaps better in helping
understand the difference w.r.t. previous semantics than the trick used
in the previous definition of the set of rejected rules where $\lia \leq \lib$ is required instead of
$\lia < \lib$. 

\begin{theorem}
	[\RD-Semantics, Alternative Characterisation {\citep{Banti2005}}]
		Let $\dprg = \seq{\prg_\lia}_{\lia < \lng}$ be a DLP and $\twib$ an
	interpretation. Given a level mapping $\lm$, let the set $\rejlm{\dprg, \twib}$ of rejected rules be defined as follows:
\[
		\rejlm{\dprg, \twib} = \{
			\rla \in \prg_\lia
			|
			\exists \lib \, \exists \rlb
			:
			\lia < \lib < \lng
			\land \rlb \in \prg_\lib
			\land \rla \confl{}{} \rlb
			\land \twib \ent \brlb
			\land \lm(\hrlb) > \lm(\brlb)
		\}.
\]
	Then, $\twib\in\modrd{\dprg}$ iff there exists a level mapping $\lm$ such that:
	\begin{enumerate}[labelindent=2pt,labelwidth=6pt,leftmargin=!]
	\item $\twib$ is a \C-model of $\all{\expv{\dprg}} \setminus \rejlm{\expv{\dprg}, \twib}$, and
	\item $\forall\atma\in\twib,\exists\rlb\in\all{\expv{\dprg}} \setminus \rejlm{\expv{\dprg}, \twib}$ such that $\hrlb=\atma\land, \lm(\hrlb) > \lm(\brlb)\land\twib\models\brlb$.
	\end{enumerate}
	\end{theorem}

This characterisation borrows from the work of \citeN{DBLP:journals/tplp/HitzlerW05} on uniform characterisations of different semantics for logic programs in terms of level mappings. In particular, this characterisation is based on the notion of 
well-supported models \citep{DBLP:journals/mlcs/Fages94}, an alternative view of stable models that characterises them as \C-models with the additional requirement that there exists some level mapping such all atoms in the model are supported by some rule whose head is that atom and the level of the that atom is greater than the level of the rule's body. When extended to the case of DLPs, besides being used to decide whether some objective literal should be true, rules are also used to reject other rules. 
Hence, the concept of well-supportedness is also adopted to the rejection mechanism, and rules can only reject other rules with the additional constraint that the level of their heads be greater than the level of their bodies.

\subsubsection{Relationship between semantics based on causal rejection}
The rule update semantics introduced above are strongly related to one
another. The above considerations show that undesired stable models of the
\AS-, \JU- and \DS-semantics were eliminated by enlarging the set of rejected
rules or by shrinking the set of default assumptions. The following theorem
shows that no additional stable models were added in the process:

\begin{theorem}
	[\cite{Leite2003,Alferes2005}]
	Let $\dprg$ be a DLP. Then,
	\[
		\modas{\dprg} \supseteq \modju{\dprg} \supseteq \modds{\dprg} \supseteq
		\modrd{\dprg}
		.
	\]
	Moreover, for each inclusion above there exists a DLP for which the
	inclusion is strict.
\end{theorem}

Furthermore, all of these semantics coincide when only acyclic DLPs are
considered, showing that the differences in the definitions of rejected rules
and default assumptions are only relevant in the presence of cyclic
dependencies between literals.

\begin{theorem}
	[\citeN{Homola2004}]
	Let $\dprg$ be an acyclic DLP. Then,
	\[
		\modas{\dprg} = \modju{\dprg} = \modds{\dprg} = \modrd{\dprg}
		.
	\]
\end{theorem}

The relation between these semantics and other formalisms has also been
studied. It has been shown by \citeN{Eiter2002} that the \AS-semantics
coincides with the non-disjunctive case of the semantics for inheritance
programs by \citeN{Buccafurri1999}.

One of the open issues with these semantics is that one cannot easily
\emph{condense} a DLP to a single logic program that could be used
\emph{instead} of the DLP to perform further updates. The first obstacle is
that the stable models of a DLP may be non-minimal, as illustrated by the following example.
\begin{example}
Consider the DLP	$\dprg_4 = \seq{\set{\atma., \atmb \lpif \atma.}, \set{\lpnot \atma \lpif
	\lpnot \atmb.}}$. According to all semantics introduced so far, we obtain
$\modju{\dprg_4} = \modas{\dprg_4} = \modds{\dprg_4} = \modrd{\dprg_4}
		= \set{\emptyset, \set{\atma, \atmb}}$.
\end{example}

Since stable models of non-disjunctive programs are subset-minimal, no such
program can have the set of stable models $\modrd{\dprg_4}$. Condensing to a
disjunctive program is also problematic because rule update semantics are
constrained to non-disjunctive programs only, so after a condensation one
would not be able to perform any further updates.

Nevertheless, the original definition of the four discussed \S-semantics (\JU-, \AS-, \DS- and
\RD-semantics) was accompanied by a translation of a DLP to a single
non-disjunctive program \emph{over an extended language} whose stable models
correspond one-to-one to the \S-models assigned to the DLP under the
respective rule update semantics. Due to the language extension, the new
program cannot simply be updated directly as a substitute for the original
DLP, but may serve as a way to study the computational properties of the rule
update semantics and as a way to implement it using existing answer-set
solvers.

In the literature we can also find the semantics proposed by \citeN{Osorio2007} which can be seen as
simpler substitutes for the \AS-semantics. Unlike the semantics discussed
above, they are not defined declaratively, instead they are specified directly
by translating the initial program and its update to a single program
\emph{over the same language}. The first translation essentially weakens the
rules from the original program by making them defeasible, i.e.,\ a rule $\olit
\lpif \brl$ is transformed into the rule $\olit \lpif \brl, \lpnot
\cmpl{\olit}$. The authors have shown that the resulting semantics is
equivalent to the \AS-semantics if a single update is performed and the
updating program contains a tautology $\olit \lpif \olit$ for every objective
literal $\olit$. Due to its simplicity, because it only deals with a single update
in a way that cannot immediately be extended to additional updates, this semantics is \emph{not} sensitive
to the addition and removal of tautologies, but it adopts the problematic
behaviour of the \AS-semantics even when the tautologies are \emph{removed}
from the updating program. For example, when considering the DLPs
\begin{align*}
	\dprg_5 &= \seq{\set{\atma., \lnot \atma.}, \emptyset}
\enspace\text{and}\\
	 \dprg_5' &= \seq{
		\set{\atma., \lnot \atma.},
		\set{\atma \lpif \atma., \lnot \atma \lpif \lnot \atma.}
	}
\end{align*}
the \AS-semantics correctly assigns no \AS-model to $\dprg_5$ although its
sensitivity to tautological updates causes $\dprg_5'$ to have two \AS-models:
$\set{\atma}$ and $\set{\lnot \atma}$. The first semantics suggested by
\citeN{Osorio2007} assigns these two interpretations, $\set{\atma}$ and $\set{\lnot \atma}$, to both $\dprg_5$ and
$\dprg_5'$, so it exhibits problematic behaviour even on DLPs that were
correctly handled by the \AS-semantics.

The second translation is more involved as it produces a program that may not
be expressible by a non-disjunctive program. The resulting update semantics is
shown to coincide with the \AS-semantics in case only a single update is
performed. Note that since the \AS-semantics coincides with the \JU-semantics
on DLPs of length two\footnote{The difference between the \AS-semantics and \JU-semantics is that according to \JU\ you can have rejected
rules rejecting other rules, while according to \AS\ rejected rules cannot reject
other rules. According to both semantics, rules of the first program of a DLP cannot reject and rules of the last program of a DLP cannot be rejected. It follows that in a DLP of length two, according to the \JU-semantics, there cannot be a rule simultaneously rejecting and being rejected, so it follows that \AS\ and \JU\ coincide for DLPs of length two.}, the above mentioned relationships between semantics
by \cite{Osorio2007} and the \AS-semantics also hold for the \JU-semantics.
Their behaviour on DLPs of length three or more has not been studied.

\subsection{Preference-Based Semantics}

\label{sec:ru:preference}

A smaller group of rule update semantics relies on syntactic transformations
and semantics for \emph{prioritised logic programs}. These semantics do not consider
default negation in heads of rules.

Formally, a prioritised logic program is a pair $\tpl{\prg, \prec}$ where
$\prg$ is a program and $\prec$ is a strict partial order over $\prg$. The
intuitive meaning of $\prec$ is that if $\rla \prec \rlb$, then $\rlb$ is more
preferred than $\rla$. There exist a number of different semantics for
prioritised logic programs \citep{DBLP:journals/ai/BrewkaE99,DBLP:journals/tplp/DelgrandeST03,Schaub2003,Zhang2003}. Their goal is to plausibly use the preference
relation $\prec$ to choose the preferred stable models among the stable models
of $\prg$, or to constrain the rules of $\prg$ used to determine the stable models.

\subsubsection{The \PRZ-semantics}
One rule update semantics of this type was defined by \citeN{Zhang2006} and
relies on the semantics for prioritised logic programs proposed by
\citeN{Zhang2003}.\footnote{%
	Note that the preference relation in these papers is reversed w.r.t.\ the
	one we use here, i.e.,\ $\rla \prec \rlb$ means in the sense of
	\citeN{Zhang2003,Zhang2006} that $\rla$ is more preferred than $\rlb$.
}
Generally speaking, according to \citeN{Zhang2003}, the semantics is assigned to a prioritised logic program
$\tpl{\prg, \prec}$ by pruning away less preferred rules, obtaining an
ordinary logic program $\prg^\prec \subseteq \prg$ called a \emph{reduct}. Formally, $\prg^<$ is a reduct of $\tpl{\prg, \prec}$ if there exists a sequence of sets 
$\prg_{i}$ ($i=0,1,\ldots$) such that
\begin{align*}
\prg_{0} = \prg,\quad\quad\quad\quad\quad
\prg_{i+1} = \prg_{i}-\prgc_{i},\quad\quad\quad\quad\quad
\prg^< = \bigcap_{i=0}^{\infty}\prg_{i},
\end{align*}
where
\begin{align*}
\prgc_{i}=\{\rla\in\prg_{i} \mid &\exists\rla'\in\prg_{i},  \text{ such that }\forall\rla''\in\prgc_{i}, \rla''<\rla' \land \rla'' \triangleleft (\prg_{i}-\prgc_{i}) \text{ and}\\
&\not\exists \prgc'_{i}\subseteq\prg_{i} \text{ such that }\exists\rla''\in\prgc_{i},\forall\rla'\in\prgc'_{i}, \rla'<\rla'' \land \rla' \triangleleft (\prg_{i}-\prgc'_{i}) \},\\
\end{align*}
and $\rla\triangleleft\prgb$ denotes that rule $\rla$ is defeated by program $\prgb$, which is true if there exists some objective literal $\olit\in\twia\in\modsm{\prgb}$ such that $\olit\in\brl^-$.

A prioritised logic program may have zero or more reducts and the preferred
stable models are all the stable models of all the reducts.
For a detailed
discussion of reducts and their properties the reader can refer to the paper by
\citeN{Zhang2003}.

Subsequently, the update semantics defined by \citeN{Zhang2006} performs an
update of a program $\prga$ by a program $\prgu$ by executing the following
steps:
\begin{enumerate}[labelindent=2pt,labelwidth=6pt,leftmargin=!]
		\item Take some stable model $\twib_\prga$ of $\prga$.

		\item Determine the set $\mathit{Update}(\twib_\prga,\prgu)$ of interpretations resulting
		from updating $\twib_\prga$ by $\prgu$, in a way similar to to the approach of \citeN{DBLP:conf/jelia/MarekT94,Marek1998}, given by:
\[
	\mathit{Update}(\twib_\prga,\prgu)
	=
	\Set{
		\twia \in \modc{\prgu}
		|
		\lnot \exists \twib \in \modc{\prgu} :
			(\twib \div \twib_\prga) \subsetneq (\twia \div \twib_\prga)
	},
\]	
where $\div$ denotes set-theoretic symmetric difference. Choose any interpretation from $\mathit{Update}(\twib_\prga,\prgu)$
and denote it by $\twib_{\seq{\prga, \prgu}}$.			

		\item Extract a maximal subset $\prga'$ of $\prga$ that is coherent with
			$\twib_{\seq{\prga, \prgu}}$, i.e., such that there exists a
				stable model of $\prga' \cup \set{\olit. | \olit \in \twib_{\seq{\prga,
				\prgu}}}$.
		\item The set $\tpl{\prga' \cup
			\prgu, \prga' \times \prgu}^<$ of reducts of the prioritised logic program $\tpl{\prga' \cup
			\prgu, \prga' \times \prgu}$ is the result of updating $\prga$ by $\prgu$.
	\end{enumerate}
	
	As explained by \citeN{Zhang2006}, the intuition behind the first two steps is
	that simply taking a maximal subset of $\prga$ coherent with $\prgu$ is too
	crude an operation because it does not take into account the source of a
	conflict.

	\begin{example}
		[Intuition For Steps 1.\ and 2.\ \citep{Zhang2006}]
		\label{ex:ru:prz intuition}
		Consider the programs
		\begin{align*}
			\begin{aligned}
				\prga:
					&& \atma &. \\
					&& \atmb &\lpif \atmc.
			\end{aligned}
			&& \text{and}
			&& \begin{aligned}
				\prgu:
					&& \atmc &\lpif \atma. \\
					&& \lnot \atmb &\lpif \atmc.
			\end{aligned}
		\end{align*}
		Since $\prga \cup \prgu$ is incoherent, some part of $\prga$ needs to be
		eliminated to regain coherence. There are two maximal subsets of $\prga$
		that are coherent with $\prgu$: $\set{\atma.}$ and $\set{\atmb \lpif
		\atmc.}$. However, intuition suggests that the former set is preferable
		since the direct conflict between rules $(\atmb \lpif \atmc.)$ and $(\lnot
		\atmb \lpif \atmc.)$ provides a justification for eliminating the rule
		$(\atmb \lpif \atmc.)$ and thus keeping the fact $(\atma.)$.
	\end{example}

	The approach taken, then, is to first consider a stable model of $\prga$ and
	update it by $\prgu$, obtaining a new interpretation $\twib_{\seq{\prga,
	\prgu}}$ that reflects the new information in $\prgu$. Afterwards, a maximal
	set of rules from $\prga$ coherent with $\twib_{\seq{\prga, \prgu}}$ is used
	to form a prioritised logic program that prefers rules from $\prgu$ over rules
	from $\prga$. The reducts of this program form the result of the update.

	Due to the possibility of having multiple reducts as possible results of the
	update, it is not completely clear how updates can be iterated. Do we choose
	one reduct and commit to it? Which one do we choose, then? Or do we simply
	consider all of the reducts and all possible evolutions? Due to these
	unresolved issues, we formally define this semantics only for DLPs of length
	two. We call it \emph{preference-based Zhang's semantics}, or
	\emph{\PRZ-semantics} for short.

	\begin{definition}
		[\PRZ-Semantics \citep{Zhang2006}]
		Let $\dprg = \seq{\prga, \prgu}$ be a DLP without default negation in heads
		of rules. The set $\modprz{\dprg}$ of \emph{\PRZ-models of $\dprg$} is the
		union of sets of stable models of all reducts obtained by performing the
		steps 1\ --\ 4\ above.
	\end{definition}

	One distinguishing feature of the \PRZ-semantics is that by relying on a
	stable model of $\prga$ for conflict resolution, it is unable to detect
	``latent'' conflicts between rules that have not been ``triggered'' in the
	initial stable model or its update. This is illustrated in the following
	example:\footnote{%
		This example does not apply to an earlier version of the \PRZ-semantics by
		\citeN{Zhang1998}. This is because the maximal subset $\prga'$ of $\prga$
		chosen for constructing the prioritised logic program is required to be
		coherent with both $\twib_{\seq{\prga, \prgu}}$ and $\prgu$, not only with
		$\twib_{\seq{\prga, \prgu}}$ as by \citeN{Zhang2006}.
	}

	\begin{example}
		[Undetected Latent Conflicts in the \PRZ-semantics]
		Consider the programs
		\begin{align*}
			\begin{aligned}
				\prga:
					&& \atma &\lpif \atmc. \\
					&& \atmb &\lpif \atmc.
			\end{aligned}
			&& \text{and}
			&& \begin{aligned}
				\prgu:
					&& \atmc&. \\
					&& \lnot \atma &\lpif \atmb.
			\end{aligned}
		\end{align*}
		and let $\dprg = \seq{\prga, \prgu}$. The single stable model of $\prga$ is
		$\twib_\prga = \emptyset$ and its update by $\prgu$ results in the
		interpretation $\twib_{\seq{\prga, \prgu}} = \set{\atmc}$ which is coherent
		with $\prga$. The resulting prioritised logic program $\tpl{\prga \cup
		\prgu, \prga \times \prgu}$ has only one reduct, $\prga \cup \prgu$, that
		has no stable model. In other words, $\modprz{\dprg} = \emptyset$ and the
		conflict between $\prga$ and $\prgu$ remained unresolved. Note also that
		$\modas{\dprg} = \modju{\dprg} = \modds{\dprg} = \modrd{\dprg} =
		\set{\set{\lnot \atma, \atmb, \atmc}}$.
	\end{example}

	The \PRZ-semantics is also sensitive to tautological
	updates:

	\begin{example}
		[Tautological Updates in the \PRZ-semantics]
		Consider the programs
		\begin{align*}
			\begin{aligned}
				\prga:
					&& \atma &\lpif \lpnot \lnot \atma. \\
					&& \lnot \atma &\lpif \lpnot \atma.
			\end{aligned}
			&& \text{and}
			&& \begin{aligned}
				\prgu:
					&& \atma &\lpif \atma.
			\end{aligned}
		\end{align*}
		Both stable models $\set{\atm}$ and $\set{\lnot \atm}$ of $\prga$ remain
		unchanged after an update by $\prgu$ and thus both rules of $\prga$ are
		retained in the resulting prioritised logic program $\tpl{\prga \cup \prgu,
		\prga \times \prgu}$. Its only reduct, however, is the program $\set{\atma
		\lpif \lpnot \lnot \atma., \atma \lpif \atma.}$, which has a single stable
		model $\set{\atma}$. The tautological update has thus discarded one of the
		stable models of $\prga$.
	\end{example}

\subsubsection{The \PRDi-, \PRWi, and \PRBi-semantics}

	Preference-based rule update semantics were also considered by
	\citeN{Delgrande2007}, utilising the semantics for prioritised logic programs
	examined by \citeN{Schaub2003}.\footnote{%
		Prioritised logic programs are called \emph{ordered logic programs} by
		\citeN{Schaub2003} and by \citeN{Delgrande2007}.
	}
	Instead of defining how a prioritised logic program $\tpl{\prg, \prec}$ can be
	characterised in terms of reducts, as done by \citeN{Zhang2003}, \citeN{Schaub2003}
	specify conditions that a stable model of $\prg$ must satisfy in order to be a
	\emph{preferred stable model of $\tpl{\prg, \prec}$}. They use three such
	conditions, defined in the literature on programs with preferences, dubbed
	\emph{\X[D]-preference}, \emph{\X[W]-preference} and \emph{\X[B]-preference},
	which yield an increasing number of preferred stable models. For further
	details about these preference strategies the reader can refer to the paper by
	\citeN{Schaub2003} and the references therein.

	Unlike in the approach by \citeN{Zhang2006}, the methodology chosen by \citeN{Delgrande2007} for performing rule
	updates is based on relatively simple transformations into a prioritised logic
	program. In order to define these transformations, we first need to introduce
	the following notation for arbitrary programs $\prga$ and $\prgu$:
	\begin{align*}
		\prga^d &= \Set{
			\olit \lpif \brl, \lpnot \cmpl{\olit}.
			|
			(\olit \lpif \brl.) \in \prga
		}, \\
		C(\prga, \prgu) &= \{
			\tpl{\rla, \rlb}
			|
			\exists \olit \in \olits :
			\rla \in \prga
			\land \rlb \in \prgu \land \hrla = \set{\olit}
			\land \hrlb = \set{\cmpl{\olit}}
		\}, \\
		c(\prga, \prgu) &= \Set{
			\rla, \rlb
			|
			\tpl{\rla, \rlb} \in C(\prga, \prgu)
		}.
	\end{align*}
	Intuitively, $\prga^d$ denotes a program obtained from $\prga$ by making all
	its rules defeasible, analogically to the semantics based on weakenings by
	\citeN{Osorio2007}. The set $C(\prga, \prgu)$ contains pairs of rules from
	$\prga$ and $\prgu$ with conflicting heads and $c(\prga, \prgu)$ contains
	rules from $\prga$ and $\prgu$ involved in such conflicts.

	\citeN{Delgrande2007} proposed three different operators for updating a
	program $\prga$ by a program $\prgu$, each of which outputs a different
	prioritised logic program:
	\begin{align*}
		\prga *_0 \prgu &= \Tpl{\prga^d \cup \prgu^d, \prga^d \times \prgu^d}, \\
		\prga *_1 \prgu &= \Tpl{\prga^d \cup \prgu^d, C(\prga^d, \prgu^d)}, \\
		\prga *_2 \prgu &= \Tpl{
			c(\prga, \prgu)^d \cup ((\prga \cup \prgu) \setminus c(\prga, \prgu)),
			C(\prga^d, \prgu^d)
		}.
	\end{align*}
	Informally, $*_0$ makes all rules from $\prga$ and $\prgu$ defeasible and
	gives preference to every rule from $\prgu$ over any rule from $\prga$. The
	operator $*_1$ produces a more cautious preference relation, only preferring
	rules from $\prgu$ over rules from $\prga$ with conflicting heads. In
	addition, the operator $*_2$ refrains from making defeasible rules that are not
	involved in any conflict.

	It is argued by \citeN{Delgrande2007} that these operators can be naturally
	generalised to account for arbitrary (finite) sequences of programs as
	follows:
	\[
		* (\seq{\prg_\lia}_{\lia < \lng}) = \begin{cases}
			\prg_0 * \prg_1 & \text{if $\lng = 2$} ,\\
			* (\seq{\prg_\lia}_{\lia < \lng - 1}) * \prg_{\lng - 1}
			& \text{if $\lng > 2$}.
		\end{cases}
	\]
	This definition is slightly incomplete since the result of
	operators $*_0$, $*_1$, and $*_2$ is not an ordinary logic program but a
	prioritised one. The question then arises as to what happens with the priority
	relation of an intermediate result, say $\prg_0 * \prg_1$, when it is further
	updated by $\prg_2$. In the following, we assume that the preference relations are merged and
	measures are taken to ensure that the merged relation remains a strict partial
	order, i.e.,\ transitivity is enforced after the merge. We can now define the
	update semantics by \cite{Delgrande2007} for arbitrary DLPs. We call them the
	\emph{\PRXi-semantics} with \X{} representing the preference strategy (i.e.,\
	\X{} is one of \X[D]{}, \X[W]{}, or \X[B]{}), and $i$ denoting the particular
	operator used for forming the prioritised logic program (i.e.,\ $i \in \set{0,
	1, 2}$).

	\begin{definition}
		[\PRXi-Semantics \citep{Delgrande2007}]
		Let $\dprg$ be a DLP without default negation in heads of rules, \X{} be one
		of \X[D]{}, \X[W]{}, or \X[B]{}, and $i \in \set{0, 1, 2}$. The set
		$\modprxi{\dprg}$ of \emph{\PRXi-models of $\dprg$} is the set of preferred
		stable models of the prioritised logic program $*_i (\dprg)$ under the
		preference strategy \X{}.
	\end{definition}

	The overall properties of these rule update semantics depend on the chosen
	operator ($*_0$, $*_1$, or $*_2$) and on the chosen preference strategy
	(\X[D]-, \X[W]-, or \X[B]-preference). Nevertheless, as \citeN{Delgrande2007} illustrated by examples, all \PRXi-semantics are sensitive to tautological
	updates. In addition, the following example shows an interesting behaviour that
	distinguishes these semantics from the previously discussed ones:

	\begin{example}
		[Default Assumptions vs. Facts \citep{Delgrande2007}]
		\label{ex:ru:default assumptions preferred to facts:1}
		Consider the programs
		\begin{align*}
			\begin{aligned}
				\prga:
					&& \lnot \atma &.
			\end{aligned}
			&& \text{and}
			&& \begin{aligned}
				\prgu:
					&& \atma &\lpif \lpnot \lnot \atma.
			\end{aligned}
		\end{align*}
		and let $\dprg = \seq{\prga, \prgu}$. For any operator $*_i$ and preference
		strategy \X{}, $\modprxi{\dprg} = \set{\set{\atma}}$. This indicates that
		the default assumption in the updating program is given preference over the
		fact in the initial program. If we interpret $\atma$ as
		$\prest{man}(\const{mary})$, then this example shows that if initially
		$\prest{man}(\const{mary})$ is known to be false and later we learn that
		\[
			\prest{man}(\vara) \lpif \lpnot \lnot \prest{man}(\vara).
			,
		\]
		meaning that by default all individuals are men, then this immediately
		changes our knowledge about $\prest{man}(\const{mary})$: we now know that
		$\prest{man}(\const{mary})$ is true!

		It seems more natural to give preference to initial facts over default
		assumptions in more recent rules. Note that $\modas{\dprg} = \modju{\dprg}
		= \set{\set{\atma}, \set{\lnot \atma}}$ and $\modds{\dprg} = \modrd{\dprg} =
		\modprz{\dprg} = \set{\set{\lnot \atma}}$, i.e.,\ the causal rejection
		semantics with unrestricted set of default assumptions allow both
		$\set{\atma}$ and $\set{\lnot \atma}$ to be stable models of $\dprg$ while
		the ``fixed'' versions of these semantics together with Zhang's
		preference-based semantics actually prefer the initial fact over the default
		assumption.
	\end{example}

	\subsection{Other Approaches}

	\label{sec:ru:other}

\subsubsection{The \RVS-semantics}
	\citeN{Sakama2003} have proposed a rule update semantics that is clearly based
	on ideas from belief revision, similarly as formula-based belief update
	operators. In particular, they define that a program $\prga' \cup \prgu$
	achieves the update of $\prga$ by $\prgu$ if $\prga'$ is a maximal subset of
	$\prga$ such that $\prga' \cup \prgu$ is coherent, i.e., it has a stable model.

	As with the \PRZ-semantics, we define the semantics of \citeN{Sakama2003} only
	for DLPs of length two because it is not clear how one should deal with multiple
	results of an update. We call the resulting semantics the
	\emph{\RVS-semantics}:

	\begin{definition}
		[\RVS-Semantics \citep{Sakama2003}]
		Let $\dprg = \seq{\prga, \prgu}$ be a DLP. The set $\modrvs{\dprg}$ of
		\emph{\RVS-models of $\dprg$} is the union of sets of stable models of all
		programs $\prga' \cup \prgu$ where $\prga'$ is a maximal subset of $\prga$
		such that $\prga' \cup \prgu$ is coherent.
	\end{definition}

	Similarly to the \PRZ-semantics, as discussed in Example~\ref{ex:ru:prz intuition}, the approach adopted by the
	\RVS-semantics pays no attention to the source of conflicts --- any solution of
	a conflict is as good as any other as long as only a minimal set of rules is
	eliminated. Another consequence is that conflicts are removed at any cost,
	even if there is no plausible way to explain why the update should restore
	coherence. This has been criticised by \citeN{Leite2003}, who argued that every
	conflict has several causes and each type of conflict should be dealt with
	accordingly. One consequence of this is that an empty or tautological update
	may restore coherence (and consistency) of an initial program. If we compare
	this to belief change principles and operators, such a behaviour is typical of
	\emph{revision} but is not desirable for \emph{updates}.
	\citeN{DBLP:journals/jair/GarciaLSPW19} introduced a family of revision operators for logic programs similar in spirit to the \RVS-semantics, except that they allow both for the addition and/or removal of rules from a logic program to achieve coherence. As with the \RVS-semantics, these operators are sensitive to empty or tautological updates.

\subsubsection{The \RVD-semantics}
	Similar ideas form the basis of the rule \emph{revision} semantics proposed by
	\citeN{Delgrande2010}. Note that since the distinction between program update
	and revision, as these terms are used in the literature, is somewhat blurry,
	in the following we also refer to this semantics as an \emph{update
	semantics}. Informally, the stable model of a sequence of programs is
	constructed by first keeping all rules from the last program and committing to
	a minimal set of default literals used to derive one of its stable models.
	Subsequently, a maximal coherent subset of the previous program is added and
	further commitments are made. This process is iterated until the first program
	of the sequence is processed, as illustrated by the following example.
\begin{example}
	Consider the DLP
	\[
		\dprg = \seq{
			\set{\atma.},
			\set{\atmb.},
			\set{\atmc \lpif \lpnot \atma., \atmc \lpif \lpnot \atmb.}
		}
		.
	\]
	We start with the last program of the sequence which has the stable model
	$\set{\atmc}$. This stable model can be derived either using the literal
	$\lpnot \atma$ or $\lpnot \atmb$ and we need to choose one of these and commit to
	it. If we pick the former, the overall set of literals we commit to at this
	stage is $\set{\lpnot \atma, \atmc}$. We then proceed to the second program
	and realise that it is coherent with our commitments as well as the rules from
	the last program. We thus add $\atmb$ to our set of commitments and the fact
	$(\atmb.)$ to the set of rules that we are going to keep. Proceeding to the
	first program of the sequence, the rule within it is inconsistent with our
	commitment to $\lpnot \atma$, so the rule needs to be discarded. The set of
	objective literals we committed to until now, namely $\set{\atmb, \atmc}$,
	forms one stable model of $\dprg$. Note that if we initially commit to $\lpnot
	\atmb$, we obtain the stable model $\set{\atma, \atmc}$.
\end{example}

To formalise this construction, \citeN{Delgrande2010} uses three-valued interpretations,
defined as pairs of disjoint interpretations $\twib = (\twib^+,\twib^-)$, which we dub 
\emph{three-valued d-interpretations} to distinguish from the three-valued interpretations used 
elsewhere in this paper, based on which a modified notion of 
reduct of a program $\prg$ without default negation in the head is 
defined as 
\[
\prg^\twib = \{\an{\hrlp \lpif \brlp\cup\lpnot\brln\setminus \twib^-}
			| 
			\rl \in \prg
			\land
			\twib^+ \cap \brln = \emptyset\}.
\] 
This reduct is used to define a special notion of three-valued answer-sets of a program $\prg$ which are those three-valued d-interpretations $\twib = (\twib^+,\twib^-)$
such that $\least{\prg^{\twib^+}}=\least{\prg^{\twib}}=\twib^+$ and for any 
$\twia = (\twib^+,\twia^-)$ such that
$\twia^- \subset \twib^-$
we have that $\least{P^{\twia}} \neq \twib^+$, 
 where $\least{\cdot}$ is as before. 
 Additionally, \citeN{Delgrande2010} defines a concept of canonical program corresponding to a three-valued d-interpretation $\twib = (\twib^+,\twib^-)$ as
\[
Pgm(\twib)=
\{ \atma. \mid \atma \in \twib^+ \}
\;\cup\;
\{ \lpif \atma. \mid \atma \in \twib^- \}.
\] 

Then, given a DLP without default negation in heads of rules $\dprg = \seq{\prg_\lia}_{\lia < \lng+1}$, an interpretation
$\twib$ is an \emph{r-answer-set of $\dprg$} iff 
there is a sequence
$\seq{(\prg^r_\lia,\twib_\lia)}_{\lia < \lng+1}$
such that 
\begin{enumerate}[labelindent=2pt,labelwidth=6pt,leftmargin=!]
\item $\prg^r_\lng = \prg_\lng$ and $\twib_\lng$ is a three-valued answer-set of $\prg_\lng$; 
\item for $\lia < \lng$,
 $\prg^r_\lia$ is a maximal set of rules of $\prg_\lia$ consistent with $\prg^r_{\lia+1} \cup Pgm(\twib_{\lia+1})$ (or $\olits$ if $\prg^r_{\lia+1} \cup Pgm(\twib_{\lia+1})$ is inconsistent) and $\twib_\lia$ is a three-valued answer-set of $\prg^r_\lia$;
\item $\twib=\twib^+_1$.
\end{enumerate}
We refer to this semantics as
	the \emph{\RVD-semantics}:

	\begin{definition}
		[\RVD-Semantics \citep{Delgrande2010}]
		Let $\dprg$ be a DLP without default negation in heads of rules. The set $\modrvd{\dprg}$ of \emph{\RVD-models of
		$\dprg$} is the set of r-answer-sets of $\dprg$.
	\end{definition}

	Similarly as the \RVS-semantics, the \RVD-semantics resolves conflicts at any
	cost, a consequence of which is that empty and tautological updates restore
	coherence and consistency. Furthermore, it exhibits the same behaviour as the
	\PRXi-semantics in Example~\ref{ex:ru:default assumptions preferred to
	facts:1}, i.e.,\ it prefers to satisfy default assumptions in further programs
	to satisfying earlier facts. It actually goes even further than the
	\PRXi-semantics, as illustrated in the following example:

	\begin{example}
		[Default Assumptions vs. Facts in the \RVD-semantics]
		\label{ex:ru:default assumptions preferred to facts:2}
		Consider the programs
		\begin{align*}
			\begin{aligned}
				\prga:
					&& \atma &.
			\end{aligned}
			&& \text{and}
			&& \begin{aligned}
				\prgu:
					&& \atmb &\lpif \lpnot \atma.
			\end{aligned}
		\end{align*}
		and let $\dprg = \seq{\prga, \prgu}$. We obtain $\modrvd{\dprg} =
		\set{\set{\atmb}}$, as opposed to $\modas{\dprg} = \modju{\dprg} =
		\modds{\dprg} = \modrd{\dprg} = \modprz{\dprg} = \modprxi{\dprg} =
		\modrvs{\dprg} = \set{\set{\atma}}$. This indicates that the default
		assumptions in the updating program are given preference over facts from the
		initial program even more aggressively than in case of the \PRXi-semantics.
		If we interpret $\atma$ as $\prest{dog}(\const{bo})$ and $\atmb$ as
		$\lnot \prest{canBark}(\const{bo})$, then this example shows that if
		initially $\prest{dog}(\const{bo})$ is known to be true and later we
		learn that
		\[
			\lnot \prest{canBark}(\vara) \lpif \lpnot \prest{dog}(\vara).
			,
		\]
		meaning that, by default, individuals that are not dogs cannot bark, then
		this immediately modifies our knowledge about $\const{bo}$: we no longer
		know whether $\prest{dog}(\const{bo})$ is true or not and, in addition,
		we conclude that $\prest{canBark}(\const{bo})$ is (explicitly) false. Poor $\const{bo}$\ldots
	\end{example}
	
	A methodology based on maximal subsets of the initial program coherent with
	its update was also used by \citeN{Osorio2007a} for updating programs under
	the \emph{pstable model semantics}. The idea is used indirectly by augmenting
	the bodies of original rules with additional literals and using an abductive
	framework to minimise the set of rules ``disabled'' by falsifying the added
	literal. We do not further consider this semantics because it diverges from
	the standard notion of a stable model and uses pstable models instead.

	Finally, there also exist approaches based on a semantic framework that
	directly encodes literal dependencies induced by rules, and performs changes
	on the dependencies instead of on the rules themselves. The advantage over
	dealing with rules is that the dependency relation is monotonic, so AGM
	postulates and operators can be applied to it directly
	\citep{Krumpelmann2010,Krumpelmann2012}. In the work of
	\citeN{Sefranek2006,Sefranek2011}, the dependency framework is used for
	specifying \emph{irrelevant updates}, an instance of which are tautological
	updates, and designing update semantics immune to such irrelevant updates.

\subsection{Fundamental Properties}

\label{sec:ru:fundamental properties}

As demonstrated above, rule update semantics are based on a number of
different approaches and constructions and provide different results even on
very simple examples. 

In this section, we indicate and examine some fundamental properties of rule update semantics. 
We call them \emph{syntactic} because they have been discussed in the context of
syntax-based semantics for rule updates and, with only two exceptions, 
their formulation requires that we refer to the
syntax of the respective DLP. The first three properties, as well as the last one, will be satisfied by all
rule update semantics that we formally introduced above. The remaining four will
only be satisfied by a subset of the semantics, serving as entry
points for comparing them. 

Recall that the distinct semantics have been defined for different classes of
DLPs, with the assumption that none contained integrity constraints. The properties defined below 
do not have that assumption. 
When we say that a semantics \S{} satisfies a particular property, we
constrain ourselves only to DLPs in the scope of the definition of \S{}.
The classes of
DLPs to which the introduced semantics are applicable is summarised in
Table~\ref{table:ru:applicability}.

The reader can
find a systematic account of the proofs of the theorems in this subsection in the PhD thesis by \cite{Slota2012Thesis}, either by 
presenting the proof or pointing to the relevant paper where the result was first proved.

\begin{table}
	\caption{Applicability of rule update semantics}
\begin{center}
		\begin{tabular}{cp{0.5cm}p{9cm}}
			\toprule
			Semantics
				&& Applicability
				\\ \midrule\midrule
			\AS{}, \JU{}, \DS{}, \RD{}
				&& Arbitrary DLPs without integrity constraints.
				\\ \midrule
			\PRXi{}, \RVD{}
				&& DLPs without default negation in heads of rules and without integrity constraints.
				\\ \midrule
			\PRZ{}, \RVS{}
				&& DLPs of length two without default negation in heads of rules and without integrity constraints.
				\\ \bottomrule
		\end{tabular}
\end{center}
 	\label{table:ru:applicability}
\end{table}

The first fundamental property captures the fact that rule update semantics
produce only \emph{supported} models. In a static setting, support
\citep{Apt1988,Dix1995a} is one of the basic conditions that logic
programming semantics are intuitively designed to satisfy. Its generalisation
to the dynamic case is straightforward.

\begin{definition}
	[Support]
	Let \S{} be a rule update semantics, $\prg$ a program, $\olit$ an objective
	literal and $\twib$ an interpretation. We say that
	\begin{itemize}[labelindent=2pt,labelwidth=6pt,leftmargin=!]
		\item \emph{$\prg$ supports $\olit$ in $\twib$} if for some rule $\rl \in
			\prg$, $\olit \in \hrl$, and $\twib \ent \brl$;

		\item \emph{$\prg$ supports $\twib$} if every objective literal $\olit \in
			\twib$ is supported by $\prg$ in $\twib$;

		\item \emph{\S{} respects support} if for every DLP $\dprg$ to which \S{}
			is applicable and every \S-model $\twib$ of $\dprg$, $\all{\dprg}$
			supports $\twib$.
	\end{itemize}
\end{definition}

In other words, a rule update semantics \S\ respects support if every objective
literal $\olit$ that is true in an \S-model of a DLP is the head
of some rule of that DLP whose body is true in the same model. Such a rule then provides a
\emph{justification} for $\olit$.

A consequence of support is that the rule update semantics satisfies \emph{language conservation}, 
defined as follows:

\begin{definition}
	[Language Conservation for Rule Updates]
	\label{def:ru:language conservation}
	Let \S{} be a rule update semantics. We say that \S{} \emph{conserves the
	language} if for every set $\lang$ of propositional variables, every DLP $\dprg =
	\seq{\prg_\lia}_{\lia < \lng}$ to which \S{} is applicable, and every \S-model
	$\twib$ of $\dprg$, if $\preds{\prg_\lia} \subseteq \lang$ for all $\lia <
	\lng$, then $\preds{\twib} \subseteq \lang$. Where $\preds{\prg_\lia}$ (resp. $\preds{\twib}$) denotes the set of all
	propositional variables appearing in $\prg_\lia$ (resp. $\twib$).
\end{definition}

Informally, if both the initial
and updating theories represent knowledge about propositional variables from the set $\lang$,
then the updated theory should not introduce knowledge about propositional variables that
do not belong to $\lang$.

Though support and language conservation are basic requirements, and
certainly too weak to be sufficient for a ``good'' rule update semantics, they
seem to be intuitive from the logic programming perspective. And, indeed,
they are satisfied by all rule update semantics that we introduced previously.

\begin{theorem}
	[Respect for Support and Language Conservation]
	\label{thm:ru:support}
	Let \X{} be one of \X[D]{}, \X[W]{}, \X[B]{} and $i \in \set{0, 1, 2}$. The
	rule update semantics \AS{}, \JU{}, \DS{}, \RD{}, \PRZ{}, \PRXi{}, \RVS{},
	and \RVD{} respect support and conserve the language.
\end{theorem}

The third fundamental property for rule update semantics expresses the usual
expectation regarding how facts should be updated by newer facts.

\begin{definition}
	[Fact Update]
	\label{def:ru:fact update}
	Let \S{} be a rule update semantics.
	We say that \emph{\S{} respects fact update} if for every finite sequence $\dprg = \seq{\prg_\lia}_{\lia < \lng}$ of
	consistent sets of facts to which
	\S{} is applicable, the unique \S-model of $\dprg$ is the interpretation
\[
		\{
			\olit \in \olits |
			\exists \lib < \lng : (\olit.) \in \prg_\lib
			\land(\forall \lia :
				\lib < \lia < \lng \mlthen
				\Set{\cmpl{\olit}., \lpnot \olit.} \cap \prg_\lia = \emptyset
			)
		\}.
\]
\end{definition}

Fact update enforces literal inertia, which forms the basis for belief update
operators such as the one by Winslett, but only for the case when both the
initial program and its updates are consistent sets of facts. Similarly as
before, all rule update semantics adhere to this property.

\begin{theorem}
	[Respect for Fact Update]
	\label{thm:ru:fact update}
	Let \X{} be one of \X[D]{}, \X[W]{}, \X[B]{} and $i \in \set{0, 1, 2}$. The
	rule update semantics \AS{}, \JU{}, \DS{}, \RD{}, \PRZ{}, \PRXi{}, \RVS{}, and
	\RVD{} respect fact update.
\end{theorem}

The fourth and fifth syntactic properties are fundamental for all semantics
based on causal rejection. The first of them is the causal rejection principle
itself.

\begin{definition}
	[Causal Rejection]
	Let \S{} be a rule update semantics.
	We say that \emph{\S{} respects causal rejection} if for every DLP $\dprg =
	\seq{\prg_\lia}_{\lia < \lng}$ to which \S{} is applicable, every \S-model
	$\twib$ of $\dprg$, all $\lia < \lng$, and all rules $\rla \in \prg_\lia$,
\[		\twib \nent \rla \enspace\enspace\enspace\text{implies}\enspace\enspace\enspace
		\exists \lib \, \exists \rlb :
			\lia < \lib < \lng
			\land \rlb \in \expv{\prg_\lib}
			\land \rla \confl{}{} \rlb
			\land \twib \ent \brlb.
\]
\end{definition}

This principle requires a \emph{cause} for every violated rule in the form of
a more recent rule with a conflicting head and a satisfied body. It is
hard-wired in the definitions of sets of rejected rules of the four rule
update semantics that are based on it.

\begin{theorem}
	[Respect for Causal Rejection]
	\label{thm:ru:causal rejection principle}
	The rule update semantics \AS{}, \JU{}, \DS{}, and \RD{} respect causal
	rejection.
\end{theorem}

\begin{example}
The following examples illustrate why each of the
	rule update semantics \PRZ{}, \PRXi{}, \RVS{}, and \RVD{} does not respect causal rejection. 
\begin{align*}
	&\dprg_1 = \seq{\set{\atma \lpif \atmb.,\lnot\atma \lpif \atmb.}, \set{\atmb.}}
\text{, where } %\enspace
 \modprz{\dprg_1}
		= \set{\set{\lnot \atma, \atmb}, \set{\atma,\atmb}};\\
	&\dprg_2 = \seq{\set{\atma.,\lnot\atma.}, \set{\atmb \lpif \atma.}}
\text{, where } %\enspace
 \modprxiX{B}{0}{\dprg_2} =  \modprxiX{B}{1}{\dprg_2} 
		= \set{\set{\lnot\atma},\set{\atma,\atmb}};\\
	&\dprg_3 = \seq{\set{\atma.,\lnot\atma.}, \set{\atmb \lpif \atma.}}
\text{, where } %\enspace
 \modprxiX{D}{0}{\dprg_3} =  \modprxiX{D}{1}{\dprg_3} =  \modprxiX{W}{0}{\dprg_3} =  \modprxiX{W}{1}{\dprg_3}
		= \set{\set{\lnot\atma}};\\
	&\dprg_4 = \seq{\set{\atma.,\lnot\atma.}, \set{\atma \lpif \atmb.,\lnot\atma \lpif \atmb.}}
\text{, where } %\enspace
 \modprxiX{B}{2}{\dprg_4} =  \modprxiX{D}{2}{\dprg_4} = \modprxiX{W}{2}{\dprg_4}
		= \set{\set{\lnot\atma},\set{\atma}};\\	
	&\dprg_5 = \seq{\set{\atma \lpif \lpnot\atma.}, \set{}}
\text{, where } %\enspace
 \modrvs{\dprg_5}
		= \set{\set{}}; \text{ and }\\
	&\dprg_6 = \seq{\set{\atma.}, \set{\atmb \lpif \lpnot\atma.}}
\text{, where } %\enspace
 \modrvd{\dprg_6}
		= \set{\set{\atmb}}.
\end{align*}
\end{example}

The sixth syntactic property stems from the fact that all rule update
semantics based on causal rejection coincide on acyclic DLPs
\citep{Homola2004,Alferes2005}. Thus, the behaviour of any rule update
semantics on acyclic DLPs can be used as a way to compare it to all these
semantics simultaneously.

\begin{definition}
	[Acyclic Justified Update]
	Let \S{} be a rule update semantics.
	We say that \emph{\S{} respects acyclic justified update} if for every
	acyclic DLP $\dprg$ to which \S{} is applicable, the set of \S-models
	of $\dprg$ coincides with $\modju{\dprg}$.
\end{definition}

\begin{theorem}
	[Respect for Acyclic Justified Update]
	\label{thm:ru:acyclic ju}
	The rule update semantics \AS{}, \JU{}, \DS{}, and \RD{} respect acyclic
	justified update.
\end{theorem}

%%%%%%%%%%%%%%%%%
The next property has been extensively
discussed in the literature, and has been at the heart of the motivation for
developing some of the variants of semantics based on causal rejection. It 
requires that the semantics be immune to tautological updates  (i.e.,\ an update composed only of
rules whose head literal also belongs to its body), the intuition being that 
such update cannot indicate any change in
the modelled world because it is always true. Immunity to tautological updates is a
desirable property of belief updates in classical logic, being a direct consequence of postulate \labbu{2}.

\begin{definition}
	[Immunity to Tautological Updates]
	Let \S{} be a rule update semantics, $\rl$ a tautological rule, and $\dprg^{\rl}$ the DLP obtained from DLP $\dprg =
	\seq{\prg_\lia}_{\lia < \lng}$ by appending $\prg_\lng=\{\rl\}$. 	
	We say that \emph{\S{} respects immunity to tautological updates} if for every DLP $\dprg =
	\seq{\prg_\lia}_{\lia < \lng}$ and every tautological rule $\rl$ such that \S{} is applicable to $\dprg^{\rl}$, $\mods{\dprg}=\mods{\dprg^{\rl}}$.
	\end{definition}

As it turns out, only the \RD-semantics is immune to tautological updates. 

\begin{theorem}
	[Immunity to Tautologies]
	\label{thm:ru:immunity_to_tau}
	The rule update semantics \RD{} respects immunity to tautological updates.
\end{theorem}

The remaining semantics are not immune to tautologies for one of the following three reasons: 
\begin{enumerate}[labelindent=2pt,labelwidth=6pt,leftmargin=!]
\item the tautology can be used to reject other rules, as is the case with most semantics based on causal rejection, namely \AS{}, \JU{}, and \DS{}, such as for example 
\begin{align*}
&\dprg_1 = \seq{\set{\atma., \lnot \atma.}, \set{\atma \lpif \atma.}} \text{, where } \modju{\dprg_1} = \modas{\dprg_1} = \modds{\dprg_1}
= \set{\set{\atma}};\\
&\dprg_2 = \seq{\set{\atma.}, \set{\lnot \atma.}, \set{\atma \lpif \atma.}} \text{, where } \modas{\dprg_2} = \set{\set{\lnot \atma} \set{\atma}};\\
&\dprg_3 = \seq{\set{\atma.}, \set{\lpnot \atma \lpif \lpnot \atma.}} \text{, where } \modju{\dprg_3} = \modas{\dprg_3} = \set{\emptyset, \set{\atma}};
\end{align*}
\item the tautology can be used as a reason not to prefer other rules, as is the case with semantics based on preferences, namely \PRZ{} and \PRXi{} (for \X{} be one of \X[D]{}, \X[W]{}, \X[B]{} and $i \in \set{0, 1, 2}$), such as for example
\begin{align*}
&\dprg_4 = \seq{\set{\atma \lpif \lpnot \lnot \atma., \lnot\atma\lpif\lpnot\atma.}, \set{\atma \lpif \atma.}} \text{, where } \modprz{\dprg_4} = \set{\set{\atma}};\\
&\dprg_5 = \seq{\set{\atma \lpif \lpnot \lnot \atma., \lnot\atma.}, \set{\atma \lpif \atma.}} \text{, where } \modprxi{\dprg_5} = \set{\set{\atma}};
\end{align*}
\item conflicts are resolved at any cost, as is the case with the \RVS{} and
	\RVD{} semantics, such as for example
\begin{align*}
&\dprg_6 = \seq{\set{\atma \lpif \lpnot \atma.}, \set{\atmb \lpif \atmb.}} \text{, where } \modrvs{\dprg_6} =\modrvd{\dprg_6} = \set{\emptyset};\\
&\dprg_7 = \seq{\set{\atma., \lnot\atma.}, \set{\atmb \lpif \atmb.}} \text{, where } \modrvs{\dprg_7} = \modrvd{\dprg_7} = \set{\set{\atma},\set{\lnot\atma}}.
\end{align*}
\end{enumerate}
	
The next two properties are no longer syntactical as the ones considered so far, but we mention them here since they have 
been often discussed together with the semantics considered in this section. 

The first one can be seen as a weaker version of immunity to tautological updates. It imposes that semantics be
immune to empty updates, and 
is obeyed by a 
larger set of semantics.

\begin{definition}
	[Immunity to Empty Updates]
	Let \S{} be a rule update semantics and $\dprg^\emptyset$ the DLP obtained from DLP $\dprg =
	\seq{\prg_\lia}_{\lia < \lng}$ by appending $\prg_\lng=\emptyset$. 	
	We say that \emph{\S{} respects immunity to empty updates} if for every DLP $\dprg =
	\seq{\prg_\lia}_{\lia < \lng}$ such that \S{} is applicable to $\dprg^\emptyset$, $\mods{\dprg}=\mods{\dprg^\emptyset}$.
	\end{definition}

\begin{theorem}
	[Immunity to Empty Updates]
	\label{thm:ru:immunity_to_empty_update}
	Let \X{} be one of \X[D]{}, \X[W]{}, or \X[B]{}. The
	rule update semantics \AS{}, \JU{}, \DS{}, \RD{}, \PRZ{}, and \PRXtwo{} respect immunity to empty updates.
\end{theorem}

There has been a considerable amount of discussion regarding whether immunity to empty updates should 
be considered a desirable property. At the heart of such discussion is the fact that most counter-examples used to
show why some semantics does not obey this property 
involve incoherent/contradictory programs whose coherence/consistency is restored through an empty update.
Two arguments are usually used to support the adequacy of this property, and why coherence/consistency should not be regained no matter what. 

The first argument is conceptual, and related to the difference between \emph{revision} and \emph{update}. Since revision is about incorporating \emph{better} knowledge about some world that did not change, there is an implicit assumption that the initial knowledge was not \emph{perfect} (\emph{complete}, \emph{correct},...), which can somehow be used to justify the removal of inconsistencies/incoherences, even if better explicit knowledge to incorporate is not available, as indicated by an empty program. On the contrary, an update is about incorporating \emph{new} knowledge about a world that changed. The knowledge about the \emph{old} world might be a bad representation of the \emph{new} world, but there is no underlying assumption that it was an imperfect representation of the \emph{old} world, i.e., it could simply be that the \emph{old} world being represented is simply incoherent/inconsistent. And, if nothing changed in that incoherent/inconsistent world, as indicated by the empty program, we should still take it to remain incoherent/inconsistent. 

The second argument is more pragmatic, and tied to the answer-set programming methodology for problem solving \citep{Marek1999,DBLP:journals/amai/Niemela99,Lifschitz1999}, where stable models correspond to solutions to the problem, and the lack of existing stable models simply means that the problem has no solutions. As a concrete example, suppose that a logic program
encodes the well-known n-queens problem, and includes some facts 
encoding the queens that have already been placed on the
chessboard.  Stable models will correspond to possible solutions to the
n-queens problem, indicating where queens should be further placed, given the ones already on the board. To enforce this,
the program would have several rules, acting as integrity constraints,
encoding that no two queens should be placed in the same
column, line or diagonal. Suppose that the initial program 
encodes a situation where two queens have already been placed in the same
column, hence without stable models. It seems clear that an update by an empty
program, encoding that no further queens have been added or removed, should
not change the fact that those two queens are still attacking each other, and no solution can
exist with them on the board, i.e., the program should still have no stable models after the update.

The last property mentioned here --- primacy of new information \citep{Dalal1988} --- is at the heart of every update operator, independently
of the base formalism for which it is defined. It corresponds to postulate \labbr{1} in belief updates, and conveys the fact that 
the models produced by a rule update semantics should conform to the new 
information, captured by the following definition:
\begin{definition}
	[Primacy of New Information]
	Let \S{} be a rule update semantics. We say that
\emph{\S{} respects primacy of new information} if for every DLP $\dprg =
	\seq{\prg_\lia}_{\lia \leq \lng}$ to which \S{}
			is applicable and
every \S-model $\twib$ of $\dprg$, $\twib\models\prg_\lng$.
\end{definition}

Primacy of new information is satisfied by all rule update semantics that we considered so far.

\begin{theorem}
	[Respect for Primacy of New Information]
	\label{thm:ru:primacy}
	Let \X{} be one of \X[D]{}, \X[W]{}, or \X[B]{} and $i \in \set{0, 1, 2}$. The
	rule update semantics \AS{}, \JU{}, \DS{}, \RD{}, \PRZ{}, \PRXi{}, \RVS{},
	and \RVD{} respect primacy of new information.
\end{theorem}

\section{The Third Era --- Semantics-Based Updates}
\label{sec:semanticUpdates}
%!TEX root =  paper.tex

Though useful in practical scenarios
\citep{Alferes2003,Saias2004,Siska2006,Ilic2008,Slota2011a}, it turned out that
most of the syntax-based semantics exhibit some undesirable behaviour. For example, except for
the semantics proposed by \citeN{Alferes2005}, a tautological
update may influence the result under all of these semantics, a behaviour that
is highly undesirable when considering knowledge updates. But, more importantly, the common feature of
all of these semantics is that they make heavy use of the syntactic structure
of programs and rules, while lacking some semantic characterisation that would allow, for example,
some adequate notion of equivalence under updates. This is a well known problem associated with these
semantics \citep{Eiter2002,Leite2003,Slota2012Thesis},
which has prevented a more thorough analysis and understanding of their semantic properties.

\subsection{Operators based on \HT-Models}

Recently, AGM revision was reformulated in the context of logic programming in
a manner analogous to belief revision in classical propositional logic, and
specific revision operators for logic programs were investigated by
\citeN{DBLP:conf/kr/DelgrandeSTW08,DBLP:journals/tocl/DelgrandeSTW13} and by \citeN{Osorio2007}. Central to this novel approach are
\emph{\HT-models}, based on the logic of here-and-there
\citep{Heyting1930,Pearce1997}, which provide a monotonic semantic
characterisation of logic programs that is strictly more expressive than the
answer-set semantics in the sense that it is possible to determine the answer-sets of a 
program from its set of \HT-models, but not vice-versa i.e., there are programs with the same set of answer-sets but different 
sets of \HT-models. Furthermore, two programs have the same set of
\HT-models if and only if they are strongly equivalent \citep{Lifschitz2001},
which means that programs $\prga, \prgb$ with the same set of \HT-models can
be modularly replaced by one another, even in the presence of additional
rules, without affecting the resulting answer-sets.

Indeed, these investigations constitute an important breakthrough in the research of
answer-set program evolution. They change the focus from the syntactic
representation of a program, where not all rules and literal occurrences are
necessarily relevant to its meaning as a whole, to its semantic
content, i.e.,\ to the information that the program is intended to represent.

Subsequently, \citeN{Slota2010b,Slota2013a} followed a similar path, but to tackle the problem of
answer-set program \emph{updates}, instead of \emph{revision} as tackled by
\citeN{DBLP:conf/kr/DelgrandeSTW08,DBLP:journals/tocl/DelgrandeSTW13}. The studied operators are semantic
in their very nature and in line with KM postulates for updates, in contrast with the
traditional syntax-based approaches to rule updates described in the previous section. Others have 
followed this path of defining change operators for logic programs characterised by their \HT-models. For example, \citeN{DBLP:conf/ecai/ZhuangDNS16} introduced a revision operator that adds and/or removes rules from a logic program while minimising the symmetric difference between \HT-models. \citeN{DBLP:journals/tocl/BinnewiesZWS18} introduced partial meet and ensconcement constructions for logic program belief change, which allowed them to define revision and contraction operators.

In this subsection, we first briefly review the revision approach of \citeN{DBLP:conf/kr/DelgrandeSTW08,DBLP:journals/tocl/DelgrandeSTW13} and the update approach of \citeN{Slota2010b,Slota2013a}. Then, we discuss a serious drawback
which extends to all rule update operators that characterise logic programs through their \HT-models. It turns out that both these revision and
the update operators are incompatible with the properties of \emph{support} and \emph{fact update}, which are at the
core of rule updates (cf. Theorems \ref{thm:ru:support} and \ref{thm:ru:fact update}). This is a very important finding as
it guides the research on rule revision and updates away from the semantic approach materialised
in AGM and KM postulates or, alternatively, to the development of semantic characterisations
of programs, richer than \HT-models, that are appropriate for describing their
dynamic behaviour.

Before we proceed, and in order to reformulate some of the postulates for programs under the
\HT-models semantics, we assume some given program conjunction and disjunction
operators $\prgand$, $\prgor$, where each assigns, to each
pair of programs, a program whose set of \HT-models is the intersection and
union, respectively, of the sets of \HT-models of argument programs. The program conjunction
operator may simply return the union of argument programs --- it is the same as the \emph{expansion operator} defined by \citeN{DBLP:conf/kr/DelgrandeSTW08,DBLP:journals/tocl/DelgrandeSTW13} ---
 while the program
disjunction operator can be defined by translating the argument programs into
the logic of here-and-there \citep{Heyting1930,Lukasiewicz1941,Pearce1997},
taking their disjunction, and transforming the resulting formula back into a
logic program (using results by \citeN{Cabalar2007}). 

In this section, we no longer restrict programs and DLPs to those without integrity constraints. 

\subsubsection{Revision based on \HT-Models}

For a rule revision operator
$\ropr$ and programs $\prga$, $\prgb$, $\prgu$, $\prgv$, acording to \citeN{DBLP:conf/kr/DelgrandeSTW08,DBLP:journals/tocl/DelgrandeSTW13}, the KM postulates for belief revision are adapted to answer-set program revision using \emph{\HT-models} as follows:
\begin{enumerate}[labelindent=24pt,labelwidth=12pt,leftmargin=!]
	\renewcommand{\theenumi}{\labpuse{\arabic{enumi}}}
	\renewcommand{\labelenumi}{\theenumi\hfill}
	\setlength{\itemsep}{.1ex}

	\item[\labprse{1}] $\prga \ropr \prgu \entHT \prgu$.
		\label{pstl:prse:1}

	\item[\labprse{2}] If $\modHT{\prga\prgand\prgb}\neq \emptyset$, then $\prga \ropr \prgu \eqHT \prga\prgand\prgb$.
		\label{pstl:prse:2}

	\item[\labprse{3}] If $\modHT{\prgu} \neq \emptyset$,
		then $\modHT{\prga \ropr \prgu} \neq \emptyset$.
		\label{pstl:prse:3}
	\item[\labprse{4}] If $\prga \eqHT \prgb$ and $\prgu \eqHT \prgv$, then $\prga \ropr
		\prgu \eqHT \prgb \ropr \prgv$.
		\label{pstl:prse:4}

	\item[\labprse{5}] $(\prga \ropr \prgu) \prgand \prgv \entHT \prga \ropr (\prgu \prgand
		\prgv)$.
		\label{pstl:prse:5}

	\item[\labprse{6}] If $\modHT{(\prga \ropr \prgu)\prgand\prgv}\neq \emptyset$ then $\prga \ropr (\prgu\prgand\prgv) \entHT (\prga \ropr \prgu)\prgand\prgv$.
		\label{pstl:prse:6}
\end{enumerate}

As discussed by \citeN{DBLP:conf/kr/DelgrandeSTW08,DBLP:journals/tocl/DelgrandeSTW13}, no $\ropr$ can be immune to tautologies or empty updates --- whenever $\prga$ is unsatisfiable and $\prgu$ is tautological or empty, immunity to tautologies and immunity to empty updates conflicts with \labprse{3}. As also discussed by \citeN{DBLP:conf/kr/DelgrandeSTW08,DBLP:journals/tocl/DelgrandeSTW13}, immunity to tautologies and empty updates could have been adopted instead of \labprse{3}.

Analogically to belief revision, a constructive characterisation of rule revision operators
satisfying conditions \labprse{1} -- \labprse{6} 
is based on an order assignment. Since the set of \HT-models of any program $\prga$
must be well-defined, i.e., $\twiab\in\modHT{\prga}$ implies that  $\twibb\in\modHT{\prga}$, not every order assignment characterises a rule revision
operator. \citeN{Slota2010b,Slota2013a} additionally define \emph{well-defined order assignments}
as those that do.

\begin{definition}
	[Rule Revision Operator Characterised by an Order Assignment]
	Let $\ropr$ be a rule revision operator and $\oas$ a preorder assignment over
	the set of all logic programs. We say that $\ropr$ is \emph{characterised by $\oas$} if for all
	programs $\prga$, $\prgu$,
	\[
		\modHT{\prga \ropr \prgu}
		=
		\min \left( \modHT{\prgu}, \pod{\prga} \right)
		.
	\]
	A preorder assignment over over the set of all programs is \emph{well-defined} if
	some rule update operator is characterised by it.
\end{definition}

Similarly as with belief revision, the following definition captures a set of conditions on the assigned orders.

\begin{definition}
	[Faithful Order Assignment Over Programs]
	A preorder assignment $\oas$ over the set of all programs is \emph{faithful} if the following three conditions hold, for all programs $\prga$, $\prgb$:
	\begin{itemize}[labelindent=2pt,labelwidth=6pt,leftmargin=!]
	\item $\text{If }\twia,\twib\in\modHT{\prga}\text{, then }\twia\not\spod{\prga}\twib$;
	\item $\text{If }\twia\in\modHT{\prga}\text{ and }\twib\not\in\modHT{\prga}\text{, then }\twia\spod{\prga}\twib$;
	\item $\text{If }\modHT{\prga}=\modHT{\prgb}\text{, then }\spod{\prga}=\spod{\prgb}$.
	\end{itemize}
\end{definition}

The representation theorem of \citeN{Delgrande2013a} states that operators
characterised by faithful order assignments  over the set of all programs are exactly those that satisfy \labprse{1}--\labprse{6}.

\begin{theorem}
	[\citeN{Delgrande2013a}]
	\label{thm:rrev:representation}
	Let $\ropr$ be a rule revision operator. Then the following conditions are
	equivalent:
	\begin{itemize}[labelindent=2pt,labelwidth=6pt,leftmargin=!]
		\item The operator $\ropr$ satisfies conditions \labprse{1}--\labprse{6}.

		\item The operator $\ropr$ is characterised by a faithful total preorder
			assignment over the set of all programs.
	\end{itemize}
\end{theorem}

\citeN{DBLP:journals/tplp/SchwindI16} provided an alternative constructive characterisation of logic programs revision operators in terms of preorders over interpretations. 
\citeN{DBLP:conf/kr/DelgrandeSTW08,DBLP:journals/tocl/DelgrandeSTW13} present two different revision operators, one analogous to the so-called set-containment based revision proposed by \citeN{Satoh1988}, and the other analogous to the so-called cardinality-based revision proposed by \citeN{Dalal1988}. Here, we recap the latter one, because, unlike the former, it obeys all six postulates \labprse{1}--\labprse{6}.

The operator is based on a notion of \emph{closeness} that is given in terms of cardinality:
\begin{definition}
Let $\stria$ and $\strib$ be sets of either interpretations or three-valued interpretations. Then, 
\[
\sigma_{\vert\vert}(\stria,\strib)=\{\tria\in\stria\vert\exists\trib\in\strib\text{ such that }\forall\tria'\in\stria,\forall\trib'\in\strib,\vert\tria'\div\trib'\vert\not<\vert\tria\div\trib\vert\},
\]
where $\div$ denotes the symmetric difference, extended for three-valued interpretations $\an{\twia, \twib}$ and $\an{\twic, \twid}$ as follows:
\[
\an{\twia, \twib}\div\an{\twic, \twid} = \an{\twia\div\twic, \twib\div\twid},
\]
and
\begin{align*}
\vert\an{\twia, \twib}\vert\leq\vert\an{\twic, \twid}\vert\text{ iff }\vert\twib\vert\leq\vert\twid\vert\text{ and if }\vert\twib\vert = \vert\twid\vert\text{ then }\vert\twia\vert\leq\vert\twic\vert;\\
\vert\an{\twia, \twib}\vert < \vert\an{\twic, \twid}\vert\text{ iff } \vert\an{\twia, \twib}\vert\leq\vert\an{\twic, \twid}\vert\text{ and } \vert\an{\twic,\twid}\vert\not\leq\vert\an{\twia, \twib}\vert.
\end{align*}
\end{definition}
The operator is then defined as follows:
\begin{definition}[Cardinality-based Program Revision \citep{DBLP:conf/kr/DelgrandeSTW08,DBLP:journals/tocl/DelgrandeSTW13}]
Let $\prga$ and $\prgu$ be two logic programs. The cardinality-based revision operator $\ropr_c$ is defined, up to equivalence of its input and output, as a logic program such that:
\begin{align*}
\modHT{\prga\ropr_c\prgb}=&\modHT{\prgb},&\text{if }\modHT{\prga}=\emptyset;\\
\modHT{\prga\ropr_c\prgb}=&\{\an{\twia, \twib}\vert\twib\in\sigma_{\vert\vert}(\modc{\prga},\modc{\prgb}),\twia\subseteq\twib,\\
&\enspace\text{ and if }\twia\subset\twib\text{ then } \an{\twia, \twib}\in\sigma_{\vert\vert}(\modHT{\prga},\modHT{\prgb})
\},&\text{otherwise}.
\end{align*}
\end{definition} 

\begin{theorem}
The cardinality based operator $\ropr_c$ satisfies \labprse{1}--\labprse{6}.
\end{theorem}

\subsubsection{Updates based on \HT-Models}

To adapt the KM postulates for updates to answer-set program
updates, \citeN{Slota2010b,Slota2013a} substitute the notion of a \emph{complete
formula} used in \labbu{7} with the notion of a \emph{basic program}, i.e., a program that either it has a unique \HT-model $\twibb$, or
a pair of \HT-models $\twiab$ and $\twibb$. In the former case, the program
exactly determines the truth values of all atoms --- the atoms in $\twib$ are
true and the remaining atoms are false. In the latter case, the program makes
atoms in $\twia$ true, the atoms in $\twib \setminus \twia$ may either be
undefined or true, as long as they all have the same truth value, and the
remaining atoms are false. The latter case needs to be
allowed in order to make the new postulate applicable to three-valued
interpretations $\twiab$ with $\twia \subsetneq \twib$ because no program has
the single \HT-model $\twiab$. 

Using \emph{\HT-models}, the adaptation of the KM postulates for updates to answer-set program
updates, which, for a rule update operator
$\uopr$ and programs $\prga$, $\prgb$, $\prgu$, $\prgv$ become:
\begin{enumerate}[labelindent=24pt,labelwidth=12pt,leftmargin=!]
	\renewcommand{\theenumi}{\labpuse{\arabic{enumi}}}
	\renewcommand{\labelenumi}{\theenumi\hfill}
	\setlength{\itemsep}{.1ex}

	\item[\labpuse{1}] $\prga \uopr \prgu \entHT \prgu$.
		\label{pstl:puse:1}

	\item[\labpuse{2}] If $\prga \entHT \prgu$, then $\prga \uopr \prgu \eqHT \prga$.
		\label{pstl:puse:2}

	\item[\labpuse{3}] If $\modHT{\prga} \neq \emptyset$ and $\modHT{\prgu} \neq \emptyset$,
		then $\modHT{\prga \uopr \prgu} \neq \emptyset$.
		\label{pstl:puse:3}
	\item[\labpuse{4}] If $\prga \eqHT \prgb$ and $\prgu \eqHT \prgv$, then $\prga \uopr
		\prgu \eqHT \prgb \uopr \prgv$.
		\label{pstl:puse:4}

	\item[\labpuse{5}] $(\prga \uopr \prgu) \prgand \prgv \entHT \prga \uopr (\prgu \prgand
		\prgv)$.
		\label{pstl:puse:5}

	\item[\labpuse{6}] If $\prga \uopr \prgu \entHT \prgv$ and $\prga \uopr \prgv \entHT
		\prgu$, then $\prga \uopr \prgu \eqHT \prga \uopr \prgv$.
		\label{pstl:puse:6}

	\item[\labpuse{7}] If $\prga$ is basic, then $(\prga \uopr \prgu) \prgand (\prga \uopr
		\prgv) \entHT \prga \uopr (\prgu \prgor \prgv)$.
		\label{pstl:puse:7}
	\item[\labpuse{8}] $(\prga \prgor \prgb) \uopr \prgu \eqHT (\prga \uopr \prgu) \prgor
		(\prgb \uopr \prgu)$.
		\label{pstl:puse:8}
\end{enumerate}

Analogically to belief updates, a constructive characterisation of rule update operators
satisfying conditions \labpuse{1} -- \labpuse{8} 
is based on an order assignment, but this time over the set of all
three-valued interpretations $\tris$. Since the set of \HT-models of any program $\prga$
must be well-defined i.e., $\twiab\in\modHT{\prga}$ implies that  $\twibb\in\modHT{\prga}$, not every order assignment characterises a rule update
operator. \citeN{Slota2010b,Slota2013a} additionally define \emph{well-defined order assignments}
as those that do.

\begin{definition}
	[Rule Update Operator Characterised by an Order Assignment]
	Let $\uopr$ be a rule update operator and $\oas$ a preorder assignment over
	$\tris$. We say that $\uopr$ is \emph{characterised by $\oas$} if for all
	programs $\prga$, $\prgu$,
	\[
		\modHT{\prga \uopr \prgu}
		=
		\bigcup_{\tria \in \modsescr{\prga}}
		\min \left( \modHT{\prgu}, \potrd \right)
		.
	\]
	A preorder assignment over $\tris$ is \emph{well-defined} if
	some rule update operator is characterised by it.
\end{definition}

Similarly as with belief updates, order assignments are required to be
faithful, i.e.\ to consider each three-valued interpretation the closest to
itself.

\begin{definition}
	[Faithful Order Assignment Over Three-valued Interpretations]
	A preorder assignment $\oas$ over $\tris$ is \emph{faithful} if for every
	three-valued interpretation $\tria$ the following condition is satisfied:
	\[
		\text{For every } \trib \in \tris \text{ with } \trib \neq \tria
		\text{ it holds that } \tria \spotrd \trib
		.
	\]
\end{definition}

Interestingly, faithful assignments characterise the same class of operators
as the larger class of semi-faithful assignments, defined as follows:

\begin{definition}
	[Semi-Faithful Order Assignment]
	A preorder assignment $\oas$ over $\tris$ is \emph{semi-faithful} if for
	every three-valued interpretation $\tria = \twiab$ the following conditions are
	satisfied (where $\tri^* = \twibb$):
	\begin{align*}
		 &\text{For every }\trib \in \tris \text{ with }\trib \neq \tria\text{  and }\trib \neq
			\tria^*\text{ , either }\tria \spotrd \trib\text{  or }\tria^* \spotrd \trib.\\
		 &\text{If }\tria^* \potrd \tria \text{, then }\tria \potrd \tria^*.
	\end{align*} 
\end{definition}

Preorder assignments need to satisfy one further condition,
related to the well-definedness of sets of \SE-models of every program. It can
naturally be seen as the semantic counterpart of \labpuse{7}.

\begin{definition}
	[Organised Preorder Assignment]
	A preorder assignment $\oas$ is \emph{organised} if for all three-valued
	interpretations $\tria$, $\trib$ and all well-defined sets of three-valued
	interpretations $\stria$, $\strib$ the following condition is satisfied:
	\begin{align*}
		& \text{If }
		\trib \in \min ( \stria, \potrd )
			\cup \min ( \stria, \potrsd )
		\text{ and }
		\trib \in \min ( \strib, \potrd )
			\cup \min ( \strib, \potrsd ), \\
		& \text{then }
		\trib \in \min ( \stria \cup \strib, \potrd )
			\cup \min (\stria \cup \strib, \potrsd ).
	\end{align*}
\end{definition}

Just as with updates in classical logic, the following representation theorem provides a constructive
characterisation of rule update operators satisfying the postulates, making it
possible to define and evaluate any operator satisfying the postulates using
an intuitive construction. 

\begin{theorem}
	[Representation theorem for rule updates \citep{Slota2013a}]
	\label{thm:representation}
	Let $\uopr$ be a rule update operator. The following conditions are
	equivalent:
\begin{enumerate}[labelindent=2pt,labelwidth=6pt,leftmargin=!]
		\item The operator $\uopr$ satisfies conditions \labpuse{1} -- \labpuse{8}.

		\item The operator $\uopr$ is characterised by a semi-faithful and
			organised preorder assignment.

		\item The operator $\uopr$ is characterised by a faithful and organised
			partial order assignment.
	\end{enumerate}
\end{theorem}

One of the benefits of dealing with rule updates on the semantic level is that
semantic properties that are rather difficult to show for syntax-based update
operators are much easier to analyse and prove. As we have seen, one of the most
widespread and counterintuitive side-effects of syntax-based rule update
semantics is that they are sensitive to tautological updates. In case of the
semantic update operators characterised in the previous theorem, such a behaviour is impossible given that the
operators satisfy \labpuse{2} and \labpuse{4}.

 \citeN{Slota2013a} additionally defined a concrete update operator that can be seen as a counterpart to the belief update operator by \citeN{Winslett1988}, which we do not present here.
 
 \subsubsection{Problems with Revision and Update Operators based on \HT-models}
The most important contribution by \citeN{Slota2010b,Slota2013a} is not 
the adaptation of the postulates to answer-set program
updates, the representation theorem, nor even the concrete update operator they defined, but rather the uncovering of a serious
drawback that extends to all rule update and revision operators based on
KM postulates and on \HT-models. In particular, it
turns out that these operators are incompatible with the properties of
\emph{support} and \emph{fact update} which are at the core of logic programming and their updates. 
	
The following theorem
shows that every rule update operator satisfying \labpuse{4}
violates either
\emph{support} or \emph{fact update}, while its proof illustrates why.

\begin{theorem} [\citep{Slota2013a}]
	\label{thm:impossibility}
	A rule update operator that satisfies \labpuse{4} either does not respect
	\emph{support} or it does not respect \emph{fact update}.
\end{theorem}
	\begin{proof}
	Let $\uopr$ be a rule update operator that satisfies \labpuse{4} and
	$\prga$, $\prgb$ and $\prgu$ the following programs:
	\begin{align*}
		\prga:\quad \atma&.
		&
		\prgb:\quad \atma &\lpif \atmb.
		&
		\prgu:\quad \lpnot \atmb. \\
		\atmb&.
		&
		\atmb&.
	\end{align*}
	Since $\prga$ is strongly
	equivalent to $\prgb$, by \labpuse{4} we obtain that $\prga \uopr \prgu$ is
	strongly equivalent to $\prgb \uopr \prgu$. Consequently, $\prga \uopr
	\prgu$ has the same answer-sets as $\prgb \uopr \prgu$. It only remains to
	observe that if $\uopr$ respects fact update, then $\prga \uopr \prgu$ has
	the unique answer-set $\set{\atm}$. But then $\set{\atm}$ is an answer-set
	of $\prgb \uopr \prgu$ in which $\atm$ is unsupported by $\prgb \cup \prgu$.
	Hence $\uopr$ does not respect support.
\end{proof}

So, any answer-set program update operator based on \HT-models and the KM
approach to belief update, as materialised in the fundamental principle
\labpuse{4}, cannot respect two basic and desirable properties: \emph{support} and \emph{fact
update}. This is a major drawback of such operators, severely
diminishing their applicability.

Moreover, the principle \labpuse{4} is also adopted as \labprse{4} for \emph{revision} of
answer-set programs based on \HT-models by \citeN{DBLP:journals/tocl/DelgrandeSTW13}.
This means that Theorem~\ref{thm:impossibility} extends to semantic program
revision operators, such as those defined by \citeN{DBLP:journals/tocl/DelgrandeSTW13}: whenever
support and fact update are expected to be satisfied by a rule revision
operator, it cannot be defined by purely manipulating the sets of \HT-models
of the underlying programs.

One question that suggests itself is whether a weaker version of the principle
\labpuse{4} can be combined with properties such as support and fact update. Its
two immediate weakenings, analogous to the weakenings of  \labbu{4} by
\citeN{Herzig1999}, are as follows:
\begin{enumerate}[labelindent=21pt,labelwidth=12pt,leftmargin=!]
	\renewcommand{\theenumi}{\labpuse{4.\arabic{enumi}}}
	\renewcommand{\labelenumi}{\theenumi\hfill}
	\setlength{\itemsep}{.1ex}

	\item[\labpuse{4.1}] If $\prga \eqHT \prgb$, then $\prga \uopr \prgu \eqHT \prgb \uopr
		\prgu$.
		\label{pstl:puse:4.1}

	\item[\labpuse{4.2}] If $\prgu \eqHT \prgv$, then $\prga \uopr \prgu \eqHT \prga \uopr
		\prgv$.
		\label{pstl:puse:4.2}
\end{enumerate}
In case of \labpuse{4.1}, it is easy to see that the proof of
Theorem~\ref{thm:impossibility} applies in the same way as with \labpuse{4}, so
\labpuse{4.1} is likewise incompatible with support and fact update.

On the other hand, principle \labpuse{4.2}, also referred to as \emph{weak independence
of syntax} (WIS) \citep{Osorio2007}, does not suffer from such severe
limitations. It is, nevertheless, violated by syntax-based rule update
semantics that assign a special meaning to occurrences of default literals in
heads of rules, as illustrated in the following example:
\begin{example}
\label{example_impossible_SE}
	Let the programs $\prga$, $\prgu$ and $\prgv$ be as follows:
	\begin{align*}
		\prga:\quad \atma&.
		&
		\prgu:\quad \lpnot \atma &\lpif \atmb.
		&
		\prgv:\quad \lpnot \atmb &\lpif \atma. \\
		\atmb&.
	\end{align*}
	Since $\prgu$ is strongly equivalent to $\prgv$, \labpuse{4.2} requires that
	$\prga \uopr \prgu$ be strongly equivalent to $\prga \uopr \prgv$. This is
	in contrast with the \JU-, \DS-, and \RD-semantics where a default literal $\lpnot
	\atma$ in the head of a rule indicates that whenever the body of the rule is
	satisfied, there is a \emph{reason for $\atma$ to cease being true}. A
	consequence of this is that an update of $\prga$ by $\prgu$ results in the
	single answer-set $\set{\atmb}$ while an update by $\prgv$ leads to the
	single answer-set $\set{\atma}$.
\end{example}
Thus, when considering the principle \labpuse{4.2}, benefits of the
declarativeness that it brings with it need to be weighed against the loss of
control over the results of updates by rules with default literals in their
heads.

\subsection{Operators based on RE-Models}

The problems identified by \citeN{Slota2010b,Slota2013a} could be mitigated if, instead of \HT-models, a richer semantic
characterisation of logic programs was used. Such a
characterisation would have to be able to distinguish between programs such as
$\prga = \set{\atma., \atmb.}$ and $\prgb = \set{\atma \lpif \atmb., \atmb.}$
because they are expected to behave differently when subject to evolution. And it would have to
distinguish between the rule $\lpnot \atma \lpif \atmb.$ and the rule $\lpnot \atmb \lpif \atma.$, inasmuch as they are expected to have different behaviour, e.g.,
when used to update the program $\prga = \set{\atma., \atmb.}$.

This is precisely the approach taken by \citeN{Slota2012}, who
defined a new monotonic characterisation of rules, dubbed \emph{robust equivalence models}, or \RE-models for short, which is
expressive enough to distinguish between the so-called abolishing rules 
$\lpnot \atma \lpif \atmb.$, $\lpnot \atmb \lpif \atma.$, and $\lpif \atma,\atmb$. Then, they introduced
a generic method for specifying semantic rule update operators in
which
\begin{enumerate}
\item a logic program is viewed as the \emph{set of sets of \RE-models of its
rules},\footnote{This 
view is closely related to the view taken by \emph{base revision operators}
\citep{Gardenfors1992,Hansson1993}.} hence acknowledging rules as the atomic pieces of
knowledge while, at the same time, abstracting away from unimportant differences
between their syntactic forms, focusing on their semantic content;
\item updates are performed by introducing additional interpretations (\emph{exceptions}) to
the sets of \RE-models of rules in the original program. 
\end{enumerate}

Instances of such generic framework
were shown to obey properties such as 
\emph{support}, \emph{fact update}, and \emph{causal rejection}, 
 thus far only obeyed by syntactic approaches. Furthermore, they have a semantic characterisation that ensures several \emph{semantic} properties
 such as replacement of equivalents captured by \labbu{4}, adapted to employ this novel
 semantic characterisation using \RE-models instead of \HT-models.
 One such instance is also shown to provide a semantic characterisation of the \JU-Semantics
for DLPs without local cycles.

The \RE-models and associated notions of equivalence are defined as follows. 

\begin{definition}[\RE-models, \RE-equivalence, and \RR-equivalence]
A three-valued interpretation $\an{\twia, \twib}$ is an \emph{\RE-model} of
a rule $\rl$ if $\twia\ent\rl^\twib$.
The set of all \RE-models of a rule $\rl$ is denoted by $\modre{\rl}$. 
Rules
$\rla$, $\rlb$ are \emph{\RE-equivalent} whenever $\modre{\rla} = \modre{\rlb}$. Programs
$\prga$ and $\prgb$ are \emph{\RE-equivalent}, denoted by $\prga\eqRE\prgb$, whenever $\modre{\prga} = \modre{\prgb}$.
The set of sets of
\RE-models of rules inside a program $\prg$ is denoted by $\modrer{\prg} = \Set{
\modre{\rl} | \rl \in \prg }$, and is used as the basis for the notion of \emph{robust rule equivalence}, or \emph{\RR-equivalence} for short.
Programs $\prga$ and $\prgb$ and \emph{\RE-equivalent}, denoted by $\prga \eqRR \prgb$,
 whenever
$\modrer{\prga^\ctau} = \modrer{\prgb^\ctau}$, where, given a program $\prga$, $\prga^\ctau = \prga \cup \set{\ctau}$, where
$\ctau$ is the \emph{canonical
	tautology}, i.e., the rule $\atm_\ctau \lpif \atm_\ctau$ given a fixed atom $\atm_\ctau$ from $\lang$.
\end{definition}
Interestingly, the affinity between \HT-models and stable models is fully retained by
\RE-models: an interpretation $\twib$ is a stable model of a program $\prg$ if and only
	if $\an{\twib, \twib} \in \modre{\prg}$ and for all $\twia \subsetneq
	\twib$, $\an{\twia, \twib} \notin \modre{\prg}$.
Furthermore, unlike with \HT-models, any set of three-valued interpretations can be
represented by a program using \RE-models \citep{Slota2012,Slota2012Thesis}.

The formalisation of the idea of viewing updates as introducing additional interpretations --- \emph{exceptions} --- to
the sets of \RE-models of rules in the original program is straightforward: an exception-driven update operator is
characterised by an \emph{exception function} \te{} that takes three inputs:
the set of \RE-models $\modre{\rl}$ of a rule $\rl \in \prga$ and the semantic
characterisations, $\modrer{\prga}$ and $\modrer{\prgu}$, of the original and
updating programs. It then returns the three-valued interpretations that are
to be introduced as exceptions to $\rl$, so the characterisation of the
updated program contains the augmented set of \RE-models,
\begin{equation}
	\label{eq:re models of rule after update}
	\modre{\rl} \cup \e \left(
		\modre{\rl}, \modrer{\prga}, \modrer{\prgu}
	\right).
\end{equation}
A rule update operator $\uopr$ is \emph{exception-driven} if for some
	exception function $\e$, $\modrer{\prga \uope \prgu}$ is equal to
\begin{equation}
	\label{eq:characterisation after update}
	\Set{
		\modre{\rl} \cup \e \left(
			\modre{\rl}, \modrer{\prga}, \modrer{\prgu}
		\right)
		|
		\rl \in \prga
	}
	\cup \modrer{\prgu},
\end{equation}
for all programs  $\prga$ and $\prgu$. In that case we also say that $\uopr$ is \emph{\te-driven}.
In words, the set of \RE-models of each rule $\rl$ from $\prga$ is
augmented with the respective exceptions while the sets of \RE-models of rules
from $\prgu$ are kept untouched.
Note that since a set of three-valued interpretations may have different syntactic representations 
as a program using \RE-models, for each exception function \te{} there is a whole class of
\te-driven rule update operators that differ in the syntactic representations
of the sets of \RE-models in \eqref{eq:characterisation after update}.\footnote{
To deal with sets of  \RE-models defined in \eqref{eq:characterisation after update} which do not correspond to a single rule, 
but rather to a set of rules, \citeN{Slota2012} consider the so-called 
\emph{rule-bases}, which can be rules or programs representing the \RE-models defined in  \eqref{eq:re
models of rule after update}, and treat them as \emph{atomic} pieces of
information. Here we omit the technical aspects of this issue and, like \citeN{Slota2012}, dub them \emph{rules}.}

\citeN{Slota2012} further investigated a constrained class of exception functions, which
they dubbed \emph{simple exception functions}, characterised by the fact 
that they produce (local) exceptions based on conflicts between pairs of rules, one from
the original and one from the updating program, while ignoring the context in
which these rules are situated, i.e., the other rules in the programs. Formally, an exception function $\e$ is 
\emph{simple} if for all $\stria \subseteq
	\tris$ and $\sstria, \sstrib \subseteq \pws{\tris}$,
	\[
		\e(\stria, \sstria, \sstrib)
			= \textstyle \bigcup_{\strib \in \sstrib} \er(\stria, \strib),
	\]
	where $\er : \pws{\tris} \times \pws{\tris} \rightarrow \pws{\tris}$ is a
	\emph{local exception function}. 
	
Despite their local nature, particular simple exception
functions generate rule update operators that satisfy the syntactic properties
of rule update semantics discussed before.
We will recap two 
simple exception functions proposed by \citeN{Slota2012}, dubbed \terone{} and \tertwo{}, inspired by rule update semantics based on causal rejection, one of them closely related to the
\JU-semantics.

Since rule update semantics based on causal rejection make use of the concepts of \emph{conflicting rules} (c.f. Def. \ref{def:conflicting_rules}) and \emph{rejected rules} (c.f. Def. \ref{def:ru:ju} for the case of the \JU-semantics), which rely on
rule syntax to which an exception function has no direct access, the semantic counterparts to these concepts
were first defined. 

Towards the definition of conflicting sets of \RE-models, two preparatory concepts are required.

First, a \emph{truth value substitution} is defined as follows: Given an
interpretation $\twib$, an atom $\atma$ and a truth value $\val \in \set{\tr,
\un, \fa}$, by $\twib[\val/\atma]$ we denote the three-valued interpretation
$\tri$ such that $\tri(\atma) = \val$ and $\tri(\atmb) = \twib(\atmb)$ for all
atoms $\atmb \neq \atma$.

This enables the introduction of the main concept needed for defining a conflict
between two sets of three-valued interpretations. Given a set of three-valued
interpretations $\stri$, an atom $\atm$, a truth value $\val_0$ and a
two-valued interpretation $\twib$, we say that \emph{$\stri$ forces $\atm$ to
have the truth value $\val_0$ w.r.t.\ $\twib$}, denoted by $\stri^\twib(\atm)
= \val_0$, if
\[
	\twib[\val/\atm] \in \stri \text{ if and only if } \val = \val_0 .
\]
In other words, the three-valued interpretation $\twib[\val_0/\atm]$ must be
the unique member of $\stri$ that either coincides with $\twib$ or differs
from it only in the truth value of $\atm$. Note that $\stri^\twib(\atm)$ stays
undefined in case no $\val_0$ with the above property exists.

Two sets of three-valued interpretations $\stria$, $\strib$ are \emph{in
conflict on atom $\atm$ w.r.t.\ $\twib$}, denoted by $\stria
\confl{\twib}{\atm} \strib$, if both $\stria^\twib(\atm)$ and
$\strib^\twib(\atm)$ are defined and $\stria^\twib(\atm) \neq
\strib^\twib(\atm)$. The following example illustrates all these concepts.

\begin{example}
	\label{ex:semantic conflict}
	Consider rules $\rl_0 = (\atma.)$, $\rl_1 = (\lpnot \atma \lpif \lpnot
	\atmb.)$ with the respective sets of \RE-models\footnote{We sometimes omit
	the usual set notation when we write interpretations. For example, instead
	of $\set{\atma, \atmb}$ we write $\atma\atmb$.}
	\begin{align*}
		\stri_0 &= \Set{
				\tpl{\atma, \atma}, \tpl{\atma, \atma\atmb},
				\tpl{\atma\atmb, \atma\atmb}
			}
			, \\
		\stri_1 &= \Set{
			\tpl{\emptyset, \emptyset}, \tpl{\emptyset, \atmb}, \tpl{\atmb, \atmb},
			\tpl{\emptyset, \atma\atmb}, \tpl{\atma, \atma\atmb},
			\tpl{\atmb, \atma\atmb}, \tpl{\atma\atmb, \atma\atmb}
		}.
	\end{align*}
	Intuitively, $\stri_0$ forces $\atma$ to $\tr$ w.r.t.\ all interpretations
	and $\rl_1$ forces $\atma$ to $\fa$ w.r.t.\ interpretations in which $\atmb$
	is false. Formally it follows that $\stri_0^\emptyset(\atma) = \tr$ because
	$\tpl{\atma, \atma}$ belongs to $\stri_0$ and neither $\tpl{\emptyset,
	\atma}$ nor $\tpl{\emptyset, \emptyset}$ belongs to $\stri_0$. Similarly, it
	follows that $\stri_1^\emptyset(\atma) = \fa$. Hence $\stri_0
	\confl{\emptyset}{\atma} \stri_1$. Using similar arguments we can conclude
	that $\stri_0 \confl{\atma}{\atma} \stri_1$. However, it does not hold that
	$\stri_0 \confl{\atma\atmb}{\atma} \stri_1$ because
	$\stri_1^{\atma\atmb}(\atma)$ is undefined.
\end{example}

We are now ready to introduce the local exception function \terone{}.

\begin{definition}[Local Exception Function \terone{}]
	The local exception function \terone{} is for all $\stria, \strib \subseteq
	\tris$ defined as
	\[
		\erone(\stria, \strib) = \Set{
			\tpl{\twia, \twib} \in \tris
			|
			\exists \atm : \stria \confl{\twib}{\atm} \strib
		} .
	\]
\end{definition}

Thus if there is a conflict on some atom w.r.t.\ $\twib$, the exceptions
introduced by \terone{} are of the form $\tpl{\twia, \twib}$ where $\twia$ can
be an arbitrary subset of $\twib$. This means that \terone{} introduces as
exceptions all three-valued interpretations that preserve false atoms from
$\twib$ while the atoms that are true in $\twib$ may be either true or
undefined. This is somewhat related to the definition of a stable model where
the default assumptions (false atoms) are fixed while the necessary truth of
the remaining atoms is checked against the rules of the program. The syntactic
properties of \terone-driven operators are as follows.

\begin{theorem}[Syntactic Properties of \terone{}  \citep{Slota2012}]
	\label{thm:syntactic properties:erone}
	Every \terone-driven rule update operator respects support and fact update.
	Furthermore, it also respects causal rejection and acyclic justified update
	w.r.t.\ DLPs of length at most two.
\end{theorem}

This means that \terone-driven rule update operators enjoy a combination of
desirable syntactic properties that operators based on \SE-models cannot
(cf.\ Theorem~\ref{thm:impossibility}). However, these operators diverge
from causal rejection, even on acyclic DLPs, when more than one update is
performed.

\begin{example}
	\label{ex:terone iterated}
	Consider again the rules $\rl_0$, $\rl_1$ and their sets of \RE-models
	$\stri_0$, $\stri_1$ from Example~\ref{ex:semantic conflict} and some
	\terone-driven rule update operator $\uopr$. Then, $\modrer{\set{\rl_0} \uopr
	\set{\rl_1}}$ will contain two elements: $\stri_0'$ and $\stri_1$, where
	\[
		\stri_0'
		= \stri_0 \cup \erone(\stri_0, \stri_1)
		= \stri_0 \cup \set{\tpl{\emptyset, \emptyset}, \tpl{\emptyset, \atma}}
		.
	\]
	An additional update by the fact $\set{\atmb.}$ then leads to the
	characterisation
	\[
		\Modrer{\biguopr \seq{\set{\rl_0}, \set{\rl_1}, \set{\atmb.}}},
	\]
	which contains three elements: $\stri_0''$, $\stri_1$ and $\stri_2$, where
	\[
		\stri_0'' = \stri_0'
			\cup \set{\tpl{\emptyset, \atmb}, \tpl{\atmb, \atmb}},
	\]
	and $\stri_2$ is the set of \RE-models of $(\atmb.)$.

	Furthermore, due to the relationship between \RE-models and stable models, the interpretation $\twib
	= \set{\atmb}$ is a stable model of $\biguopr \seq{\set{\rl_0},
	\set{\rl_1}, \set{\atmb.}}$ because $\tpl{\atmb, \atmb}$ belongs to all sets
	of models in the set of sets of models $\modrer{\biguopr \seq{\set{\rl_0},
	\set{\rl_1}, \set{\atmb.}}}$ and $\tpl{\emptyset, \atmb}$ does not belong to
	$\stri_2$. However, $\twib$ does not respect causal rejection and it is not
	a \JU-model of $\tpl{\set{\rl_0}, \set{\rl_1}, \set{\atmb.}}$.
\end{example}

This shortcoming of \terone{} was overcome by \citeN{Slota2012} as follows:

\begin{definition}[Local Exception Functions \tertwo{}]
	The local exception functions \tertwo{} is for all $\stria,
	\strib \subseteq \tris$ defined as
	\begin{align*}
		\ertwo(\stria, \strib) &= \Set{
			\tpl{\twia, \twic} \in \tris \mid
				\exists \twib \, \exists \atm : \stria \confl{\twib}{\atm} \strib
				\land \twia \subseteq \twib \subseteq \twic
				\land (\atm \in \twic \setminus \twia \mlthen \twic = \twib)
			} .
	\end{align*}
\end{definition}

The function \tertwo{} introduces more exceptions than
\terone{}. A conflict on $\atm$ w.r.t. $\twib$ leads to the introduction of
interpretations in which atoms either maintain the truth value they had in
$\twib$, or they become undefined. They must also satisfy an extra condition:
when $\atm$ becomes undefined, no other atom may pass from false to undefined.
This leads to operators that satisfy all syntactic properties.

\begin{theorem}[Syntactic Properties of \tertwo{} \citep{Slota2012}]
	Let $\uopr$ be a \tertwo-driven rule update operator. Then
	$\uopr$ respects support, language conservation, fact update, causal rejection, acyclic justified
	update, immunity to tautological and empty updates, and primacy of new information.
\end{theorem}

Under \RR \ equivalence, the postulate that requires update operators to be syntax indepent can be defined as follows, for a rule update operator $\uopr$ and programs $\prga$, $\prgb$, $\prgu$, and $\prgv$:

\labpurr{4}
				 If $\prga \eqRR \prgb$ and $\prgu \eqRR \prgv$, then
					$\prga \uopr \prgu \eqRR \prgb \uopr \prgv$.
					
\begin{theorem}[Syntax Independence of \tertwo{} \citep{Slota2012}]
Let $\uopr$ be a \tertwo-driven rule update operator. Then, $\uopr$ respects \labpurr{4}.
\end{theorem}

It is worth noting that \tertwo-driven
operators are very closely related to the \JU-semantics, even on
programs with cycles. They diverge from it only on programs with tautologies.

\begin{theorem}[\citep{Slota2012}]
	\label{prop:exctwo vs ju}
	Let $\dprg$ be a DLP, $\twib$ an interpretation and $\uopr$ a
	\tertwo-driven rule update operator. Then,
	\begin{itemize}[labelindent=2pt,labelwidth=6pt,leftmargin=!]
		\item $\moduopr{\dprg} \subseteq \modju{\dprg}$ and
		\item if $\all{\dprg}$ contains no tautologies, then $\modju{\dprg}
			\subseteq \moduopr{\dprg}$.
	\end{itemize}
\end{theorem}

This means that up to the case of tautologies, \tertwo{} can be seen as semantic characterisation of the justified update
semantics: it leads to stable models that, typically, coincide with justified
update models.

\begin{example}
	Consider again the program $\prga$ from the example in the introduction, which contains the rules
	\begin{align*}
		\prest{goHome} \lpif \lpnot \prest{money}.&
		&\prest{goRestaurant} \lpif \prest{money}.&
		&\prest{money}.
	\end{align*}
		and its update $\prgu$ with the rules
	\begin{align*}
		\lpnot \prest{money} &\lpif \prest{robbed}.
		& \prest{robbed} &.
	\end{align*}
Unsurprisingly, \tertwo{} would only introduce exceptions to the third rule of $\prga$. A
\tertwo-driven rule update operator $\uopr$ would produce,
as the result of $\prga\uopr\prgu$, a program \RR-equivalent to:
	\begin{align*}
		\prest{goHome} &\lpif \lpnot \prest{money}. \\
		\prest{goRestaurant} &\lpif \prest{money}.\\
		\prest{money} &\lpif \lpnot \prest{robbed}. \\
		\lpnot \prest{money} &\lpif \prest{robbed}.\\
		\prest{robbed} &.
	\end{align*}
\end{example}

\begin{example}
Going back to Example \ref{example_impossible_SE} where the programs $\prga$, $\prgu$, and $\prgv$ were as follows:
	\begin{align*}
		\prga:\quad \atma&.
		&
		\prgu:\quad \lpnot \atma &\lpif \atmb.
		&
		\prgv:\quad \lpnot \atmb &\lpif \atma. \\
		\atmb&.
	\end{align*}

A
\tertwo-driven rule update operator $\uopr$ would produce, up to \RR-equivalence, the following results, which respect both \emph{support} and \emph{fact update}:

\noindent\begin{minipage}{.5\linewidth}
\begin{align*}
		\prga \uopr \prgu:\quad \atma &\lpif \lpnot \atmb.\\
		                            \quad \atmb&.\\
		                            \quad \lpnot\atma &\lpif \atmb.
	\end{align*}
\end{minipage}%
\begin{minipage}{.5\linewidth}
\begin{align*}
		\prga \uopr \prgv:\quad \atma&.\\
		                            \quad \atmb&\lpif \lpnot \atma\\
		                            \quad \lpnot\atmb &\lpif \atma.
	\end{align*}
\end{minipage}
\end{example}

The new monotonic characterisation of rules --- \RE-models --- and
the generic method for specifying semantic rule update operators in
which a logic program is viewed as the \emph{set of sets of \RE-models of its
rules} and updates are performed by introducing additional interpretations to
the sets of \RE-models of rules in the original program, allowed the definition of 
concrete update operators that enjoy a combination of syntactic as well as semantic properties that had
never been reconciled before. 

Acknowledging rules as first-class objects by viewing a program as the set of sets of
their models essentially amounts to adopting the view used by \emph{base revision operators} \citep{Gardenfors1992,Hansson1993} where a theory is composed of a set of formulas, each considered an atomic piece of knowledge that could be falsified by an update. This view has also been more recently adopted within the context of 
description logic updates \citep{Liu2006,Giacomo2009,Calvanese2010,Lenzerini2011}. 

Nevertheless, departing from \HT-models and using \RE-models instead may raise a few eyebrows. After all, characterising strong equivalence between programs through \HT-models is one of the great landmarks in the history of answer-set programming. However, strong equivalence characterised by \HT-models only considers the union of rules, while other kinds of operations must also be considered when dealing with belief change such as \emph{updates} and \emph{revisions}, e.g., falsifying an atom that was previously true. Just as the rule $\atma \lpif \atmb.$ specifies $\atmb$ as a justification for $\atma$, the rule $\lpnot \atma \lpif \atmb.$ should be seen as specifying $\atmb$ as a justification for $\lpnot \atma$ and the rule $\lpnot \atmb \lpif \atma.$ as specifying $\atma$ as a justification for $\lpnot \atmb$. Whereas the latter two rules are \HT-equivalent --- each can be modularly replaced by the other, the remaining rules staying the same --- they provide different justifications in a dynamic setting, e.g., when used to update the program $\prga = \set{\atma., \atmb}$: the former provides a justification to make $\atma$ false while the latter a justification to make $\atmb$ false. Distinguishing between these two rules, and between them and the integrity constraint $\lpif \atma,\atmb.$, is precisely the difference between \HT-models and \RE-models.

Other properties that are obeyed by 
a \tertwo-driven rule update operator $\uopr$ include, for example, 
\emph{initialisation} ($\emptyset \uopr \prgu \equiv \prgu$), 
\emph{non-interference} (If $\prgu$, $\prgv$ are over disjoint alphabets,
					then $(\prga \uopr \prgu) \uopr \prgv
						\equiv (\prga \uopr \prgv) \uopr \prgu$),
\emph{absorption} ($(\prga \uopr \prgu) \uopr \prgu \equiv \prga \uopr \prgu$), 
\emph{augmentation} (If $\prgu \subseteq \prgv$, then
					$(\prga \uopr \prgu) \uopr \prgv \equiv \prga \uopr \prgv$), among others.
A more thorough discussion on \tertwo-driven rule update operators' properties as well as on the reasons why they fail other 
semantic properties drawn from the KM postulates are given by \citeN{Slota2012}.

Despite this success, a closer inspection shows that 
 \RR-equivalence might still be
slightly too strong for characterising updates, because programs such
as $\set{\atma.}$ and $\set{\atma., \atma \lpif \atmb.}$ are not considered
 \RR-equivalent even though we expect the same behaviour from them when they are
updated. A notion of program equivalence that
is weaker than \RR-equivalence but stronger than \RE-equivalence
so that both \labbu{4} and properties such as $\prga \cup \prgu \ent \prga \uopr \prgu$ can
be achieved under a single notion of program equivalence still needs to be found.
\citeN{Slota2011} speculate that this could be solved by
adopting a weaker equivalence --- \SMR-equivalence --- which discards
rules whose set of models are supersets of the set of models of other rules. However, such equivalence is too weak
since, when instantiated with \RE-models, programs such as $\set{\lpnot
\atmb.}$ and $\set{\lpnot \atmb., \atma \lpif \atmb.}$ are \SMR-equivalent
although, when updated by $\set{\atmb.}$, different results are expected for
each of them.

\section{State of Affairs --- Comments and Outlook}
%!TEX root =  paper.tex

\cite{McCarthy98} described \emph{elaboration tolerance} as ``\emph{the ability to accept changes to a person's or computer program's representation of facts about a subject without having to start all over}", arguing that human-level AI will require representation formalisms that are elaboration tolerant. Representing knowledge as a DLP --- a sequence of logic programs --- equipped with automated mechanisms to deal with overlapping, possibly conflicting information, as those provided by the rule update semantics surveyed in this paper, constitutes one significant step towards equipping logic programming with elaboration tolerance. It allows for the kind of incremental specification proposed by McCarthy where one can focus on specifying what is new or what has changed, adding it to the end of the sequence, without having to rework a new encompassing representation.

One of the most important lessons learnt so far is that the syntactic nature of answer-set programming cannot be 
ignored when programs are subjected to a belief change operation such as an update, or even a revision.
Whereas this was at the heart of the rejection of model-based updates, which led to the myriad of approaches developed
during the second era --- syntax-based updates --- it was partially forgotten at the beginning of the third era --- semantics-based updates ---
with the promise that characterising a program through its \HT-models would move us away from having to 
rely on the program's syntax. Theorem \ref{thm:impossibility} killed this hope by showing that such a characterisation of programs 
based on \HT-models simply cannot be used if fundamental properties such as \emph{support} and 
\emph{fact update} are required. And these are indeed two fundamental properties. \emph{Fact update} is a very
straightforward, simple and rather undisputed property, which corresponds to the simple unconditional change operations --- insertion and deletion --- performed on 
extensional (relational) databases, e.g., corresponding to the SQL commands \texttt{INSERT/DELETE} \prest{fact} \texttt{FROM P}. 
The second property, \emph{support}, is related to the fact that implication in rules plays a significantly different role than material implication in classical
logic, e.g., not allowing for the contrapositive, instead being tied to the intuitionistic logic of here-and-there \citep{Heyting1930,Pearce1997} where the notion of truth is tied to the notion of proof/justification, somehow provided by rules from their bodies to their heads. Accordingly, unsupported atoms are usually not accepted in logic programming.

The negative result encoded by Theorem~\ref{thm:impossibility} should certainly make us look back
to the so-called syntactical approaches through a different set of lenses, less critical of the fact that
its syntactic features play a central role, and see whether they indeed provide a viable alternative, 
such as the \RD-semantics, which is the only one immune to tautological updates while maintaining
all other syntactical properties discussed here. But at the same time, Theorem~\ref{thm:impossibility} should not
drive us away from the pursuit of a semantical characterisation of updates that somehow reconciles the syntactical
properties such as \emph{support}, \emph{fact update}, and \emph{causal rejection} with the semantic properties encoded 
in both the AGM and KM postulates. 

Acknowledging the importance of these syntactical properties, \citeN{DBLP:journals/tocl/BinnewiesZWS18} also adopted \HT-models as the underlying semantic characterisation of logic programs to introduce partial meet and ensconcement constructions for logic program belief change, which allowed them to define syntax-preserving operators that obey \emph{support}.

The introduction of the \emph{abstract exception-driven
update} abstract framework  \citep{Slota2012a} seems to be another important step in the direction of reconciling the syntactical nature of logic programming with the semantic properties encoded in the AGM family of postulates. On the one hand, it served as a framework to
capture updates of logic programs, namely the \JU-semantics, reconciling semantical as well as syntactical properties. It did so by using \RE-models to characterise rules, and viewing a program as the \emph{set of sets of the \RE-models of its rules}. This way, it acknowledged rules as the atomic pieces of knowledge while, at the same time, abstracting away from unimportant differences
between their syntactic forms, focusing on their semantic content. On the other hand, this abstract framework was shown to also capture several update
operators used for ontology updates, such as the model-based
Winslett's operator, or the formula-based WIDTIO and \emph{bold} operators
\citep{Liu2006,Giacomo2009,Calvanese2010,Lenzerini2011}. 

The \emph{exception-driven update} framework also
seems to provide a promising vehicle
to reconcile updates in classical logic with updates of logic programs, opening 
up a promising avenue
to investigate updates of hybrid knowledge bases composed of rules and ontologies \citep{Slota2012Thesis,DBLP:journals/ai/SlotaLS15}.
Additionally, the logic programming instantiation of the exception-driven update framework
shed new light into the long lasting problem of \emph{state condensing}\footnote{\emph{State condensing} is the problem of finding
a single logic program that faithfully represents
a sequence of logic programs (DLP), i.e., that (i) is written in the
same alphabet, (ii) has the same set of stable models, and (iii) is equivalent to the
sequence of programs when subject to further updates.}, which was 
solved by \citeN{Slota2013} for the \JU{} and \AS{} semantics, by resorting to more expressive
classes of answer-set programs, namely nested and disjunctive. 

The historical account of the research on updating logic programs found in this paper is almost entirely focused on the declarative side of the problem, leaving 
out most procedural and computational aspects. Our choice to follow this path was somehow grounded on the fact that over the years the procedural
and computational aspects have played a secondary role in this line of research. As far as we know, there are currently no efficient native implementations available to compute the \S-models for any of the semantics \S\ presented in this paper. The absence of efficient native implementations is more important given the known negative results regarding \emph{state condensing}, which in general prevent the direct usage of existing efficient solvers, since it is in general not possible to construct a single logic program that corresponds to the result of the updates encoded in a DLP, written in the same language $\lang$ as the DLP. Several authors have addressed the procedural and computational aspects of updating logic programs through the definition of general transformations that convert any DLP into a single logic program --- written in a language that extends to the original language $\lang$ with new propositional symbols to allow the representation of the \emph{temporal} aspect of the DLP, and other notions such as the rejection of rules --- whose stable models, restricted to the original language $\lang$, correspond to the \S-models according to the rule update semantics \S{} in question. Each of the semantics based on causal rejection has one such corresponding transformation. With these transformations, the \S-models of the DLP can be computed through the resulting logic program using existing efficient solvers such as \emph{clasp} \citep{DBLP:journals/aicom/GebserKKOSS11} and \emph{dlv} \citep{DBLP:journals/tocl/LeonePFEGPS06}, even though the syntactical overhead introduced by the transformations could result in a significant computational cost. \citeN{Delgrande2007}, on the other hand, define their semantics through a transformation into a prioritised logic program, which doesn't need the extension of the language $\lang$, but for which there are no efficient native implementations, as far as we know. Hence, the resulting prioritised logic program would have to be further converted into a logic program to enable the use of existing efficient solvers, with a similar syntactical overhead as the transformations mentioned above. Other semantics either do not mention procedural and computational aspects, or provide transformations that only deal with DLPs of length two, whose result cannot be further updated, as discussed earlier in this paper. All in all, whereas the known results regarding the procedural and computational aspects of many of the semantics presented in this paper are enough to allow the development of prototypical implementations --- some of which have actually been created by the authors of the original semantics, even if no longer easily available --- more research is needed to develop native efficient implementations that could support large applications.

Despite the progress towards a full understanding of updates of logic programs, many open 
questions remain, whose difficulty in answering seems to always be tied to the tension created by
the syntactical nature of logic programs, and the wish for a semantical characterisation
of belief change operations. We discuss some of the most important ones below.

\emph{Can other syntactical rule update semantics be captured by the \emph{exception-driven update} framework?} 
The need to detect
non-tautological irrelevant updates
\citep{Alferes2005,Sefranek2006,Sefranek2011} to capture, e.g., the \RD-semantics poses a number of challenges. For instance, simple exception functions
based on local exceptions such as \terthree\ cannot distinguish an update of $\set{\atma.}$ by
$\prgu = \set{\lpnot \atma \lpif \lpnot \atmb., \lpnot \atmb \lpif \lpnot
\atma.}$, where it is plausible to introduce the exception $\an{\emptyset,
\emptyset}$, from an update of $\set{\atma., \atmb.}$ by $\prgu$, where such
an exception should not be introduced due to the cyclic dependency of
justifications to reject $\atma.$ and $\atmb.$ In such situations,
context-aware functions need to be used. Such functions would also have the
potential of satisfying properties such as \labbu{3} and associativity ($\prga \uopr (\prgu \uopr \prgv) \equiv (\prga \uopr \prgu) \uopr \prgv$).

\emph{Can we characterise rule update semantics through 
binary operators on some class of programs over the same alphabet so that they can be iterated, hence addressing the state condensing problem?}
Whereas answering the previous question on this list might certainly provide hints on how to characterise rule update semantics through 
binary operators,
it may well be a problem without solution for some 
of the existing semantics, even resorting to more expressive classes of logic programs.
This is probably the 
case with the \PRZ{} and the \RVS{} semantics because of how they are defined, which involves the construction of
possibly more than one program, using maximality criteria. Given its properties and relationship with the \JU{} and \AS{} semantics, for which a solution already exists, 
perhaps the bigger reward lies in addressing this question for the \RD{} semantics.

 \emph{What is the appropriate logic corresponding to the \RE-models?} Just as the logic of here-and-there \citep{Heyting1930,Pearce1997}, a nonclassical logic extending intuitionistic logic, 
served as a basis for a logical characterisation of answer-set programming,
finding the nonclassical logic that subsumes \RE-models and provides an adequate basis to characterise the dynamics of logic programs would be most useful. The
difficulties in extending even \RE-models to more general classes of logic programs hints on the difficulty of this task. 

\emph{Can a characterisation based on \RE-models bring added value when used to describe logic programs that undergo other belief change operations such as {revision}, {contraction}, {erasure}, {forgetting} and {ensconcement}?} Ultimately, the answer to this question boils down to whether the rule $\lpnot \atma \lpif \atmb.$, the rule $\lpnot \atmb \lpif \atma.$, and the integrity constraint $\lpif \atma,\atmb.$ should be considered equivalent, or not, within the belief change operation in question. In the context of updates of logic programs, as we have seen throughout this paper, some semantics distinguish them, as for example those based on causal rejection, where updating the program $\set{\atma.,\atmb.}$ with $\lpnot \atma \lpif \atmb.$ results in the model $\set{\atmb}$ while updating it with $\lpnot \atmb \lpif \atma.$ would result in the model $\set{\atma}$. If this distinction makes sense in other belief change operations --- for example, one might conceive a revision operator with a similar behaviour --- then a characterisation based on \RE-models might be more suitable than using, e.g., \HT-models. 

\emph{How can the syntax-based updates be extended to deal with disjunctive logic programs?} Interestingly, updating disjunctive logic programs under the answer-set semantics has received very little attention so far. Whereas extending some of the syntax-based approaches seems straightforward --- in fact, the \RVS-semantics was defined by \citeN{Sakama2003} for disjunctive logic programs --- dealing with semantics such as those based on the causal rejection principle might prove more difficult. For example, conflicts that provide a cause for rejecting rules would no longer be restricted to pairs of rules.

\emph{How to precisely characterise the related though distinct belief change operations of revision and update within the context of answer-set programming?} At an abstract level, the difference between these two forms of belief change rests on whether we are acquiring better information about a static world (revision) or whether we are acquiring newer information about a changing world (update). The influence of such difference when dealing with belief change in answer-set programming has been limited and, to some extent, restricted to the extent to which one seeks to deal with inconsistencies/contradictions: whereas in revision they are removed at any cost, even as the result of an empty/tautological belief change operation, in updates such an empty/tautological operation should have no effect. More is needed in order to better understand and characterise what revision and update have in common, and in difference, when taken within the context of answer-set programming.

 \emph{Does it make sense to consider other causes for rejection within causal-rejection based semantics?} The answer to this question is, to some extent, related to the previous question, inasmuch as it relates to finding other conflicts that, if not resolved, could result in a contradiction/incoherence which should be avoided in the context of an update. But other avenues for extending the causes for rejection exist such as, for example, explicitly declaring certain pairs of otherwise unrelated atoms as being in conflict.

\ 
\newline

All in all, the quest for the \emph{perfect} approach to updating answer-set programs has been quite fruitful.
We are not there yet --- and we may never be --- but we are certainly much closer.

 % Acknowledgments
\section*{Acknowledgments}
The authors would like to thank Jos{\'e} Alferes, Martin Bal{\'{a}}z, Federico Banti, Antonio Brogi, Martin Homola, Lu{\'i}s Moniz Pereira, Halina Przymusinska, Teodor C. Przymusinski, and Theresa Swift, with whom they 
worked on the topic of this paper over the years, as well as Ricardo Gon\c calves and Matthias Knorr for valuable comments on an earlier draft of this paper. 
The authors would also like to thank the anonymous reviewers for their insightful comments and suggestions, which greatly helped us improve this paper.
The authors were partially supported by Funda\c{c}\~ao para a Ci\^encia e Tecnologia through projects FORGET (PTDC/CCI-INF/32219/2017) and RIVER (PTDC/CCI-COM/30952/2017), and strategic project NOVA LINCS (UIDB/04516/2020).

% Bibliography
\bibliographystyle{acmtrans}
\bibliography{bibliography}

\begin{thebibliography}{}

\bibitem[\protect\citeauthoryear{Alchourr{\'o}n, G{\"a}rdenfors, and
  Makinson}{Alchourr{\'o}n et~al\mbox{.}}{1985}]{Alchourron1985}
{\sc Alchourr{\'o}n, C.~E.}, {\sc G{\"a}rdenfors, P.}, {\sc and} {\sc Makinson,
  D.} 1985.
\newblock On the logic of theory change: Partial meet contraction and revision
  functions.
\newblock {\em Journal of Symbolic Logic\/}~{\em 50,\/}~2, 510--530.

\bibitem[\protect\citeauthoryear{Alferes, Banti, Brogi, and Leite}{Alferes
  et~al\mbox{.}}{2005}]{Alferes2005}
{\sc Alferes, J.~J.}, {\sc Banti, F.}, {\sc Brogi, A.}, {\sc and} {\sc Leite,
  J.~A.} 2005.
\newblock The refined extension principle for semantics of dynamic logic
  programming.
\newblock {\em Studia Logica\/}~{\em 79,\/}~1, 7--32.

\bibitem[\protect\citeauthoryear{Alferes, Brogi, Leite, and Pereira}{Alferes
  et~al\mbox{.}}{2003}]{Alferes2003}
{\sc Alferes, J.~J.}, {\sc Brogi, A.}, {\sc Leite, J.~A.}, {\sc and} {\sc
  Pereira, L.~M.} 2003.
\newblock An evolvable rule-based e-mail agent.
\newblock In {\em Proceedings of the 11th Portuguese Conference Artificial
  Intelligence (EPIA 2003)}, {F.~Moura-Pires} {and} {S.~Abreu}, Eds. Lecture
  Notes in Computer Science, vol. 2902. Springer, Beja, Portugal, 394--408.

\bibitem[\protect\citeauthoryear{Alferes, Leite, Pereira, Przymusinska, and
  Przymusinski}{Alferes et~al\mbox{.}}{2000}]{Alferes2000}
{\sc Alferes, J.~J.}, {\sc Leite, J.~A.}, {\sc Pereira, L.~M.}, {\sc
  Przymusinska, H.}, {\sc and} {\sc Przymusinski, T.~C.} 2000.
\newblock Dynamic updates of non-monotonic knowledge bases.
\newblock {\em The Journal of Logic Programming\/}~{\em 45,\/}~1-3
  (September/October), 43--70.

\bibitem[\protect\citeauthoryear{Alferes and Pereira}{Alferes and
  Pereira}{1996}]{Alferes1996}
{\sc Alferes, J.~J.} {\sc and} {\sc Pereira, L.~M.} 1996.
\newblock Update-programs can update programs.
\newblock In {\em Proceedings of the 6th Workshop on Non-Monotonic Extensions
  of Logic Programming (NMELP 1996)}, {J.~Dix}, {L.~M. Pereira}, {and} {T.~C.
  Przymusinski}, Eds. Lecture Notes in Computer Science, vol. 1216. Springer,
  Bad Honnef, Germany, 110--131.

\bibitem[\protect\citeauthoryear{Apt and Bezem}{Apt and Bezem}{1991}]{Apt1991}
{\sc Apt, K.~R.} {\sc and} {\sc Bezem, M.} 1991.
\newblock Acyclic programs.
\newblock {\em New Generation Computing\/}~{\em 9,\/}~3/4, 335--364.

\bibitem[\protect\citeauthoryear{Apt, Blair, and Walker}{Apt
  et~al\mbox{.}}{1988}]{Apt1988}
{\sc Apt, K.~R.}, {\sc Blair, H.~A.}, {\sc and} {\sc Walker, A.} 1988.
\newblock Towards a theory of declarative knowledge.
\newblock In {\em Foundations of Deductive Databases and Logic Programming},
  {J.~Minker}, Ed. Morgan Kaufmann, San Francisco, CA, USA, 89--148.

\bibitem[\protect\citeauthoryear{Banti, Alferes, Brogi, and Hitzler}{Banti
  et~al\mbox{.}}{2005}]{Banti2005}
{\sc Banti, F.}, {\sc Alferes, J.~J.}, {\sc Brogi, A.}, {\sc and} {\sc Hitzler,
  P.} 2005.
\newblock The well supported semantics for multidimensional dynamic logic
  programs.
\newblock In {\em Proceedings of the 8th International Conference on Logic
  Programming and Nonmonotonic Reasoning (LPNMR 2005)}, {C.~Baral}, {G.~Greco},
  {N.~Leone}, {and} {G.~Terracina}, Eds. Lecture Notes in Computer Science,
  vol. 3662. Springer, Diamante, Italy, 356--368.

\bibitem[\protect\citeauthoryear{Binnewies, Zhuang, Wang, and
  Stantic}{Binnewies et~al\mbox{.}}{2018}]{DBLP:journals/tocl/BinnewiesZWS18}
{\sc Binnewies, S.}, {\sc Zhuang, Z.}, {\sc Wang, K.}, {\sc and} {\sc Stantic,
  B.} 2018.
\newblock Syntax-preserving belief change operators for logic programs.
\newblock {\em {ACM} Transactions on Computational Logic\/}~{\em 19,\/}~2,
  12:1--12:42.

\bibitem[\protect\citeauthoryear{Brewka and Eiter}{Brewka and
  Eiter}{1999}]{DBLP:journals/ai/BrewkaE99}
{\sc Brewka, G.} {\sc and} {\sc Eiter, T.} 1999.
\newblock Preferred answer sets for extended logic programs.
\newblock {\em Artificial Intelligence\/}~{\em 109,\/}~1-2, 297--356.

\bibitem[\protect\citeauthoryear{Buccafurri, Faber, and Leone}{Buccafurri
  et~al\mbox{.}}{1999}]{Buccafurri1999}
{\sc Buccafurri, F.}, {\sc Faber, W.}, {\sc and} {\sc Leone, N.} 1999.
\newblock Disjunctive logic programs with inheritance.
\newblock In {\em Proceedings of the 1999 International Conference on Logic
  Programming (ICLP 1999)}, {D.~D. Schreye}, Ed. The MIT Press, Las Cruces, New
  Mexico, USA, 79--93.

\bibitem[\protect\citeauthoryear{Cabalar and Ferraris}{Cabalar and
  Ferraris}{2007}]{Cabalar2007}
{\sc Cabalar, P.} {\sc and} {\sc Ferraris, P.} 2007.
\newblock Propositional theories are strongly equivalent to logic programs.
\newblock {\em Theory and Practice of Logic Programming\/}~{\em 7,\/}~6,
  745--759.

\bibitem[\protect\citeauthoryear{Cabalar, Pearce, and Valverde}{Cabalar
  et~al\mbox{.}}{2007}]{Cabalar2007a}
{\sc Cabalar, P.}, {\sc Pearce, D.}, {\sc and} {\sc Valverde, A.} 2007.
\newblock Minimal logic programs.
\newblock In {\em Proceedings of the 23rd International Conference on Logic
  Programming (ICLP 2007)}, {V.~Dahl} {and} {I.~Niemel{\"a}}, Eds. Lecture
  Notes in Computer Science, vol. 4670. Springer, Porto, Portugal, 104--118.

\bibitem[\protect\citeauthoryear{Calvanese, Kharlamov, Nutt, and
  Zheleznyakov}{Calvanese et~al\mbox{.}}{2010}]{Calvanese2010}
{\sc Calvanese, D.}, {\sc Kharlamov, E.}, {\sc Nutt, W.}, {\sc and} {\sc
  Zheleznyakov, D.} 2010.
\newblock Evolution of {DL-Lite} knowledge bases.
\newblock In {\em Proceedings of the 9th International Semantic Web Conference
  (ISWC 2009)}, {P.~F. Patel-Schneider}, {Y.~Pan}, {P.~Hitzler}, {P.~Mika},
  {L.~Zhang}, {J.~Z. Pan}, {I.~Horrocks}, {and} {B.~Glimm}, Eds. Lecture Notes
  in Computer Science, vol. 6496. Springer, Shanghai, China, 112--128.

\bibitem[\protect\citeauthoryear{Dalal}{Dalal}{1988}]{Dalal1988}
{\sc Dalal, M.} 1988.
\newblock Investigations into a theory of knowledge base revision.
\newblock In {\em Proceedings of the 7th National Conference on Artificial
  Intelligence (AAAI 1988)}, {H.~E. Shrobe}, {T.~M. Mitchell}, {and} {R.~G.
  Smith}, Eds. AAAI Press / The MIT Press, St. Paul, MN, USA, 475--479.

\bibitem[\protect\citeauthoryear{{De~Giacomo}, Lenzerini, Poggi, and
  Rosati}{{De~Giacomo} et~al\mbox{.}}{2009}]{Giacomo2009}
{\sc {De~Giacomo}, G.}, {\sc Lenzerini, M.}, {\sc Poggi, A.}, {\sc and} {\sc
  Rosati, R.} 2009.
\newblock On instance-level update and erasure in description logic ontologies.
\newblock {\em Journal of Logic and Computation\/}~{\em 19,\/}~5, 745--770.

\bibitem[\protect\citeauthoryear{Delgrande}{Delgrande}{2010}]{Delgrande2010}
{\sc Delgrande, J.~P.} 2010.
\newblock A {P}rogram-{L}evel {A}pproach to {R}evising {L}ogic {P}rograms under
  the {A}nswer {S}et {S}emantics.
\newblock {\em Theory and Practice of Logic Programming\/}~{\em 10,\/}~4-6
  (July), 565--580.

\bibitem[\protect\citeauthoryear{Delgrande, Peppas, and Woltran}{Delgrande
  et~al\mbox{.}}{2013}]{Delgrande2013a}
{\sc Delgrande, J.~P.}, {\sc Peppas, P.}, {\sc and} {\sc Woltran, S.} 2013.
\newblock Agm-style belief revision of logic programs under answer set
  semantics.
\newblock In {\em Proceeding of the 12th International Conference on Logic
  Programming and Nonmonotonic Reasoning (LPNMR 2013)}, {P.~Cabalar} {and}
  {T.~C. Son}, Eds. Lecture Notes in Computer Science, vol. 8148. Springer,
  264--276.

\bibitem[\protect\citeauthoryear{Delgrande, Schaub, and Tompits}{Delgrande
  et~al\mbox{.}}{2003}]{DBLP:journals/tplp/DelgrandeST03}
{\sc Delgrande, J.~P.}, {\sc Schaub, T.}, {\sc and} {\sc Tompits, H.} 2003.
\newblock A framework for compiling preferences in logic programs.
\newblock {\em Theory and Practice of Logic Programming\/}~{\em 3,\/}~2,
  129--187.

\bibitem[\protect\citeauthoryear{Delgrande, Schaub, and Tompits}{Delgrande
  et~al\mbox{.}}{2007}]{Delgrande2007}
{\sc Delgrande, J.~P.}, {\sc Schaub, T.}, {\sc and} {\sc Tompits, H.} 2007.
\newblock A preference-based framework for updating logic programs.
\newblock In {\em Proceedings of the 9th International Conference on Logic
  Programming and Nonmonotonic Reasoning (LPNMR 2007)}, {C.~Baral},
  {G.~Brewka}, {and} {J.~S. Schlipf}, Eds. Lecture Notes in Computer Science,
  vol. 4483. Springer, Tempe, AZ, USA, 71--83.

\bibitem[\protect\citeauthoryear{Delgrande, Schaub, Tompits, and
  Woltran}{Delgrande et~al\mbox{.}}{2008}]{DBLP:conf/kr/DelgrandeSTW08}
{\sc Delgrande, J.~P.}, {\sc Schaub, T.}, {\sc Tompits, H.}, {\sc and} {\sc
  Woltran, S.} 2008.
\newblock Belief revision of logic programs under answer set semantics.
\newblock In {\em Proceedings of the 11th International Conference on
  Principles of Knowledge Representation and Reasonin (KR 2008)}, {G.~Brewka}
  {and} {J.~Lang}, Eds. {AAAI} Press, 411--421.

\bibitem[\protect\citeauthoryear{Delgrande, Schaub, Tompits, and
  Woltran}{Delgrande et~al\mbox{.}}{2013}]{DBLP:journals/tocl/DelgrandeSTW13}
{\sc Delgrande, J.~P.}, {\sc Schaub, T.}, {\sc Tompits, H.}, {\sc and} {\sc
  Woltran, S.} 2013.
\newblock A model-theoretic approach to belief change in answer set
  programming.
\newblock {\em ACM Transactions on Computational Logic\/}~{\em 14,\/}~2, 14.

\bibitem[\protect\citeauthoryear{Delgrande and Wang}{Delgrande and
  Wang}{2015}]{DBLP:conf/aaai/DelgrandeW15}
{\sc Delgrande, J.~P.} {\sc and} {\sc Wang, K.} 2015.
\newblock A syntax-independent approach to forgetting in disjunctive logic
  programs.
\newblock In {\em Proceedings of the 29th AAAI Conference on Artificial
  Intelligence (AAAI 2015)}, {B.~Bonet} {and} {S.~Koenig}, Eds. {AAAI} Press,
  1482--1488.

\bibitem[\protect\citeauthoryear{Dix}{Dix}{1995}]{Dix1995a}
{\sc Dix, J.} 1995.
\newblock A classification theory of semantics of normal logic programs: {II}.
  {W}eak properties.
\newblock {\em Fundamenta Informaticae\/}~{\em 22,\/}~3, 257--288.

\bibitem[\protect\citeauthoryear{Eiter, Fink, Sabbatini, and Tompits}{Eiter
  et~al\mbox{.}}{2002}]{Eiter2002}
{\sc Eiter, T.}, {\sc Fink, M.}, {\sc Sabbatini, G.}, {\sc and} {\sc Tompits,
  H.} 2002.
\newblock On properties of update sequences based on causal rejection.
\newblock {\em Theory and Practice of Logic Programming\/}~{\em 2,\/}~6,
  721--777.

\bibitem[\protect\citeauthoryear{Eiter, Gottlob, and Mannila}{Eiter
  et~al\mbox{.}}{1997}]{44EiterGM1997}
{\sc Eiter, T.}, {\sc Gottlob, G.}, {\sc and} {\sc Mannila, H.} 1997.
\newblock Disjunctive datalog.
\newblock {\em ACM Transactions on Database Systems\/}~{\em 22,\/}~3, 364--418.

\bibitem[\protect\citeauthoryear{Erdem, Gelfond, and Leone}{Erdem
  et~al\mbox{.}}{2016}]{DBLP:journals/aim/ErdemGL16}
{\sc Erdem, E.}, {\sc Gelfond, M.}, {\sc and} {\sc Leone, N.} 2016.
\newblock Applications of answer set programming.
\newblock {\em {AI} Magazine\/}~{\em 37,\/}~3, 53--68.

\bibitem[\protect\citeauthoryear{Erdem and Patoglu}{Erdem and
  Patoglu}{2018}]{DBLP:journals/ki/ErdemP18}
{\sc Erdem, E.} {\sc and} {\sc Patoglu, V.} 2018.
\newblock Applications of {ASP} in robotics.
\newblock {\em K{\"{u}}nstliche Intelligenz\/}~{\em 32,\/}~2-3, 143--149.

\bibitem[\protect\citeauthoryear{Fages}{Fages}{1994}]{DBLP:journals/mlcs/Fages94}
{\sc Fages, F.} 1994.
\newblock Consistency of clark's completion and existence of stable models.
\newblock {\em Methods of Logic in Computer Science\/}~{\em 1,\/}~1, 51--60.

\bibitem[\protect\citeauthoryear{Falkner, Friedrich, Schekotihin, Taupe, and
  Teppan}{Falkner et~al\mbox{.}}{2018}]{DBLP:journals/ki/FalknerFSTT18}
{\sc Falkner, A.~A.}, {\sc Friedrich, G.}, {\sc Schekotihin, K.}, {\sc Taupe,
  R.}, {\sc and} {\sc Teppan, E.~C.} 2018.
\newblock Industrial applications of answer set programming.
\newblock {\em K{\"{u}}nstliche Intelligenz\/}~{\em 32,\/}~2-3, 165--176.

\bibitem[\protect\citeauthoryear{Ferm{\'{e}} and Hansson}{Ferm{\'{e}} and
  Hansson}{2011}]{DBLP:journals/jphil/FermeH11a}
{\sc Ferm{\'{e}}, E.~L.} {\sc and} {\sc Hansson, S.~O.} 2011.
\newblock {AGM} 25 years - twenty-five years of research in belief change.
\newblock {\em Journal of Philosophical Logic\/}~{\em 40,\/}~2, 295--331.

\bibitem[\protect\citeauthoryear{Garcia, Lef{\`{e}}vre, St{\'{e}}phan, Papini,
  and W{\"{u}}rbel}{Garcia
  et~al\mbox{.}}{2019}]{DBLP:journals/jair/GarciaLSPW19}
{\sc Garcia, L.}, {\sc Lef{\`{e}}vre, C.}, {\sc St{\'{e}}phan, I.}, {\sc
  Papini, O.}, {\sc and} {\sc W{\"{u}}rbel, E.} 2019.
\newblock A semantic characterization {ASP} base revision.
\newblock {\em Journal of Artificial Intelligence Research\/}~{\em 66},
  989--1029.

\bibitem[\protect\citeauthoryear{G{\"a}rdenfors}{G{\"a}rdenfors}{1992}]{Gardenfors1992}
{\sc G{\"a}rdenfors, P.} 1992.
\newblock {\em Belief Revision}.
\newblock Cambridge University Press, Chapter Belief Revision: An Introduction,
  1--28.

\bibitem[\protect\citeauthoryear{Gebser, Kaufmann, Kaminski, Ostrowski, Schaub,
  and Schneider}{Gebser
  et~al\mbox{.}}{2011}]{DBLP:journals/aicom/GebserKKOSS11}
{\sc Gebser, M.}, {\sc Kaufmann, B.}, {\sc Kaminski, R.}, {\sc Ostrowski, M.},
  {\sc Schaub, T.}, {\sc and} {\sc Schneider, M.~T.} 2011.
\newblock Potassco: The potsdam answer set solving collection.
\newblock {\em {AI} Communincations\/}~{\em 24,\/}~2, 107--124.

\bibitem[\protect\citeauthoryear{Gelfond and Lifschitz}{Gelfond and
  Lifschitz}{1988}]{Gelfond1988}
{\sc Gelfond, M.} {\sc and} {\sc Lifschitz, V.} 1988.
\newblock The stable model semantics for logic programming.
\newblock In {\em Proceedings of the 5th International Conference and Symposium
  on Logic Programming (ICLP/SLP 1988)}, {R.~A. Kowalski} {and} {K.~A. Bowen},
  Eds. MIT Press, Seattle, Washington, 1070--1080.

\bibitem[\protect\citeauthoryear{Gelfond and Lifschitz}{Gelfond and
  Lifschitz}{1991}]{Gelfond1991}
{\sc Gelfond, M.} {\sc and} {\sc Lifschitz, V.} 1991.
\newblock Classical negation in logic programs and disjunctive databases.
\newblock {\em New Generation Computing\/}~{\em 9,\/}~3-4, 365--385.

\bibitem[\protect\citeauthoryear{Gon{\c{c}}alves, Knorr, and
  Leite}{Gon{\c{c}}alves et~al\mbox{.}}{2016}]{DBLP:conf/epia/GoncalvesKL16}
{\sc Gon{\c{c}}alves, R.}, {\sc Knorr, M.}, {\sc and} {\sc Leite, J.} 2016.
\newblock The ultimate guide to forgetting in answer-set programming.
\newblock In {\em Proceedings of the 15th International Conference on
  Principles of Knowledge Representation and Reasoning (KR 2016)}, {C.~Baral},
  {J.~Delgrande}, {and} {F.~Wolter}, Eds. {AAAI} Press.

\bibitem[\protect\citeauthoryear{Hansson}{Hansson}{1993}]{Hansson1993}
{\sc Hansson, S.~O.} 1993.
\newblock Reversing the {Levi} identity.
\newblock {\em Journal of Philosophical Logic\/}~{\em 22,\/}~6 (December),
  637--669.

\bibitem[\protect\citeauthoryear{Herzig and Rifi}{Herzig and
  Rifi}{1999}]{Herzig1999}
{\sc Herzig, A.} {\sc and} {\sc Rifi, O.} 1999.
\newblock Propositional belief base update and minimal change.
\newblock {\em Artificial Intelligence\/}~{\em 115,\/}~1, 107--138.

\bibitem[\protect\citeauthoryear{Heyting}{Heyting}{1930}]{Heyting1930}
{\sc Heyting, A.} 1930.
\newblock Die formalen {R}egeln der intuitionistischen {L}ogik.
\newblock {\em Sitzungsberichte der Preussischen {A}kademie der
  {W}issenschaften\/}, 42--56.
\newblock Reprint in \emph{{L}ogik-{T}exte: Kommentierte {A}uswahl zur
  {G}eschichte der Modernen {L}ogik}, Akademie-Verlag, 1986.

\bibitem[\protect\citeauthoryear{Hitzler and Wendt}{Hitzler and
  Wendt}{2005}]{DBLP:journals/tplp/HitzlerW05}
{\sc Hitzler, P.} {\sc and} {\sc Wendt, M.} 2005.
\newblock A uniform approach to logic programming semantics.
\newblock {\em Theory and Practice of Logic Programming\/}~{\em 5,\/}~1-2,
  93--121.

\bibitem[\protect\citeauthoryear{Homola}{Homola}{2004}]{Homola2004}
{\sc Homola, M.} 2004.
\newblock Dynamic logic programming: Various semantics are equal on acyclic
  programs.
\newblock In {\em Proceedings of the 5th International Workshop on
  Computational Logic in Multi-Agent Systems (CLIMA V)}, {J.~A. Leite} {and}
  {P.~Torroni}, Eds. Lecture Notes in Computer Science, vol. 3487. Springer,
  Lisbon, Portugal, 78--95.

\bibitem[\protect\citeauthoryear{Ilic, Leite, and Slota}{Ilic
  et~al\mbox{.}}{2008}]{Ilic2008}
{\sc Ilic, M.}, {\sc Leite, J.}, {\sc and} {\sc Slota, M.} 2008.
\newblock Explicit dynamic user profiles for a collaborative filtering
  recommender system.
\newblock In {\em Proceedings of the 11th Ibero-American Conference on
  Artificial Intelligence (IBERAMIA 2008)}, {H.~Geffner}, {R.~Prada}, {I.~M.
  Alexandre}, {and} {N.~David}, Eds. Vol. LNAI 5290. Springer-Verlag, 352--361.

\bibitem[\protect\citeauthoryear{Inoue and Sakama}{Inoue and
  Sakama}{2004}]{Inoue2004}
{\sc Inoue, K.} {\sc and} {\sc Sakama, C.} 2004.
\newblock Equivalence of logic programs under updates.
\newblock In {\em Proceedings of the 9th European Conference on Logics in
  Artificial Intelligence (JELIA 2004)}, {J.~J. Alferes} {and} {J.~A. Leite},
  Eds. Lecture Notes in Computer Science, vol. 3229. Springer, Lisbon,
  Portugal, 174--186.

\bibitem[\protect\citeauthoryear{Katsuno and Mendelzon}{Katsuno and
  Mendelzon}{1989}]{Katsuno1989}
{\sc Katsuno, H.} {\sc and} {\sc Mendelzon, A.~O.} 1989.
\newblock A unified view of propositional knowledge base updates.
\newblock In {\em Proceedings of the 11th International Joint Conference on
  Artificial Intelligence (IJCAI 1989)}, {N.~S. Sridharan}, Ed. Morgan
  Kaufmann, 1413--1419.

\bibitem[\protect\citeauthoryear{Katsuno and Mendelzon}{Katsuno and
  Mendelzon}{1991}]{Katsuno1991}
{\sc Katsuno, H.} {\sc and} {\sc Mendelzon, A.~O.} 1991.
\newblock On the difference between updating a knowledge base and revising it.
\newblock In {\em Proceedings of the 2nd International Conference on Principles
  of Knowledge Representation and Reasoning (KR 1991)}, {J.~F. Allen},
  {R.~Fikes}, {and} {E.~Sandewall}, Eds. Morgan Kaufmann Publishers, Cambridge,
  MA, USA, 387--394.

\bibitem[\protect\citeauthoryear{Keller and Winslett}{Keller and
  Winslett}{1985}]{Keller1985}
{\sc Keller, A.~M.} {\sc and} {\sc Winslett, M.} 1985.
\newblock On the use of an extended relational model to handle changing
  incomplete information.
\newblock {\em IEEE Transactions on Software Engineering\/}~{\em 11,\/}~7,
  620--633.

\bibitem[\protect\citeauthoryear{Kr{\"u}mpelmann}{Kr{\"u}mpelmann}{2012}]{Krumpelmann2012}
{\sc Kr{\"u}mpelmann, P.} 2012.
\newblock Dependency semantics for sequences of extended logic programs.
\newblock {\em Logic Journal of the IGPL\/}~{\em 20,\/}~5, 943--966.

\bibitem[\protect\citeauthoryear{Kr{\"u}mpelmann and
  Kern-Isberner}{Kr{\"u}mpelmann and Kern-Isberner}{2010}]{Krumpelmann2010}
{\sc Kr{\"u}mpelmann, P.} {\sc and} {\sc Kern-Isberner, G.} 2010.
\newblock On belief dynamics of dependency relations for extended logic
  programs.
\newblock In {\em Proceedings of the 13th International Workshop on
  Non-Monotonic Reasoning (NMR 2010)}, {T.~Meyer} {and} {E.~Ternovska}, Eds.
  Toronto, Canada.

\bibitem[\protect\citeauthoryear{Leite}{Leite}{2003}]{Leite2003}
{\sc Leite, J.~A.} 2003.
\newblock {\em Evolving Knowledge Bases}. Frontiers of Artificial Intelligence
  and Applications, xviii + 307 p. Hardcover, vol.~81.
\newblock IOS Press.

\bibitem[\protect\citeauthoryear{Leite and Pereira}{Leite and
  Pereira}{1998}]{Leite1997}
{\sc Leite, J.~A.} {\sc and} {\sc Pereira, L.~M.} 1998.
\newblock Generalizing updates: From models to programs.
\newblock In {\em Proceedings of the 3rd International Workshop on Logic
  Programming and Knowledge Representation (LPKR 1997)}, {J.~Dix}, {L.~M.
  Pereira}, {and} {T.~C. Przymusinski}, Eds. Lecture Notes in Computer Science,
  vol. 1471. Springer, 224--246.

\bibitem[\protect\citeauthoryear{Lenzerini and Savo}{Lenzerini and
  Savo}{2011}]{Lenzerini2011}
{\sc Lenzerini, M.} {\sc and} {\sc Savo, D.~F.} 2011.
\newblock On the evolution of the instance level of {DL-Lite} knowledge bases.
\newblock In {\em Proceedings of the 24th International Workshop on Description
  Logics (DL 2011)}, {R.~Rosati}, {S.~Rudolph}, {and} {M.~Zakharyaschev}, Eds.
  CEUR Workshop Proceedings, vol. 745. CEUR-WS.org, Barcelona, Spain.

\bibitem[\protect\citeauthoryear{Leone, Pfeifer, Faber, Eiter, Gottlob, Perri,
  and Scarcello}{Leone et~al\mbox{.}}{2006}]{DBLP:journals/tocl/LeonePFEGPS06}
{\sc Leone, N.}, {\sc Pfeifer, G.}, {\sc Faber, W.}, {\sc Eiter, T.}, {\sc
  Gottlob, G.}, {\sc Perri, S.}, {\sc and} {\sc Scarcello, F.} 2006.
\newblock The {DLV} system for knowledge representation and reasoning.
\newblock {\em ACM Transactions on Computational Logic\/}~{\em 7,\/}~3,
  499--562.

\bibitem[\protect\citeauthoryear{Lifschitz}{Lifschitz}{1999}]{Lifschitz1999}
{\sc Lifschitz, V.} 1999.
\newblock Action languages, answer sets, and planning.
\newblock In {\em The Logic Programming Paradigm: A 25-Year Perspective},
  {K.~R. Apt}, {V.~W. Marek}, {M.~Truszczynski}, {and} {D.~S. Warren}, Eds.
  Springer Berlin Heidelberg, Berlin, Heidelberg, 357--373.

\bibitem[\protect\citeauthoryear{Lifschitz}{Lifschitz}{2008}]{DBLP:conf/iclp/Lifschitz08}
{\sc Lifschitz, V.} 2008.
\newblock Twelve definitions of a stable model.
\newblock In {\em Proceedings of the 24th International Conference on Logic
  Programming (ICLP 2008)}, {M.~G. de~la Banda} {and} {E.~Pontelli}, Eds.
  Lecture Notes in Computer Science, vol. 5366. Springer, 37--51.

\bibitem[\protect\citeauthoryear{Lifschitz, Pearce, and Valverde}{Lifschitz
  et~al\mbox{.}}{2001}]{Lifschitz2001}
{\sc Lifschitz, V.}, {\sc Pearce, D.}, {\sc and} {\sc Valverde, A.} 2001.
\newblock Strongly equivalent logic programs.
\newblock {\em ACM Transactions on Computational Logic\/}~{\em 2,\/}~4,
  526--541.

\bibitem[\protect\citeauthoryear{Liu, Lutz, Mili{\v{c}}i{\'c}, and Wolter}{Liu
  et~al\mbox{.}}{2006}]{Liu2006}
{\sc Liu, H.}, {\sc Lutz, C.}, {\sc Mili{\v{c}}i{\'c}, M.}, {\sc and} {\sc
  Wolter, F.} 2006.
\newblock Updating description logic {ABoxes}.
\newblock In {\em Proceedings of the 10th International Conference on
  Principles of Knowledge Representation and Reasoning (KR 2006)},
  {P.~Doherty}, {J.~Mylopoulos}, {and} {C.~A. Welty}, Eds. AAAI Press, Lake
  District of the United Kingdom, 46--56.

\bibitem[\protect\citeauthoryear{{\L}ukasiewicz}{{\L}ukasiewicz}{1941}]{Lukasiewicz1941}
{\sc {\L}ukasiewicz, J.} 1941.
\newblock Die {L}ogik und das {G}rundlagenproblem.
\newblock In {\em {L}es Entretiens de Z{\"u}rich sue les Fondements et la
  m{\'e}thode des sciences math{\'e}matiques 1938}. Z{\"u}rich, 82--100.

\bibitem[\protect\citeauthoryear{Marek and Truszczynski}{Marek and
  Truszczynski}{1994}]{DBLP:conf/jelia/MarekT94}
{\sc Marek, V.~W.} {\sc and} {\sc Truszczynski, M.} 1994.
\newblock Revision specifications by means of programs.
\newblock In {\em Proceedings of the 4th European Workshop on Logics in
  Artificial Intelligence (JELIA 2094)}, {C.~MacNish}, {D.~Pearce}, {and}
  {L.~M. Pereira}, Eds. Lecture Notes in Computer Science, vol. 838. Springer,
  122--136.

\bibitem[\protect\citeauthoryear{Marek and Truszczynski}{Marek and
  Truszczynski}{1998}]{Marek1998}
{\sc Marek, V.~W.} {\sc and} {\sc Truszczynski, M.} 1998.
\newblock Revision programming.
\newblock {\em Theoretical Computer Science\/}~{\em 190,\/}~2, 241--277.

\bibitem[\protect\citeauthoryear{Marek and Truszczy{\'{n}}ski}{Marek and
  Truszczy{\'{n}}ski}{1999}]{Marek1999}
{\sc Marek, V.~W.} {\sc and} {\sc Truszczy{\'{n}}ski, M.} 1999.
\newblock Stable models and an alternative logic programming paradigm.
\newblock In {\em The Logic Programming Paradigm: A 25-Year Perspective},
  {K.~R. Apt}, {V.~W. Marek}, {M.~Truszczynski}, {and} {D.~S. Warren}, Eds.
  Springer Berlin Heidelberg, Berlin, Heidelberg, 375--398.

\bibitem[\protect\citeauthoryear{McCarthy}{McCarthy}{1998}]{McCarthy98}
{\sc McCarthy, J.} 1998.
\newblock Elaboration tolerance.
\newblock In {\em Working Papers of the Fourth International Symposium on
  Logical formalizations of Commonsense Reasoning (Commonsense 1998)}.

\bibitem[\protect\citeauthoryear{Newtono}{Newtono}{1726}]{Principia}
{\sc Newtono, I.} 1726.
\newblock {\em Philosophi\ae\ Naturalis Principia Mathematica}.
\newblock Editio tertia \& aucta emendata. Apud Guil \& Joh. Innys, Regi\ae\
  Societatis typographos.

\bibitem[\protect\citeauthoryear{Niemel{\"{a}}}{Niemel{\"{a}}}{1999}]{DBLP:journals/amai/Niemela99}
{\sc Niemel{\"{a}}, I.} 1999.
\newblock Logic programs with stable model semantics as a constraint
  programming paradigm.
\newblock {\em Annals of Mathematics and Artificial Intelligence\/}~{\em
  25,\/}~3-4, 241--273.

\bibitem[\protect\citeauthoryear{Osorio and Cuevas}{Osorio and
  Cuevas}{2007}]{Osorio2007}
{\sc Osorio, M.} {\sc and} {\sc Cuevas, V.} 2007.
\newblock Updates in answer set programming: An approach based on basic
  structural properties.
\newblock {\em Theory and Practice of Logic Programming\/}~{\em 7,\/}~4,
  451--479.

\bibitem[\protect\citeauthoryear{Osorio and Zepeda}{Osorio and
  Zepeda}{2007}]{Osorio2007a}
{\sc Osorio, M.} {\sc and} {\sc Zepeda, C.} 2007.
\newblock Update sequences based on minimal generalized pstable models.
\newblock In {\em Proceedings of the 6th Mexican International Conference on
  Artificial Intelligence (MICAI 2007)}, {A.~F. Gelbukh} {and} {A.~F.~K.
  Morales}, Eds. Lecture Notes in Computer Science, vol. 4827. Springer,
  Aguascalientes, Mexico, 283--293.

\bibitem[\protect\citeauthoryear{Pearce}{Pearce}{1997}]{Pearce1997}
{\sc Pearce, D.} 1997.
\newblock A new logical characterisation of stable models and answer sets.
\newblock In {\em Proceedings of the 6th Workshop on Non-Monotonic Extensions
  of Logic Programming (NMELP 1996)}, {J.~Dix}, {L.~M. Pereira}, {and} {T.~C.
  Przymusinski}, Eds. Lecture Notes in Computer Science, vol. 1216. Springer,
  Bad Honnef, Germany, 57--70.

\bibitem[\protect\citeauthoryear{Przymusinski and Turner}{Przymusinski and
  Turner}{1995}]{DBLP:conf/lpnmr/PrzymusinskiT95}
{\sc Przymusinski, T.~C.} {\sc and} {\sc Turner, H.} 1995.
\newblock Update by means of inference rules.
\newblock In {\em Proceedings of the 3rd International Conference on Logic
  Programming and Nonmonotonic Reasoning (LPNMR 1995)}, {V.~W. Marek} {and}
  {A.~Nerode}, Eds. Lecture Notes in Computer Science, vol. 928. Springer,
  156--174.

\bibitem[\protect\citeauthoryear{Przymusinski and Turner}{Przymusinski and
  Turner}{1997}]{DBLP:journals/jlp/PrzymusinskiT97}
{\sc Przymusinski, T.~C.} {\sc and} {\sc Turner, H.} 1997.
\newblock Update by means of inference rules.
\newblock {\em The Journal of Logic Programming\/}~{\em 30,\/}~2, 125--143.

\bibitem[\protect\citeauthoryear{Saias and Quaresma}{Saias and
  Quaresma}{2004}]{Saias2004}
{\sc Saias, J.} {\sc and} {\sc Quaresma, P.} 2004.
\newblock A methodology to create legal ontologies in a logic programming based
  web information retrieval system.
\newblock {\em Artificial Intelligence and Law\/}~{\em 12,\/}~4, 397--417.

\bibitem[\protect\citeauthoryear{Sakama and Inoue}{Sakama and
  Inoue}{2003}]{Sakama2003}
{\sc Sakama, C.} {\sc and} {\sc Inoue, K.} 2003.
\newblock An abductive framework for computing knowledge base updates.
\newblock {\em Theory and Practice of Logic Programming\/}~{\em 3,\/}~6,
  671--713.

\bibitem[\protect\citeauthoryear{Satoh}{Satoh}{1988}]{Satoh1988}
{\sc Satoh, K.} 1988.
\newblock Nonmonotonic reasoning by minimal belief revision.
\newblock In {\em Proceedings of the International Conference on Fifth
  Generation Computer Systems, (FGCS 1988)}. {OHMSHA} Ltd. Tokyo and
  Springer-Verlag, 455--462.

\bibitem[\protect\citeauthoryear{Schaub and Wang}{Schaub and
  Wang}{2003}]{Schaub2003}
{\sc Schaub, T.} {\sc and} {\sc Wang, K.} 2003.
\newblock A semantic framework for preference handling in answer set
  programming.
\newblock {\em Theory and Practice of Logic Programming\/}~{\em 3,\/}~4-5,
  569--607.

\bibitem[\protect\citeauthoryear{Schwind and Inoue}{Schwind and
  Inoue}{2016}]{DBLP:journals/tplp/SchwindI16}
{\sc Schwind, N.} {\sc and} {\sc Inoue, K.} 2016.
\newblock Characterization of logic program revision as an extension of
  propositional revision.
\newblock {\em Theory and Practice of Logic Programming\/}~{\em 16,\/}~1,
  111--138.

\bibitem[\protect\citeauthoryear{{\v{S}}efr{\'a}nek}{{\v{S}}efr{\'a}nek}{2006}]{Sefranek2006}
{\sc {\v{S}}efr{\'a}nek, J.} 2006.
\newblock Irrelevant updates and nonmonotonic assumptions.
\newblock In {\em Proceedings of the 10th European Conference on Logics in
  Artificial Intelligence (JELIA 2006)}, {M.~Fisher}, {W.~van~der Hoek},
  {B.~Konev}, {and} {A.~Lisitsa}, Eds. Lecture Notes in Computer Science, vol.
  4160. Springer, Liverpool, UK, 426--438.

\bibitem[\protect\citeauthoryear{{\v{S}}efr{\'a}nek}{{\v{S}}efr{\'a}nek}{2011}]{Sefranek2011}
{\sc {\v{S}}efr{\'a}nek, J.} 2011.
\newblock Static and dynamic semantics: Preliminary report.
\newblock In {\em Proceedings of the 10th Mexican International Conference on
  Artificial Intelligence (MICAI 2011)}, {I.~Z. Batyrshin} {and} {G.~Sidorov},
  Eds. IEEE Computer Society, Los Alamitos, CA, USA, 36--42.

\bibitem[\protect\citeauthoryear{Siska}{Siska}{2006}]{Siska2006}
{\sc Siska, J.} 2006.
\newblock Dynamic logic programming and world state evaluation in computer
  games.
\newblock In {\em Proceedings of the 20th Workshop on Logic Programming (WLP
  2006)}, {M.~Fink}, {H.~Tompits}, {and} {S.~Woltran}, Eds. INFSYS Research
  Report, vol. 1843-06-02. Technische Universit{\"a}t Wien, Austria, Vienna,
  Austria, 64--70.

\bibitem[\protect\citeauthoryear{Slota}{Slota}{2012}]{Slota2012Thesis}
{\sc Slota, M.} 2012.
\newblock Updates of hybrid knowledge bases.
\newblock Ph.D. thesis, Universidade Nova de Lisboa.

\bibitem[\protect\citeauthoryear{Slota and Leite}{Slota and
  Leite}{2010}]{Slota2010b}
{\sc Slota, M.} {\sc and} {\sc Leite, J.} 2010.
\newblock On semantic update operators for answer-set programs.
\newblock In {\em Proceedings of the 19th European Conference on Artificial
  Intelligence (ECAI 2010)}, {H.~Coelho}, {R.~Studer}, {and} {M.~Wooldridge},
  Eds. Frontiers in Artificial Intelligence and Applications, vol. 215. IOS
  Press, Lisbon, Portugal, 957--962.

\bibitem[\protect\citeauthoryear{Slota and Leite}{Slota and
  Leite}{2011}]{Slota2011}
{\sc Slota, M.} {\sc and} {\sc Leite, J.} 2011.
\newblock Back and forth between rules and {SE}-models.
\newblock In {\em Proceedings of the 11th International Conference on Logic
  Programming and Nonmonotonic Reasoning (LPNMR 2011)}, {J.~P. Delgrande} {and}
  {W.~Faber}, Eds. Lecture Notes in Computer Science, vol. 6645. Springer,
  Vancouver, Canada, 174--186.

\bibitem[\protect\citeauthoryear{Slota and Leite}{Slota and
  Leite}{2012a}]{Slota2012}
{\sc Slota, M.} {\sc and} {\sc Leite, J.} 2012a.
\newblock Robust equivalence models for semantic updates of answer-set
  programs.
\newblock In {\em Proceedings of the 13th International Conference on
  Principles of Knowledge Representation and Reasoning (KR 2012)}, {G.~Brewka},
  {T.~Eiter}, {and} {S.~A. McIlraith}, Eds. AAAI Press, Rome, Italy, 158--168.

\bibitem[\protect\citeauthoryear{Slota and Leite}{Slota and
  Leite}{2012b}]{Slota2012a}
{\sc Slota, M.} {\sc and} {\sc Leite, J.} 2012b.
\newblock A unifying perspective on knowledge updates.
\newblock In {\em Proceedings of the 13th European Conference on Logics in
  Artificial Intelligence (JELIA 2012)}, {L.~F. del Cerro}, {A.~Herzig}, {and}
  {J.~Mengin}, Eds. Logics in Artificial Intelligence (LNAI), vol. 7519.
  Springer, Toulouse, France, 372--384.

\bibitem[\protect\citeauthoryear{Slota and Leite}{Slota and
  Leite}{2013}]{Slota2013}
{\sc Slota, M.} {\sc and} {\sc Leite, J.} 2013.
\newblock On condensing a sequence of updates in answer-set programming.
\newblock In {\em Proceedings of the 23rd International Joint Conference on
  Artificial Intelligence (IJCAI 2013)}, {F.~Rossi}, Ed. IJCAI/AAAI,
  1097--1103.

\bibitem[\protect\citeauthoryear{Slota and Leite}{Slota and
  Leite}{2014}]{Slota2013a}
{\sc Slota, M.} {\sc and} {\sc Leite, J.} 2014.
\newblock The rise and fall of semantic rule updates based on {SE}-models.
\newblock {\em Theory and Practice of Logic Programming\/}~{\em 14,\/}~6,
  869--907.

\bibitem[\protect\citeauthoryear{Slota, Leite, and Swift}{Slota
  et~al\mbox{.}}{2011}]{Slota2011a}
{\sc Slota, M.}, {\sc Leite, J.}, {\sc and} {\sc Swift, T.} 2011.
\newblock Splitting and updating hybrid knowledge bases.
\newblock {\em Theory and Practice of Logic Programming\/}~{\em 11,\/}~4-5,
  801--819.

\bibitem[\protect\citeauthoryear{Slota, Leite, and Swift}{Slota
  et~al\mbox{.}}{2015}]{DBLP:journals/ai/SlotaLS15}
{\sc Slota, M.}, {\sc Leite, J.}, {\sc and} {\sc Swift, T.} 2015.
\newblock On updates of hybrid knowledge bases composed of ontologies and
  rules.
\newblock {\em Artificial Intelligence\/}~{\em 229}, 33--104.

\bibitem[\protect\citeauthoryear{Wang, Wang, and Zhang}{Wang
  et~al\mbox{.}}{2013}]{WangWZ13}
{\sc Wang, Y.}, {\sc Wang, K.}, {\sc and} {\sc Zhang, M.} 2013.
\newblock Forgetting for answer set programs revisited.
\newblock In {\em Proceedings of the 23rd International Joint Conference on
  Artificial Intelligence (IJCAI 2013)}, {F.~Rossi}, Ed. IJCAI/AAAI.

\bibitem[\protect\citeauthoryear{Wang, Zhang, Zhou, and Zhang}{Wang
  et~al\mbox{.}}{2012}]{WangZZZ12}
{\sc Wang, Y.}, {\sc Zhang, Y.}, {\sc Zhou, Y.}, {\sc and} {\sc Zhang, M.}
  2012.
\newblock Forgetting in logic programs under strong equivalence.
\newblock In {\em Proceedings of the 13th International Conference on
  Principles of Knowledge Representation and Reasoning (KR 2012)}, {G.~Brewka},
  {T.~Eiter}, {and} {S.~A. McIlraith}, Eds. AAAI Press, 643--647.

\bibitem[\protect\citeauthoryear{Wang, Zhang, Zhou, and Zhang}{Wang
  et~al\mbox{.}}{2014}]{WangZZZ14}
{\sc Wang, Y.}, {\sc Zhang, Y.}, {\sc Zhou, Y.}, {\sc and} {\sc Zhang, M.}
  2014.
\newblock Knowledge forgetting in answer set programming.
\newblock {\em Journal of Artificial Intelligence Research\/}~{\em 50}, 31--70.

\bibitem[\protect\citeauthoryear{Winslett}{Winslett}{1988}]{Winslett1988}
{\sc Winslett, M.} 1988.
\newblock Reasoning about action using a possible models approach.
\newblock In {\em Proceedings of the 7th National Conference on Artificial
  Intelligence (AAAI 1988)}, {H.~E. Shrobe}, {T.~M. Mitchell}, {and} {R.~G.
  Smith}, Eds. AAAI Press / The MIT Press, Saint Paul, MN, USA, 89--93.

\bibitem[\protect\citeauthoryear{Winslett}{Winslett}{1990}]{Winslett1990}
{\sc Winslett, M.} 1990.
\newblock {\em Updating Logical Databases}.
\newblock Cambridge University Press, New York, USA.

\bibitem[\protect\citeauthoryear{Zhang}{Zhang}{2003}]{Zhang2003}
{\sc Zhang, Y.} 2003.
\newblock Two results for prioritized logic programming.
\newblock {\em Theory and Practice of Logic Programming\/}~{\em 3,\/}~2,
  223--242.

\bibitem[\protect\citeauthoryear{Zhang}{Zhang}{2006}]{Zhang2006}
{\sc Zhang, Y.} 2006.
\newblock Logic program-based updates.
\newblock {\em ACM Transactions on Computational Logic\/}~{\em 7,\/}~3,
  421--472.

\bibitem[\protect\citeauthoryear{Zhang and Foo}{Zhang and
  Foo}{1998}]{Zhang1998}
{\sc Zhang, Y.} {\sc and} {\sc Foo, N.~Y.} 1998.
\newblock Updating logic programs.
\newblock In {\em Proceedings of the 13th European Conference on Artificial
  Intelligence (ECAI 1998)}, {H.~Prade}, Ed. John Wiley and Sons, Chichester,
  Brighton, UK, 403--407.

\bibitem[\protect\citeauthoryear{Zhang and Foo}{Zhang and
  Foo}{2005}]{Zhang2005}
{\sc Zhang, Y.} {\sc and} {\sc Foo, N.~Y.} 2005.
\newblock A unified framework for representing logic program updates.
\newblock In {\em Proceedings of the 20th National Conference on Artificial
  Intelligence (AAAI 2005)}, {M.~M. Veloso} {and} {S.~Kambhampati}, Eds. AAAI
  Press / The MIT Press, Pittsburgh, Pennsylvania, USA, 707--713.

\bibitem[\protect\citeauthoryear{Zhuang, Delgrande, Nayak, and Sattar}{Zhuang
  et~al\mbox{.}}{2016}]{DBLP:conf/ecai/ZhuangDNS16}
{\sc Zhuang, Z.}, {\sc Delgrande, J.~P.}, {\sc Nayak, A.~C.}, {\sc and} {\sc
  Sattar, A.} 2016.
\newblock Reconsidering agm-style belief revision in the context of logic
  programs.
\newblock In {\em Proceedings of the 22nd European Conference on Artificial
  Intelligence (ECAI 2016)}, {G.~A. Kaminka}, {M.~Fox}, {P.~Bouquet},
  {E.~H{\"{u}}llermeier}, {V.~Dignum}, {F.~Dignum}, {and} {F.~van Harmelen},
  Eds. Frontiers in Artificial Intelligence and Applications, vol. 285. {IOS}
  Press, 671--679.

\end{thebibliography}

\end{document}